\documentclass[10pt,twocolumn,letterpaper]{article}
\usepackage[accsupp]{axessibility}
\usepackage{iccv}
\usepackage{times}
\usepackage{epsfig}
\usepackage{graphicx}
\usepackage{amsmath}
\usepackage{amssymb}
\usepackage{algpseudocode}
\usepackage{caption} 
\usepackage{adjustbox}
\usepackage[ruled, lined, linesnumbered, commentsnumbered, longend]{algorithm2e}
\usepackage{enumitem}
\usepackage{multirow}
\usepackage[x11names]{xcolor}
\usepackage{colortbl}
\usepackage{todonotes}
\usepackage{booktabs}
\usepackage{soul}
\usepackage{makecell}
\setuptodonotes{size=\footnotesize}
\SetKwInOut{KwIn}{Input}
\SetKwInOut{KwOut}{Output}
\SetKwInOut{KwRe}{Return}

\DeclareRobustCommand{\hlgray}[1]{{\sethlcolor{Snow2}\hl{#1}}}
\DeclareRobustCommand{\hlcyan}[1]{{\sethlcolor{LightCyan1}\hl{#1}}}

\makeatletter
\newcommand\HUGE{\@setfontsize\Huge{38}{47}}
\makeatother    
\SetCommentSty{mycommfont}
\makeatletter
\def\thickhline{%
  \noalign{\ifnum0=`}\fi\hrule \@height \thickarrayrulewidth \futurelet
   \reserved@a\@xthickhline}
\def\@xthickhline{\ifx\reserved@a\thickhline
               \vskip\doublerulesep
               \vskip-\thickarrayrulewidth
             \fi
      \ifnum0=`{\fi}}
\makeatother

\newlength{\thickarrayrulewidth}
\usepackage[pagebackref,breaklinks,colorlinks]{hyperref}
\hypersetup{linkcolor=red,anchorcolor=blue,citecolor=cyan,filecolor=blue,urlcolor=blue}

\usepackage[capitalize]{cleveref}
\crefname{section}{Sec.}{Secs.}
\Crefname{section}{Section}{Sections}
\Crefname{table}{Table}{Tables}
\crefname{table}{Tab.}{Tabs.}
\crefname{equation}{Eq.}{Eqs.}
\iccvfinalcopy 

\begin{document}

\title{Adversarial Bayesian Augmentation for Single-Source Domain Generalization}

\author{
Sheng Cheng \qquad Tejas Gokhale \qquad Yezhou Yang\\
Active Perception Group, Arizona State University\\
{\tt\small \{scheng53, tgokhale, yz.yang\}@asu.edu}
}

\maketitle
\ificcvfinal\thispagestyle{empty}\fi

\begin{abstract}
Generalizing to unseen image domains is a challenging problem primarily due to the lack of diverse training data, inaccessible target data, and the large domain shift that may exist in many real-world settings.
As such data augmentation is a critical component of domain generalization methods that seek to address this problem.
We present Adversarial Bayesian Augmentation (ABA), a novel algorithm that learns to generate image augmentations in the challenging single-source domain generalization setting.
ABA draws on the strengths of adversarial learning and Bayesian neural networks to guide the generation of diverse data augmentations -- these synthesized image domains aid the classifier in generalizing to unseen domains.
We demonstrate the strength of ABA on several types of domain shift including style shift, subpopulation shift, and shift in the medical imaging setting.
ABA outperforms all previous state-of-the-art methods, including pre-specified augmentations, pixel-based and convolutional-based augmentations.
Code: \url{https://github.com/shengcheng/ABA}.

\end{abstract}

\section{Introduction}

Improving the generalization of deep neural networks to out-of-distribution samples is a fundamental yet challenging problem in machine learning and computer vision~\cite{wang2018learning, krueger2021out, muandet2013domain}.
Typically, neural networks are trained and tested on data samples from the same distribution (under the \textit{i.i.d.} assumption); under this setting, image classifiers have achieved impressive performances.
However, in real-world applications, the distribution of test samples can drastically differ from the training samples~\cite{volpi2018generalizing,nguyen2015deep}. 
This is especially problematic when the process of acquiring labeled samples from the target test domain is expensive or infeasible, making it difficult to apply semi-supervised learning for domain adaptation~\cite{zhang2021semi,yao2015semi}.
Therefore, there is a need to develop techniques that enable deep neural networks to capture the domain-invariant patterns in the data~\cite{muandet2013domain, xiao2021bit}, facilitating improved generalization to out-of-distribution samples.

In the multi-source domain generalization (MSDG) setting, where there are multiple source domains for training, domain label information can be leveraged to learn the domain shift~\cite{muandet2013domain, d2019domain, xiao2021bit}.
Prior information about the target domain is also useful to design specific data augmentation methods to tackle domain shift.
For instance, if it is known that the target domain contains sketches, skeletonizing the source images is a good solution~\cite{geirhos2018imagenettrained}; if it is known that the target domain contains geometric transformations, rotation/translation/scaling would be a suitable augmentation~\cite{jaderberg2015spatial}; or if attributes of the target domain are known they can be used for learning data augmentations~\cite{gokhale2021attribute}.
However, these methods assume that we know the properties of the target domain -- such knowledge is not available in the single-source domain generalization (SSDG) setting.
In the SSDG setting, where only one domain is available for training, it is more challenging to address the domain shift issue.
In this paper, we focus on the strict SSDG setting, where only one source domain is available for training and no prior knowledge is available about the target domain.

\begin{figure}
    \centering
    \includegraphics[width=\linewidth]{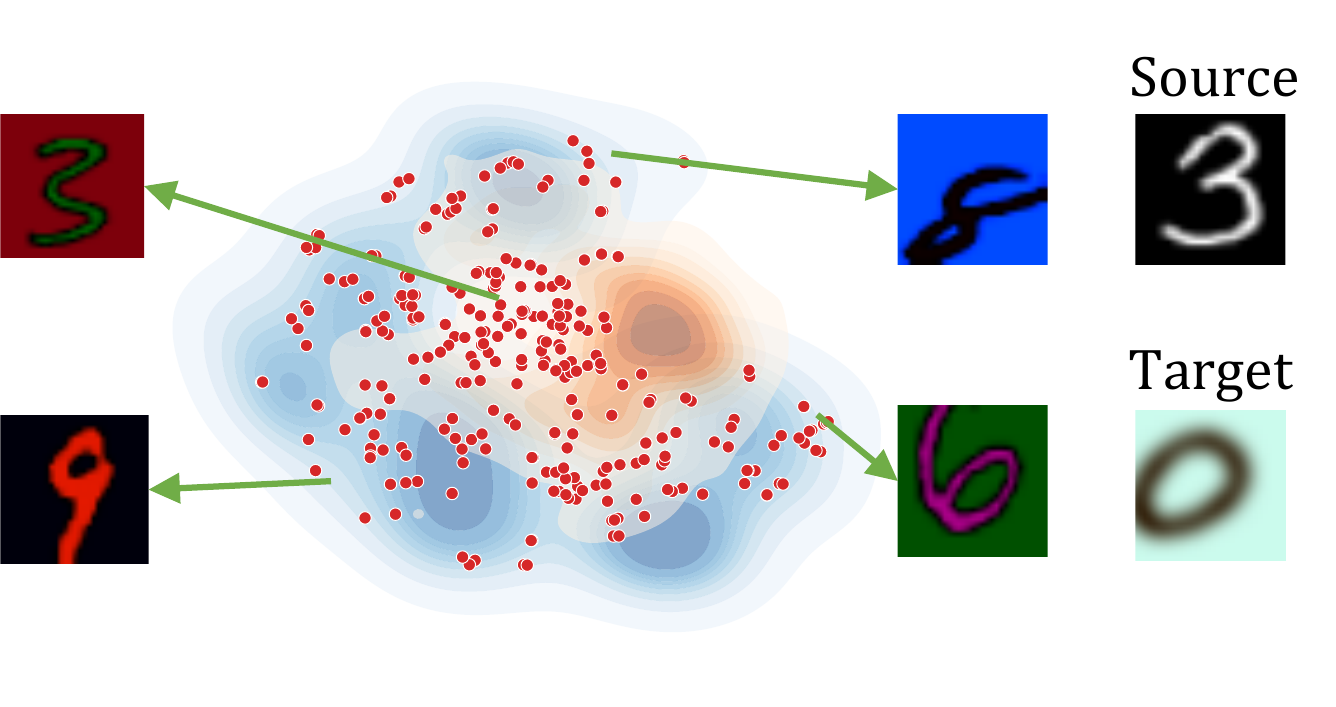}
    \caption{An illustration of the diversity introduced by Adversial Bayesian Augmentations.
    The blue and orange surfaces represent the source (seen) and target (unseen) domains respectively.
    The red dots represent the samples augmented by ABA; these augmentations expose the classifier to regions closer to the target domain, thereby improving image classifiers' generalization to unseen domains.
    }
    \label{fig:teaser}
\end{figure}

\begin{figure*}
    \centering
             \includegraphics[width=\linewidth]{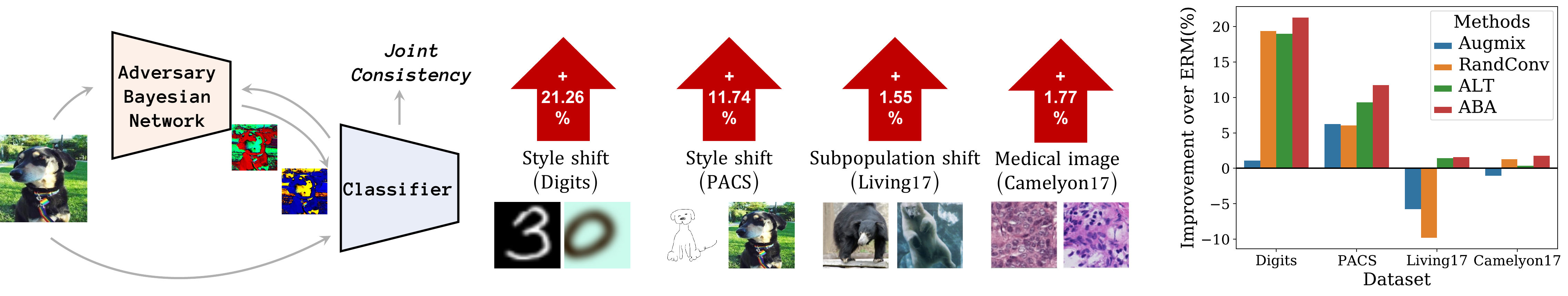}
    
    \caption{The Adversarial Bayesian Augmentation framework (left panel), the improvement of our method ($\text{ABA}_{\text{5-layers}}$) over ERM on each dataset, where samples in source and target domain are displayed under the name of the dataset (middle panel), and summarization of results on a wide range of domain generalization benchmarks (right panel).
    }
    \label{fig:overview}
\end{figure*}

Recent work in SSDG focuses on augmenting the data in order to simulate the presence of out-of-distribution domains.
One way involves learning-free data augmentation methods, such as RandConv~\cite{xu2021robust}, Augmix~\cite{hendrycks*2020augmix} and JiGen~\cite{carlucci2019domain} -- here the data augmentation is pre-specified and does not evolve or adapt during training.
Another approach is based on adversarial perturbations, which involves generating adversarial samples to improve generalization, such as Augmax~\cite{wang2021augmax}, ADA~\cite{volpi2018generalizing}, M-ADA~\cite{qiao2020learning}, and ALT~\cite{gokhale2023improving}.
Although the Bayesian neural networks as the backbone of the classifier show good generalization ability to out-of-distribution samples intrinsically~\cite{louizos2015variational, xiao2021bit, daxberger2019bayesian},
and some papers~\cite{saatci2017bayesian} use Bayesian neural networks for generating images, none of the work directly augments the data by Bayesian neural network for domain generalization.

In this paper, we present a novel approach called Adversarial Bayesian Augmentation, dubbed \textit{ABA}, which draws on the strengths of adversarial learning and Bayesian neural networks to generate more diverse data and improve generalization on different domains, as shown in Figure~\ref{fig:overview}. 
Specifically, the adversarial learning-based methods, which explore a wider augmentation space, already outperforms learning-free methods~\cite{gokhale2023improving, wang2021augmax} on SSDG.
The introduction of weight uncertainty by the Bayesian neural network further enhances the strength of data augmentation, as shown in Figure~\ref{fig:teaser}. 
Our experimental results demonstrate the superior performance of ABA compared to existing methods.

\noindent The key contributions and findings of the paper thus are:
\begin{itemize}[nosep,noitemsep,leftmargin=*]
    \item We introduce a novel data augmentation method, dubbed ABA, which combines adversarial learning and Bayesian neural network, to improve the diversity of training data for single-source domain generalization setting.
    \item We empirically validate the effectiveness of our proposed method on four datasets, covering three types of domain generalization, namely style generalization, subpopulation generalization, and medical imaging generalization. Our method outperforms all existing state-of-art methods on all four datasets.
    \item We investigate the core driving forces from ABA which enable the generation of diverse data and conduct comprehensive ablation study on how the model hyperparameters impact the overall system's performance. 
\end{itemize}

\section{Related Work}
\noindent\textbf{Domain Generalization.}
Domain generalization aims to learn representations that can be transferred to unseen domains.  
Recent research has explored a range of techniques, including feature fusion~\cite{shen2019situational}, meta learning~\cite{li2018learning}, adversarial learning~\cite{volpi2018generalizing, qiao2020learning}, style transformation~\cite{nam2021reducing}. 
Typically, there are two settings of domain generalization: single-source domain generalization (SSDG), where only one source domain is available for training, and multi-source domain generalization (MSDG), where multiple source domains are provided.
However, in practice, data from multiple sources can be expensive or sometimes infeasible. 
In this paper, we focus on SSDG setting.

\medskip\noindent\textbf{Single-Source Domain Generalization.}
Since there is no extra information available about the target domain, most methods for SSDG focus on data augmentation to generate more diverse samples. 
For example, JiGen~\cite{carlucci2019domain} decomposes the image into grids and randomly shuffles the patches to create the augmented image.
RandConv~\cite{xu2021robust} uses random convolutions as data augmentation, preserving the shape and local texture information.
ADA~\cite{volpi2018generalizing} and M-ADA~\cite{qiao2020learning} adversarially augment image at the pixel level.
SagNet~\cite{nam2021reducing} transfers the style of the image.
Augmix~\cite{hendrycks*2020augmix} and Augmax~\cite{wang2021augmax} compose data augmentation operations with random or learned mixing coefficients.

\medskip\noindent\textbf{Bayesian Neural Network.}
In a standard deep neural network, the weight of the network takes the single values learned from data. 
In contrast, the Bayesian neural network (BNN) aims to estimate the distribution of weights, which provides the capability of uncertainty estimation~\cite{gal2016dropout}, robustness to over-fitting~\cite{gal2015bayesian}, and resistance to adversarial attacks~\cite{pang2021evaluating}.
However, the estimation of the posterior of the weights is often intractable. 
Current research has focused on techniques such as Bayes By Backprop (BBB)~\cite{blundell2015weight} with local reparameterization~\cite{kingma2015variational, molchanov2017variational}, Variational Inference (VI)~\cite{kingma2013auto}, and Flipout approximation~\cite{wen2018flipout}.
Variational Bayesian inference, coupled with domain invariance learning~\cite{xiao2021bit} can improve domain generalization. However, this method is only adapted to MSDG settings.

\medskip\noindent\textbf{Adversarial Training.}
Adversarial training has been proposed as a solution to mitigate against the vulnerability of neural networks to input perturbations~\cite{szegedy2013intriguing,goodfellow2014explaining, madry2018towards}. 
AdvBNN~\cite{liu2018advbnn} proposed an adversarial-trained Bayesian neural network that is robust to strong adversarial attacks, which we adopt with a similar formulation of the min-max problem.
Recent work~\cite{stutz2019disentangling} has shown that on-manifold adversarial samples can improve both robustness and generalization.
Adversarial training has also been adopted for domain generalization.
ADA~\cite{volpi2018generalizing} and M-ADA~\cite{qiao2020learning} employ adversarial data augmentation at the pixel level to generate difficult examples for improving domain generalization.
ESDA~\cite{volpi2019addressing} adversarially learns image transformation, while Augmax~\cite{wang2021augmax} learns the combination of image augmentation operations in an adversarial manner.
ALT~\cite{gokhale2023improving} goes further and builds an additional image-to-image transformation network to learn adversarial augmentations.
Recent work shows trade-offs between reliability metrics such as accuracy, robustness, and fairness \cite{zhang2019theoretically,gokhale2022generalized,moayeri2022explicit,tran2022fairness}.

\section{Proposed Method}
Let $\mathcal{S}$ and $\mathcal{T}$ represent the source and target domains respectively, which share the same label space. 
The training set is a subset in the source domain and contains $N$ training pairs, denoted as $\{(x_i, y_i)\}_{i=1}^N \subset \mathcal{S}$. 
The objective of SSDG is to use $\mathcal{S}$ to learn parameters $\theta$ of a classifier $f$ which also can generalize well to target domain $\mathcal{T}$.

To accomplish this, since no information is available from the target domain $\mathcal{T}$, previous works focus on data augmentation, denoted as $g$. 
For example, in RandConv~\cite{xu2021robust}, $g$ is a random convolutional layer, while in Augmix~\cite{hendrycks*2020augmix}, $g$ is a composition of image augmentation operations. 
To learn the representation invariant to data augmentation, a consistency regularization loss is typically used to encourage consistent prediction between the clean image and the augmented image. 
The Kullback-Leibler (KL) divergence loss is commonly used for consistency loss.

\subsection{Adversarial Bayesian Augmentation} \label{sec:ABA}

In this paper, we design $g$ as a $L$-layer Bayesian convolutional neural network, parameterized by $\Phi ~{=}~ \{ \phi_l \}_{l=1}^L$, where $\phi_i ~{\in}~ \mathbb{R}^{k_l\times k_l\times C_{in(l)} \times C_{out(l)}}$ are the parameters of each Bayesian convolutional layer. 
Following the setting in~\cite{xu2021robust}, we randomly sample $k_l$ from $\mathcal{K} ~{=}~ \{1,3, ...n\}$. 
$C_{in(l)}$ and $C_{out(l)}$ represent the number of input and output channels for each layer convolutional kernel.
Since $g$ is an image augmentation function, the number of input channels for the first and last layer are equal to the number of image channels (3 for color images and 1 for grayscale images).

To perform Bayesian inference, we need to estimate the posterior distribution $p(\phi_l|x, y)$, which is intractable in closed form. 
To approximate it, we adopt the variational Bayesian inference approach and use a variational distribution $q(\phi_l)$.
This distribution is obtained by minimizing the KL divergence between $q(\phi_l)$ and true posterior distribution $p(\phi_l|x, y)$.
To enable efficient sampling of the variational distribution, we re-parameterize as $\phi_l=\mu_l + \sigma_l \epsilon_l$, where $\epsilon_l$ is sampled from the standard normal distribution, which allows us to compute the gradients of $\mu_l$ and $\sigma_l$.
We denote $\boldsymbol{\mu} = \{\mu_l\}_{l=1}^L$ and $\boldsymbol{\sigma} = \{\sigma_l\}_{l=1}^L$. So $\Phi=\{\boldsymbol{\mu}, \boldsymbol{\sigma}\}$.

\begin{algorithm}[t]
\small
\caption{Learning with Adversarial Bayesian Augmentation (ABA)}\label{alg:algo}
\KwIn{$\{x_i, y_i\}_{i=1}^{N}$}
\KwOut{Classifier $f$ parameters $\theta^*$}
    \For{$t \leftarrow 1$ \KwTo $T$}{
        \eIf{$t < T_{\text{warmup}} $}{
            $\theta \leftarrow \theta - \gamma \bigtriangledown \mathcal{L}_{\text{cls}}$
         }{
         \tcc{Training ABA}
            $\Phi \leftarrow \Phi_0$ \\
            \For{$m \leftarrow 1$ \KwTo $T_{adv}$}{
                $y_g = f(g(x, \Phi), \theta)$ \leavevmode \\
                $\Phi \leftarrow \Phi -  \eta \bigtriangledown \mathcal{L}_{\text{ELBO}}$ \tcp{See \eqref{eq:elbo}}
            }
            \tcc{Train classifier}
            $\Phi \leftarrow \boldsymbol{\mu} + \boldsymbol{\sigma} \odot \epsilon$ \tcp{Sample parameters} \leavevmode 
            $\theta \leftarrow \theta - \gamma \bigtriangledown (\mathcal{L}_{\text{cls}} + \alpha \mathcal{L}_{\text{KL}})$ \tcp{See \eqref{eq:cls},\eqref{eq:kl}}
         }
    }
\textbf{Return $\theta$} 
\end{algorithm}

The optimization of ABA is formulated as a min-max problem.
Initially, we optimize the $g$ network using adversarial training to augment images that can fool the classifier $f$. 
To achieve this, we use the evidence lower bound (ELBO) of the variational Bayesian network as the loss function. ELBO is a lower bound on the log marginal likelihood of the observed data and is defined as follows:
\begin{align} \label{eq:elbo}
    \mathcal{L}_{\mathrm{ELBO}} = &\frac{1}{N}\sum_{i=1}^N\mathbb{E}_{g\sim q(\Phi)}[\log(y_i|g(x_i), \theta) ]  \nonumber \\ 
    &-\beta \sum_{l=1}^L\mathrm{KL}(q(\phi_l)||p(\phi_l)),
\end{align}
where the prior distribution $p(\phi_l)$ of each layer follows $\mathcal{N}(0, \frac{1}{k_l\times k_l \times C_{in(l)}})$, which is commonly used in network initialization~\cite{he2016deep}.
Theoretically, the coefficient $\beta$ for the KL term should be 1. 
However, in practice, for small datasets or large models, smaller $\beta$ ($0 < \beta < 1$) is preferred~\cite{liu2018advbnn}.

Starting from a random initialization, the parameters of $g$ are iteratively updated by maximizing the negative of ELBO.
In contrast to adv-BNN~\cite{liu2018advbnn}, which constrains the adversarial samples bounded by $\ell_p$ norm, we control the strength of adversarial samples by adjusting the learning rate $\eta$ and the number of adversarial steps $T_{adv}$.
The final augmented images $x_g$ are obtained through Bayesian inference using the optimized parameters $\Phi^*$ and clamped to the image range. 
Note that we can sample multiple augmented images from Bayesian inference, and we sample twice denoted as $x_{g_1}$ and $x_{g_2}$.
These augmented images can be used for classifier learning in the presence of domain shift.

Next, we optimize the classifier $f$ with a loss function consisting of two terms: a cross-entropy loss, which is 
\begin{equation} \label{eq:cls}
    \mathcal{L}_{\mathrm{cls}} = \mathrm{CrossEntropy}(f(x_{g_1}, \theta), y),
\end{equation}
 and a consistency regularization loss, which helps to keep the prediction consistent on augmented data, defined as:
\begin{equation} \label{eq:kl}
    \mathcal{L}_{\mathrm{KL}} = \mathrm{KL}(p_c||\Bar{p}) + \mathrm{KL}(p_{g_1}||\Bar{p}) + \mathrm{KL}(p_{g_2}||\Bar{p}),
\end{equation}
where $p_c, p_{g_1}, p_{g_2}$ denotes the softmax prediction of $f$ on clean image $x$ and augmented images $x_{g_1}, x_{g_2}$ respectively. 
$\Bar{p}$ is the average of $p_c, p_{g_1},$ and $p_{g_2}$.

\begin{table*}[th]
    \centering
    \small
    \begin{tabular}{lcccccc}
    \toprule
    \textbf{Method} & \textbf{MNIST-10K} & \textbf{MNIST-M} & \textbf{SVHN} & \textbf{USPS} & \textbf{SYNTH} & \textbf{Target Avg.} \\
    \midrule
    ERM & 98.40 \scriptsize(0.84) & 58.87 \scriptsize(3.73) & 33.41 \scriptsize(5.28) & 79.27 \scriptsize(2.70) & 42.43 \scriptsize(5.46) & 53.50 \scriptsize(4.23)   \\
    ADA & N/A & 60.41 & 35.51 & 77.26 & 45.32 & 54.62 \\
    M-ADA& 99.30 & 67.94 & 42.55 & 78.53 & 48.95 & 59.49\\
    ESDA & 99.30 \scriptsize(0.10) & 81.60 \scriptsize(1.60) & 48.90 \scriptsize(5.20) & 84.00 \scriptsize(1.20) & 62.20 \scriptsize(1.30) & 69.12 \scriptsize(2.33)\\
    AdvBNN & 98.23 \scriptsize(0.08) & 71.79 \scriptsize(0.69) & 44.85 \scriptsize(0.55) & 46.05 \scriptsize(0.53) & 44.99 \scriptsize(0.54) & 51.92 \scriptsize(0.51)\\
    Augmix & 98.53 \scriptsize(0.18) & 53.36 \scriptsize(1.59) & 25.96 \scriptsize(0.80) & 96.12 \scriptsize(0.72) & 42.90 \scriptsize(0.60) & 54.59 \scriptsize(0.50) \\
    \hline
     {\color{purple} \scriptsize 1-layer convolutional-based augmentations} \\
    RandConv & 98.85 \scriptsize(0.04) & \textbf{87.76 \scriptsize(0.83)} & \textbf{57.62 \scriptsize(2.09)} & 83.36 \scriptsize(0.96) & 62.88 \scriptsize(0.78) & 72.88 \scriptsize(0.58)\\
    $\text{ALT}_{\text{1-layer}}$ & 98.41 \scriptsize(0.15) & 72.80 \scriptsize(2.06) & 47.07 \scriptsize (1.88) & 94.79 \scriptsize(0.88) & 66.27 \scriptsize(1.56) & 70.23 \scriptsize(1.22)\\
    $\text{ALT}_{\text{1-layer+RandConv}}$ & 98.54 \scriptsize(0.10) & 75.77 \scriptsize(1.51) & 49.90 \scriptsize(1.62) & 95.64 \scriptsize(0.62) & 68.61 \scriptsize(1.75) & 72.47 \scriptsize(1.18)\\
    \rowcolor{LightCyan1}
        $\text{ABA}_{\text{1-layer}}$ & 98.82 \scriptsize(0.09) & 78.81 \scriptsize(1.64) & 51.88 \scriptsize(1.93) & \textbf{96.22 \scriptsize(0.26)} &\textbf{71.25 \scriptsize(1.27)} & \textbf{74.57 \scriptsize(0.52)}\\
    \rowcolor{LightCyan1} 
        $\text{ABA}_{\text{1-layer+RandConv}}$ & 98.78 \scriptsize(0.09) & 78.62 \scriptsize(0.92) & 52.04 \scriptsize(1.13) & 96.16 \scriptsize(0.16) & 71.23 \scriptsize(0.93) & 74.51 \scriptsize(0.70)\\
    \hline
    \color{purple} \scriptsize 3-layer convolutional-based augmentations \\
    \rowcolor{LightCyan1}
        $\text{ABA}_{\text{3-layers}}$ & 98.73 \scriptsize(0.10) & \textbf{80.94 \scriptsize(0.39)} & 55.88 \scriptsize(0.70) & 96.34 \scriptsize(0.54) & 73.09 \scriptsize(0.34) & 76.56 \scriptsize(0.06)\\
    \rowcolor{LightCyan1}
        $\text{ABA}_{\text{3-layers+RandConv}}$ & 98.67 \scriptsize(0.11) & 80.05 \scriptsize(0.81) & \textbf{56.87 \scriptsize(1.05)} & \textbf{96.55 \scriptsize(0.34)} & \textbf{73.40 \scriptsize(0.19)} & \textbf{76.72 \scriptsize(0.41)} \\
    \hline 
    \color{purple} \scriptsize 5-layer convolutional-based augmentations \\
    $\text{ALT}_{\text{5-layer}}$ & 98.46 \scriptsize(0.27) & 74.28 \scriptsize(1.36) & 52.25 \scriptsize(1.54) & 94.99 \scriptsize(0.68) & 68.44 \scriptsize(0.98) & 72.49 \scriptsize(0.87)\\
    $\text{ALT}_{\text{5-layer+RandConv}}$ & 98.46 \scriptsize(0.25) & 76.90 \scriptsize(1.42) & 53.78 \scriptsize(1.97) & 95.40 \scriptsize(0.72) & 69.40 \scriptsize(1.07) & 73.87 \scriptsize(1.03) \\
    \rowcolor{Snow2}
    $\text{ALT}_{\text{5-layer+Augmix}}$ & 98.55 \scriptsize(0.11) & 75.98 \scriptsize(0.89) & 55.01 \scriptsize(1.34) & 96.17 \scriptsize(0.45) & 68.93 \scriptsize(2.17) & 74.38 \scriptsize(0.86) \\
    \rowcolor{LightCyan1}
    $\text{ABA}_{\text{5-layer}}$ & 98.78 \scriptsize(0.06) & \textbf{80.54 \scriptsize(0.53)} & 52.45 \scriptsize(1.21) & 95.81 \scriptsize(0.47) & 70.25 \scriptsize(1.21) & 74.76 \scriptsize(0.52)\\
    \rowcolor{LightCyan1}
    $\text{ABA}_{\text{5-layer+RandConv}}$ & 98.76 \scriptsize(0.12) & 79.69 \scriptsize(0.35) & 54.09 \scriptsize(1.27) & \textbf{96.42 \scriptsize(0.35)} & \textbf{71.55 \scriptsize(0.96)} & 75.44 \scriptsize(0.61)\\
    \rowcolor{LightCyan1}
    $\text{ABA}_{\text{5-layer+Augmix}}$ & 98.66 \scriptsize(0.16) & 80.24 \scriptsize(0.51) & \textbf{56.43 \scriptsize(0.59)} & 96.14 \scriptsize(0.64) & 70.91 \scriptsize(0.83) & \textbf{75.93 \scriptsize(0.60)}\\
    \bottomrule
    \end{tabular}
    \caption{\textbf{SSDG accuracy on Digits dataset.} The source domain is MNIST-10K. The target domains are MNIST-M, SVHN, USPS, SYNTH. We report the mean (and standard deviation) of 5 runs. 
}
    \label{tab:digits}
\end{table*}

\medskip\noindent\textbf{Comparison with other convolutional-based augmentations.}
For only one layer network, ABA adversarially learns the distribution of parameters, while in RandConv~\cite{xu2021robust}, the parameters are sampled from a fixed distribution, namely $\mathcal{N}(0, \frac{1}{C_{in}\times H \times W})$, where $H$ and $W$ are the size of the convolutional kernel, $C_{in}$ is the input channel. 
For multiple layers network, compared with ALT~\cite{gokhale2023improving}, ABA learns a \textit{Bayesian} convolutional neural network adversarially rather than a standard convolutional neural network.  

\medskip\noindent\textbf{Implementation.}
Algorithm~\ref{alg:algo} depicts the implementation details.
For network design,  the activation of multiple layers ABA is \texttt{LeakyRelu}.
The second augmented image $x_{g_2}$ can be obtained not only through Bayesian inference, but also obtained from other data augmentation techniques, such as RandConv~\cite{xu2021robust}, Augmix~\cite{hendrycks*2020augmix}.
We train the classifier for a total of $T$ iterations. 
At first $T_{warmup}$ iterations, we train the classifier without any data augmentation methods. 
After $T_{warmup}$, for each iteration, we randomly initialize the $g$ and update its parameters by adversarial Bayesian training. 
The learning rate of adversarial learning is $\eta$.
After $T_{adv}$ steps adversarial learning, we sample the augmented images via Bayesian inference and clamp them to the image range.
We then use the augmented images, along with the clean image, to train the classifier $f$ using the classification loss and consistency regularization. 
The learning rate of the classifier is $\gamma$ and the weight of consistency regularization is $\alpha$.

\section{Experiments}
In this section, we validate our method on four datasets that represent three types of generalization: style shift, 
subpopulation shift, 
and domain shift in medical imaging.
We train our models on the source domain and evaluate them on the test set of each target domain.

We compare our approach against several state-of-the-art methods
\footnote{In \cref{tab:digits,tab:pacs,tab:wilds} we highlight the previous best model in \hlgray{gray}, variants of ABA better than the previous best in \hlcyan{blue}, and the best accuracy in \textbf{bold}.}
using seven variants.
For fair comparison with RandConv~\cite{xu2021robust}, we use $\text{ABA}_{\text{1-layer}}$, \ie ABA with a 1-layer Bayesian neural network. 
To match the number of convolutional layers in ALT~\cite{gokhale2023improving}, we use $\text{ABA}_{\text{5-layer}}$, \ie ABA with a 5-layer Bayesian convolutional neural network.
In the variants $\text{ABA}_{\text{5-layer+RandConv}}$ and $\text{ABA}_{\text{5-layer+Augmix}}$, the second augmented image is generated by RandConv or Augmix instead of Bayesian neural network.

\subsection{Style Shift}
We validate our method on two popular style-shift benchmark datasets:
(1) \textbf{Digits}
is composed of digit images from MNIST-10K~\cite{lecun1998gradient}, MNIST-M~\cite{ganin2016domain}, SVHN~\cite{Netzer2011}, USPS~\cite{denker1988neural}, SYNTH~\cite{ganin2015unsupervised}.
Following the setting in ~\cite{volpi2018generalizing}, MNIST-10K is the source domain containing 10,000 images from MNIST, and the other four datasets are target domains.
(2) \textbf{PACS}~\cite{li2017deeper}
consists of images from four domains: photo, art painting, sketch, and cartoon, and 7 classes. 
We select one domain as the source domain and the other three as the target domains.

\subsubsection{Digits}
\noindent\textbf{Baselines.}
Our baselines include Empirical Risk Minimization (ERM), ADA~\cite{volpi2018generalizing}, M-ADA~\cite{qiao2020learning}, AdvBNN~\cite{liu2018advbnn}, 
ESDA~\cite{volpi2019addressing}, RandConv~\cite{xu2021robust}, Augmix\cite{hendrycks*2020augmix} and ALT~\cite{gokhale2023improving}.
For fair comparison with RandConv and ALT, we implement the 1-layer and 5-layer ABA.
The classifier architecture for $f$ is DigitNet~\cite{volpi2018generalizing}, with $T~{=}~10000$ iterations, batch size 512, learning rate $\gamma ~{=}~ 1e^{-4}$ and Adam optimizer.
For ABA $g$, we set the weight of consistency loss term $\alpha ~{=}~ 3$, adversarial learning rate $\eta~{=}~5e^{-6}$, number of adversarial steps $T_{adv} ~{=}~ 10$, warm-up steps $T_{warmup} ~{=}~ 5$, and factor for KL term of ELBO $\beta~{=}~1$.

\medskip\noindent\textbf{Results.}
Table~\ref{tab:digits} shows 
that pixel-level adversarial perturbation methods such as ADA and M-ADA, and the composition of image augmentation method like Augmix, only marginally improve SSDG performance, while AdvBNN even downgrades the performance. 
However, convolutional-based augmentations, even with just one layer, can significantly enhance performance.
Among the 1-layer convolutional augmentations, the weights learned adversarially (ALT) do not perform better than the weights randomly sampled from a fixed distribution (RandConv).
However, our 1-layer ABA outperforms both, achieving a 1.69\% improvement over RandConv and a 4.34\% improvement over 1-layer ALT.
Combining a RandConv module does not improve performance much for 1-layer ABA, but still outperforms 1-layer ALT with RandConv by 2.04\%.
A 5-layer ABA can further improve performance by a small margin 0.19\%, similar to the observation of ALT. 
Adding a RandConv module improves performance by a small margin of 0.68\% compared to 5-layer ABA and by 1.57\% compared to 5-layer ALT with RandConv.
Adding an Augmix module improves by 1.17\% compared to 5-layer ABA and by 1.55\% compared to 5-layer ALT with Augmix.
We achieve state-of-art results by 3-layer ABA with RandConv, with an accuracy of 76.72\%.
We note that RandConv outperforms all state-of-the-art methods in some domains, such as MNIST-M and SVHN, but does not consistently achieve superior results in other domains.

\begin{table*}[th]
    \centering
    \small
    \begin{tabular}{lccccc}
    \toprule
    \textbf{Method} & \textbf{Photo} & \textbf{Cartoon} & \textbf{Art} & \textbf{Sketch} & \textbf{Avg.} \\
    \midrule
    ERM & 38.93 & 70.00 & 68.83 & 39.36 & 54.28\\
    JiGen & 41.70 & 72.23 & 67.70 & 36.83 & 54.61\\
    SagNet & 48.53 & 75.66 & 73.20 & 50.06 & 61.86\\
    ADA & 44.63 & 71.96 & 72.43 & 45.73 & 58.68\\
    AdvBNN & 45.93 \scriptsize(0.41) & 60.24 \scriptsize(0.95) & 75.33 \scriptsize(0.95) & 26.19 \scriptsize(1.23) & 51.92 \scriptsize(1.15) \\
    Augmix & 45.24 \scriptsize(1.12) & 74.66 \scriptsize(1.09) & 71.47 \scriptsize(0.64) & 47.72 \scriptsize(1.72) & 60.51 \scriptsize(1.14)\\
    \hline
    \color{purple} \scriptsize 1-layer convolutional-based augmentations \\
    RandConv &49.80 \scriptsize(4.23) &67.90 \scriptsize(1.55)& 69.63 \scriptsize(2.15)& \textbf{54.06 \scriptsize(1.96)} & 60.34 \scriptsize(2.47)\\   
    $\text{ALT}_{\text{1-layer}}$ & 50.83 \scriptsize(2.13) & 75.00 \scriptsize(0.62)& 73.87 \scriptsize(1.31) & 47.83 \scriptsize(1.95) & 61.88 \scriptsize(1.50)\\
    $\text{ALT}_{\text{1-layer+RandConv}}$ & 52.24 \scriptsize(0.82)  & 75.16 \scriptsize(0.67) & 73.46 \scriptsize(1.29)& 49.21 \scriptsize(2.14)& 62.51 \scriptsize(1.23)\\

        $\text{ABA}_{\text{1-layer}}$ &\textbf{54.49 \scriptsize(1.35)}& 75.61 \scriptsize(0.89) & 75.59 \scriptsize(1.56)& 52.84 \scriptsize(2.80) & \textbf{64.63 \scriptsize(1.65)}\\
    $\text{ABA}_{\text{1-layer+RandConv}}$ & 52.32 \scriptsize(1.82) & \textbf{76.01 \scriptsize(0.56)} & \textbf{75.77 \scriptsize(1.64)} & 50.20 \scriptsize(1.93) & 63.58 \scriptsize(1.49)\\
    \hline
    \color{purple} \scriptsize 3-layer convolutional-based augmentations \\
    \rowcolor{LightCyan1}
        $\text{ABA}_{\text{3-layers}}$ & \textbf{58.86 \scriptsize(0.83)} & \textbf{77.49 \scriptsize(0.57)}& 75.34 \scriptsize(0.89)& \textbf{53.76 \scriptsize(2.46)}& \textbf{66.36 \scriptsize(1.19)}\\ 
    \rowcolor{LightCyan1}
        $\text{ABA}_{\text{3-layers+RandConv}}$ & 56.95 \scriptsize(0.80) & 77.21 \scriptsize(0.85) & \textbf{75.34 \scriptsize(0.52)} & 53.52 \scriptsize(0.90)  & 65.76 \scriptsize(0.15)\\
    \hline 
    \color{purple} \scriptsize 5-layer convolutional-based augmentations \\
    $\text{ALT}_{\text{5-layer}}$ & 54.33 \scriptsize(1.08) &  75.96 \scriptsize(1.12) & 74.06 \scriptsize(1.09) & 50.03 \scriptsize(2.41) & 63.60 \scriptsize(1.43)\\ 
    $\text{ALT}_{\text{5-layer+RandConv}}$ & 55.66 \scriptsize(0.50) & 76.23 \scriptsize(0.80) & 73.96 \scriptsize(0.54) & 50.86 \scriptsize(0.79)& 64.18 \scriptsize(0.66)\\
    \rowcolor{Snow2}
        $\text{ALT}_{\text{5-layer+Augmix}}$ & 55.09 \scriptsize(1.87) & \textbf{77.36 \scriptsize(0.73)} & \textbf{75.69 \scriptsize(1.21)} & 50.72 \scriptsize(1.41) & 64.72 \scriptsize(1.30)\\
    \rowcolor{LightCyan1}
        $\text{ABA}_{\text{5-layer}}$ & \textbf{59.04 \scriptsize(1.43)} & 77.16 \scriptsize(0.35) & 74.71 \scriptsize(0.76)& 53.18 \scriptsize(2.07)& \textbf{66.02 \scriptsize(1.15)}\\
    \rowcolor{LightCyan1}
        $\text{ABA}_{\text{5-layer+RandConv}}$ & 57.59 \scriptsize(1.26) & 76.66 \scriptsize(0.24)& 75.61 \scriptsize(1.02) & \textbf{54.12 \scriptsize(1.33)} & 66.00 \scriptsize(0.96)\\
    \rowcolor{LightCyan1}
        $\text{ABA}_{\text{5-layer+Augmix}}$ & 57.87 \scriptsize(0.22) & 77.29 \scriptsize(0.78)& 74.70 \scriptsize(0.96) & 52.35 \scriptsize(0.03) & 65.55 \scriptsize(0.49)\\
        
    \bottomrule
    \end{tabular}
    \caption{\textbf{SSDG accuracy on PACS.} Each column is the average accuracy on the target domains trained on the given source domain. We report the mean (and standard deviation) of 5 runs. More details about the accuracy of the source domain to each target domain are in the Appendix.}
    \label{tab:pacs}
\end{table*}
 
\subsubsection{PACS}
\noindent\textbf{Baselines.}
Our baseline methods include the Empirical Risk Minimization (ERM), JiGEn~\cite{carlucci2019domain},
ADA~\cite{volpi2018generalizing}, AdvBNN~\cite{liu2018advbnn}, Augmix~\cite{hendrycks*2020augmix}, Randconv~\cite{xu2021robust}, SagNet~\cite{nam2021reducing},
and ALT~\cite{gokhale2023improving}.
To train the model, we use the pre-trained ResNet18 model and set the training iterations $T=2000$, with batch size 32, learning rate $\gamma=4e^{-4}$, SGD optimizer with cosine annealing learning rate scheduler. 
For AdvBNN, due to the computing complexity, we only bayesianize the linear layers of the networks.
For ABA, we set the weight of consistency loss term $\alpha = 3$, adversarial learning rate $\eta=5e^{-5}$, number of adversarial steps $T_{adv} = 10$, the warm-up steps $T_{warmup} = 4$ and factor for KL term of ELBO $\beta=0.1$.

\medskip\noindent\textbf{Results.}
Our experiments on the PACS dataset are summarized in Table~\ref{tab:pacs}.
For PACS, each domain can be considered as a source domain, while the remaining three domains serve as target domains.
we report the average accuracy across all target domains for a given source domain in each column, as well as the average accuracy across all four source domains.
As PACS contains images with different styles, methods such as SagNet and RandConv that transfer style and preserve shape and texture information can improve performance.
In comparison, JiGen and ADA only marginally improve accuracy, while AdvBNN downgrades the performance.
Similarly to the Digits dataset, leveraging convolutional-based augmentations provides significant performance improvements, with five variants of ALT performing better than other baseline models.
However, our proposed ABA method outperforms ALT, with 1-layer ABA achieving a 2.75\% improvement over 1-layer ALT and 5-layer ABA achieving a 2.42\% improvement over 5-layer ALT.
Adding RandConv modules further improves performance, with $\text{ABA}_{\text{1-layer+RandConv}}$ and $\text{ABA}_{\text{5-layer+RandConv}}$ achieving improvements of 1.07\% and 1.82\% over $\text{ALT}_{\text{1-layer+RandConv}}$ and $\text{ALT}_{\text{5-layer+RandConv}}$, respectively.
Adding Augmix modules on 5-layer ABA still performs better than 5-layer ALT with Augmix by 0.83\%. 
We achieve the state-of-art results by 3-layer ABA, with an accuracy of 66.36\%.
We observe that RandConv performs well in the Sketch domain, which is consistent with the intuition that it preserves the shape and texture information.

\subsection{Subpopulation Shift}
We validate our method on subpopulation shift with the Living17 dataset~\cite{santurkar2021breeds}.
Living17 contains images from ImageNet~\cite{deng2009imagenet} from 17 superclasses, each of which contains 4 subclasses based on WordNet hierarchy~\cite{Fellbaum2005-FELWAW}.
For example, \textit{labrador} and \textit{husky} are subclasses of the superclass \textit{dog}.
We choose 2 subclasses in each superclass as the source domain and the rest of the subclasses as the target domain, following the setting in~\cite{santurkar2021breeds}. See the appendix for details.

\begin{table*}[ht]
    \centering
    \small
    \begin{tabular}{c@{\hskip 0.3in}c}
        \centering
        \begin{tabular}{lcc@{}}
        \toprule
        \textbf{Method} & \textbf{Source} & \textbf{Target}\\
        \midrule
             ERM & 95.84 \scriptsize(0.13) & 70.97 \scriptsize(0.80) \\
             AdvBNN & 92.59 \scriptsize(0.47) & 60.76 \scriptsize(0.27)\\   
             Augmix & 94.58 \scriptsize(0.17) & 65.22 \scriptsize(0.90) \\
             Augmax & 94.00 \scriptsize(0.30) & 63.75 \scriptsize(0.50) \\
             RandConv & 87.23 \scriptsize(1.54) & 61.18 \scriptsize(1.65) \\
             $\text{ALT}_{\text{5-layer}}$ & 94.98 \scriptsize(0.12) & 72.38 \scriptsize(0.84)\\
             \rowcolor{Snow2}
             $\text{ALT}_{\text{5-layer+RandConv}}$ & 94.91 \scriptsize(0.15) & 72.47 \scriptsize(0.63)\\
             \rowcolor{white}
             $\text{ALT}_{\text{5-layer+Augmix}}$ & 95.03 \scriptsize(0.09) & 71.97 \scriptsize(0.62)\\
             \rowcolor{LightCyan1}
             $\text{ABA}_{\text{5-layer}}$ & 95.18 \scriptsize(0.31) & 72.52 \scriptsize(0.55) \\
             $\text{ABA}_{\text{5-layer+RandConv}}$ & 95.30 \scriptsize(0.24) &\textbf{72.69 \scriptsize(1.16)} \\
             \rowcolor{white}
             $\text{ABA}_{\text{5-layer+Augmix}}$ & 95.32 \scriptsize(0.12) & 72.41 \scriptsize(0.38) \\
             \bottomrule
        \end{tabular}
        \quad
        \begin{tabular}{lccccc@{}}
        \toprule
        \textbf{Method} & \textbf{Hospital 1,2,3} & \textbf{Hospital 4} & \textbf{Hospital 5} & \textbf{Target Avg.}\\
        \midrule
            ERM &  97.96 \scriptsize(0.04)&90.58 \scriptsize(0.63) & 82.26 \scriptsize(3.91) & 86.42 \scriptsize(1.68)\\
            AdvBNN & 96.48 \scriptsize(0.55) & 87.30 \scriptsize(2.14) & 80.79  \scriptsize(1.69) & 84.04 \scriptsize(1.65)\\
            Augmix & 96.15 \scriptsize(0.19) & 85.92 \scriptsize(1.28) & 84.86 \scriptsize(1.58) & 85.39 \scriptsize(1.36)\\ 
            Augmax & 95.17 \scriptsize(0.13) & 79.61 \scriptsize(1.49) & 85.51 \scriptsize(1.69) & 82.56 \scriptsize(1.30)\\
            RandConv & 97.98 \scriptsize(0.09) & 90.64 \scriptsize(1.24) & 84.75 \scriptsize(3.94) & 87.70 \scriptsize(1.69)\\
            $\text{ALT}_{\text{5-layer}}$ & 96.95 \scriptsize(0.10) & 90.82 \scriptsize(0.82) & 82.70 \scriptsize(1.22) & 86.76 \scriptsize(0.82)\\
            $\text{ALT}_{\text{5-layer+RandConv}}$ & 97.09 \scriptsize(0.01) & 91.09 \scriptsize(1.01) & 85.80 \scriptsize(1.59) & 88.44 \scriptsize(1.16)\\
            \rowcolor{Snow2} $\text{ALT}_{\text{5-layer+Augmix}}$ & 97.06 \scriptsize(0.10) & 91.77 \scriptsize(0.42) & 86.11 \scriptsize(3.35) & 88.94 \scriptsize(1.60)\\
            \rowcolor{white}
            $\text{ABA}_{\text{5-layer}}$ & 97.29 \scriptsize(0.14) & 91.19 \scriptsize(0.64) & 85.20 \scriptsize(3.45) & 88.19 \scriptsize(1.91)\\
            \rowcolor{LightCyan1}
            $\text{ABA}_{\text{5-layer+RandConv}}$ & 97.28 \scriptsize(0.10) & 90.89 \scriptsize(0.65) & \textbf{88.47 \scriptsize(0.96)} & 89.68 \scriptsize(0.80) \\
            $\text{ABA}_{\text{5-layer+Augmix}}$ & 97.23 \scriptsize(0.06) & \textbf{91.85 \scriptsize(0.78)} & 87.92 \scriptsize(0.59) & \textbf{89.88 \scriptsize(0.37)} \\
            \bottomrule
        \end{tabular}
    \end{tabular}

    \caption{\textbf{SSDG accuracy on Living17 (left panel) and Camelyon17 (right panel).} For Living17, the source domain is the two subclasses in each superclass and the target domain is the remaining two subclasses. For Camelyon17, the source domain is images from hospital 1,2,3 and the target domain is hospital 4 and 5. We report mean (and standard deviation) of 5 runs.}
    \label{tab:wilds}
\end{table*}

\begin{figure*}
    \centering
    \begin{adjustbox}{width=\linewidth}
    \begin{tabular}{ccc}
         \includegraphics{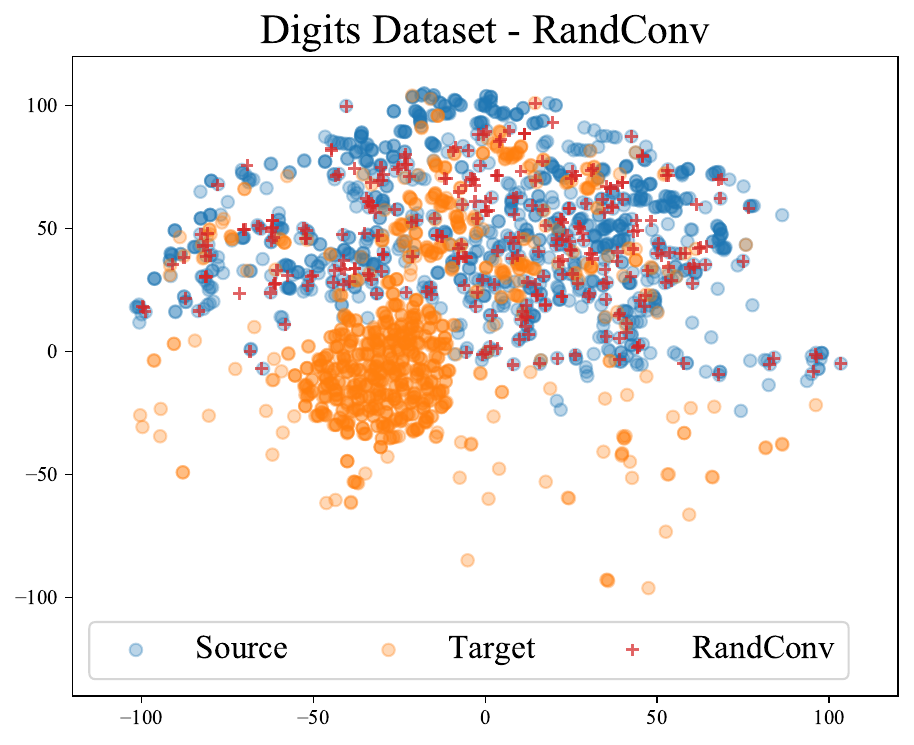} &
         \includegraphics{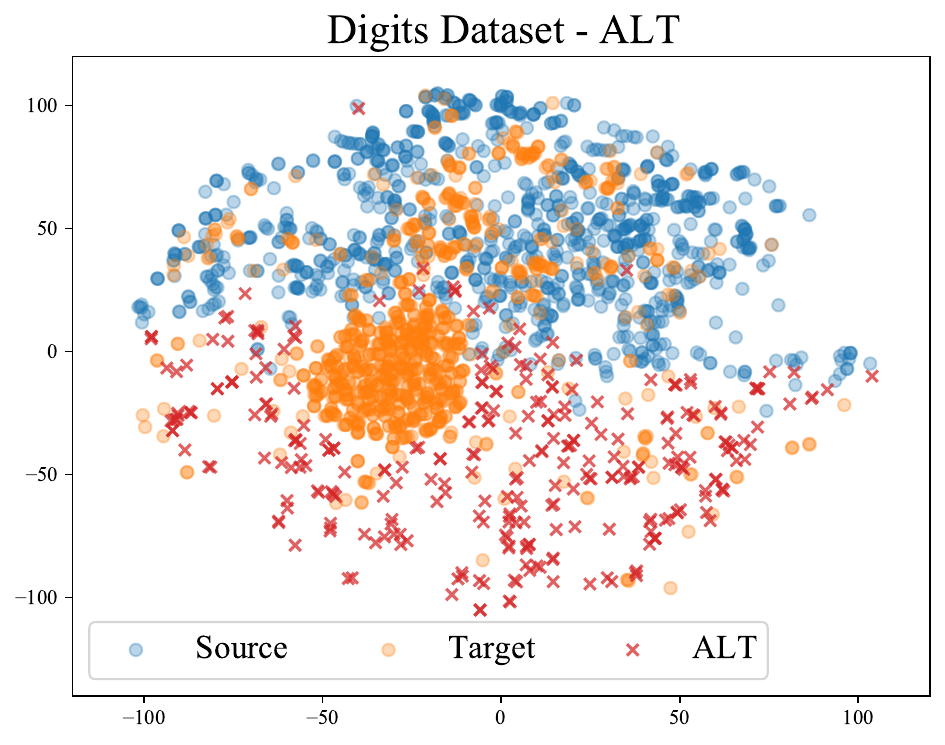} &
         \includegraphics{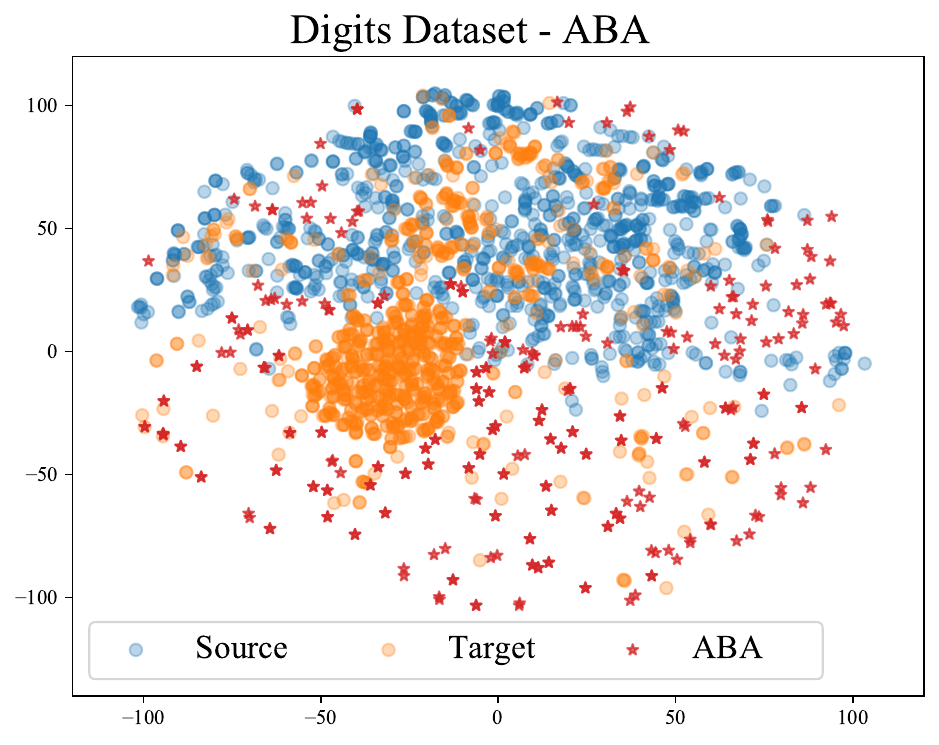} \\

    \end{tabular}
    \end{adjustbox}
    \caption{TSNE plot for source domain, target domain and augmented image distribution by RandConv, ALT, ABA.}
    \label{fig:tsne}
\end{figure*}

\medskip\noindent\textbf{Baselines.}
The baselines for Living17 dataset includes the Empirical Risk Minimization (ERM), AdvBNN~\cite{liu2018advbnn}, Augmix~\cite{hendrycks*2020augmix}, Augmax~\cite{wang2021augmax}, Randconv~\cite{xu2021robust} and ALT~\cite{gokhale2023improving}.
We consider three variants of ALT in our evaluation, by adding RandConv and Augmix module. 
we do not perform any hyperparameter tuning for Living17 and directly apply identical training settings and hyperparameters from PACS.

\medskip\noindent\textbf{Results.}
Table~\ref{tab:wilds} (left panel) presents our results on the Living17 dataset. 
We observe that several baseline models (including AdvBNN, Augmix and its adversarial variant Augmax) even worsen the performance on the target domain.
RandConv also causes a decrease in performance, although this could be due to it affecting coverage on the source domain, as the source domain accuracy is low.
In comparison, both ALT and our proposed method with RandConv module outperform ERM, with ALT achieving a 1.50\% improvement and our method achieving a 1.72\% improvement, which achieves the best result on Living17 dataset.

\subsection{Domain Shift in Medical Imaging}
{Camelyon17~\cite{bandi2018detection}}
is the medical imaging dataset for binary classification of tumor detection in the center $32 \times 32$ region.
The dataset is collected from 5 different hospitals.
Following the setting in~\cite{wilds2021}, we combine the data from the first 3 hospitals as the source domain and the remaining 2 hospitals as target domains. 
Note that while some multi-source domain generalization methods utilize domain labels (hospital numbers) for training, we do not use this information, but simply combine data from three hospitals as the single source domain.

\medskip\noindent\textbf{Baselines.}
The baseline models and experiment settings are same as the subpopulation shift experiment, but with difference of using  ResNet50 model as the feature extractor. 

\medskip\noindent\textbf{Results.}
We present the results of our experiments on the Camelyon17 dataset in Table~\ref{tab:wilds} (right panel).
We observe that AdvBNN performs worse than ERM.
Augmix, which composites image augmentation operations, leads to a downgrade in performance, while Augmax, which generates more adversarial samples, performs even worse.
However, RandConv can improve the performance, which is consistent with the properties of dataset that rely on shape and texture information.
ALT also improves accuracy, while $\text{ALT}_{\text{5-layer}}$ performs slightly lower than RandConv about 0.94\%, but $\text{ALT}_{\text{5-layer+Augmix}}$ achieves an improvement of 1.24\% over RandConv.
Our $\text{ABA}_{\text{5-layer}}$ achieves an improvement of 1.43\% compared with $\text{ALT}_{\text{5-layer}}$, and similarly, $\text{ABA}_{\text{5-layer+RandConv}}$ and $\text{ABA}_{\text{5-layer+Augmix}}$ outperforms $\text{ALT}_{\text{5-layer+RandConv}}$ and $\text{ALT}_{\text{5-layer+Augmix}}$ by 1.24\%, 0.94\%, achieving the best results on the Camelyon17 dataset.

\begin{figure*}[ht]
    \centering
    \begin{adjustbox}{width=\linewidth}
        \begin{tabular}{c@{\hskip 0.3in}c@{\hskip 0.3in}c @{\hskip 0.3in}c}
             \includegraphics{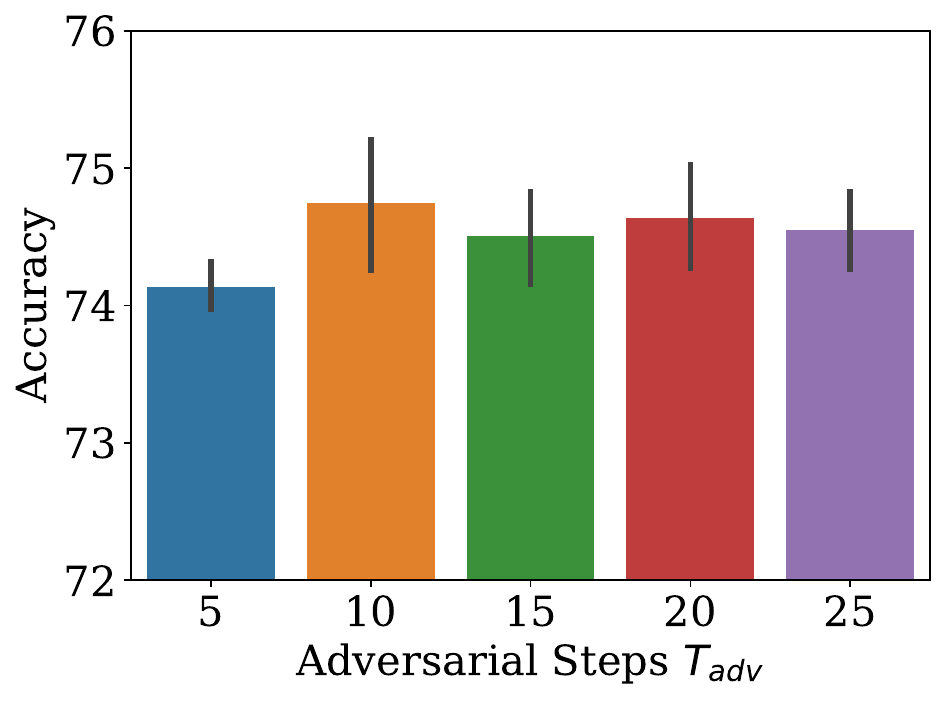} &
             \includegraphics{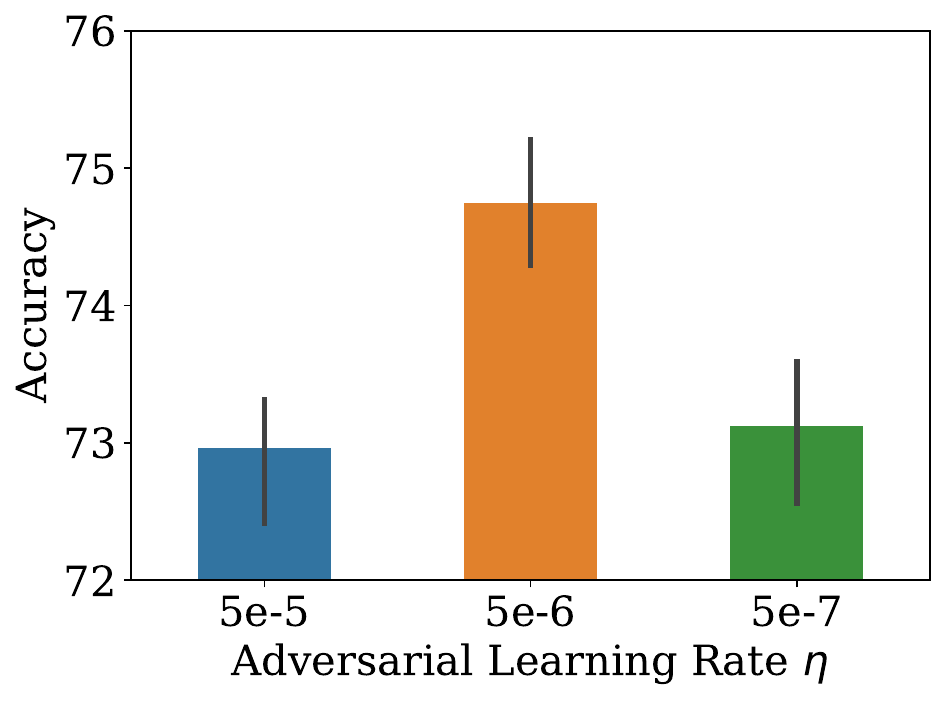} &
             \includegraphics{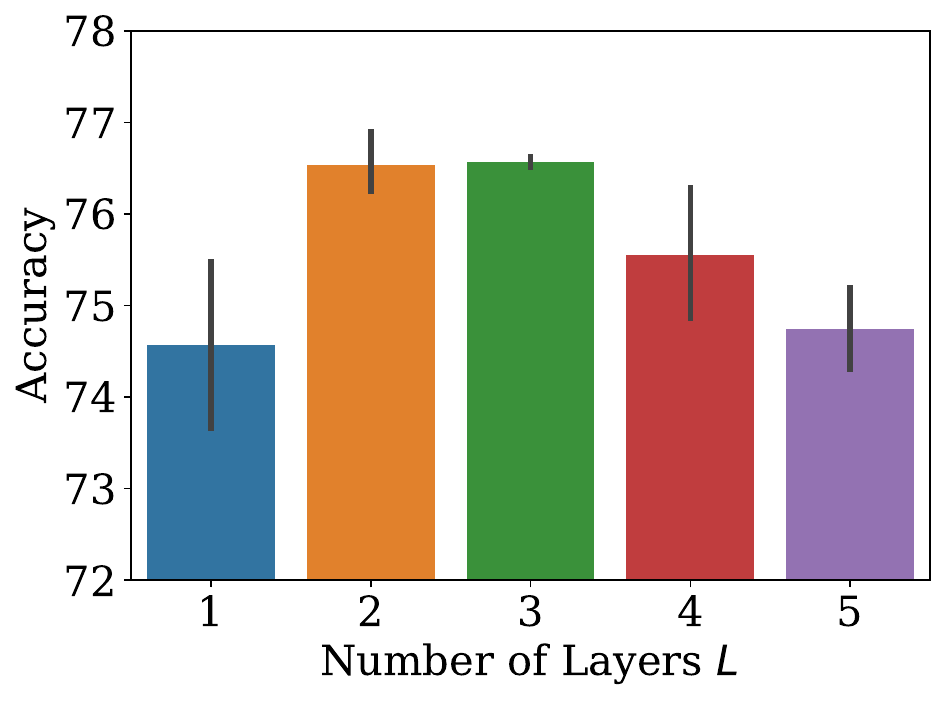} &
             \includegraphics{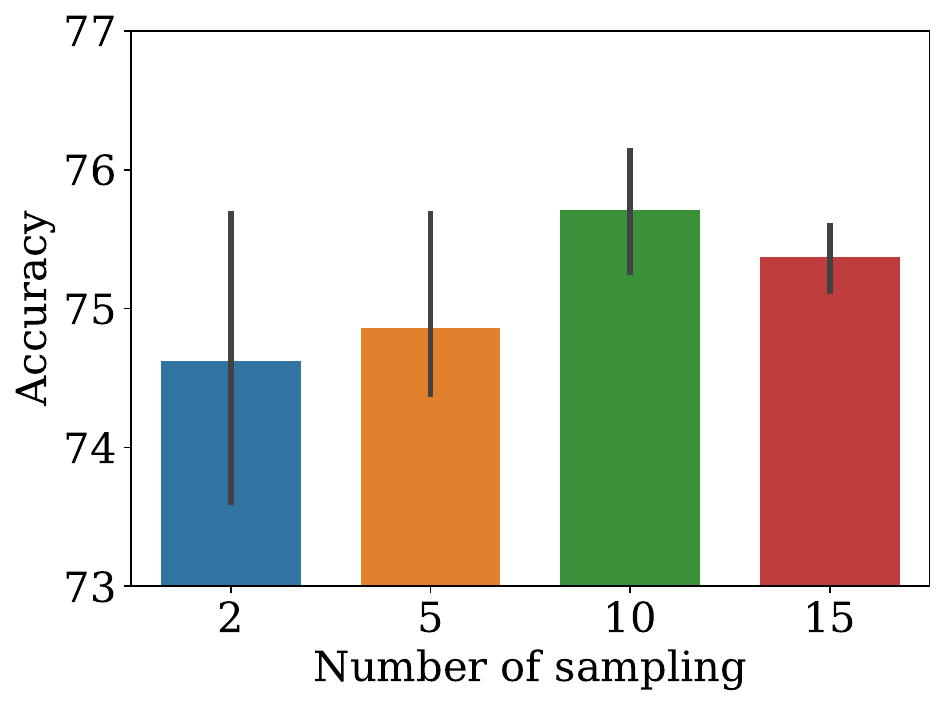}
        \end{tabular}
    \end{adjustbox}
    
    \caption{From left to right, we show the impact of hyperparameters on SSDG performance (Digits dataset): adversarial steps, adversarial learning rate, number of layers, and number of sampling per Bayesian convolutional layer.}
    \label{fig:hyper}
\end{figure*}

\begin{figure}
    \centering
    \begin{adjustbox}{width=\linewidth}
        \begin{tabular}{lc}
             \rotatebox[origin=c]{90}{\HUGE Input Image} &
             \makecell{
                \includegraphics{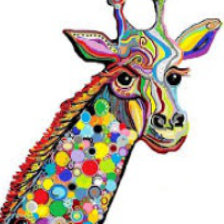}
                \includegraphics{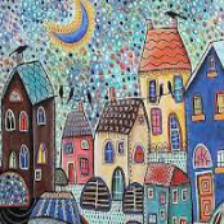}
                \includegraphics{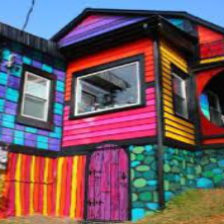}
                \includegraphics{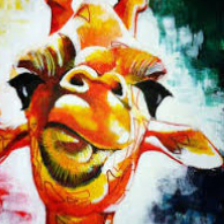}
                }
             \\
             \rotatebox[origin=c]{90}{\HUGE RandConv} &
             \makecell{
                \includegraphics{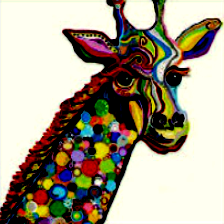}
                \includegraphics{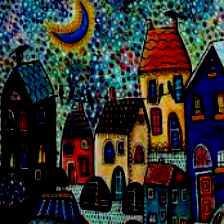}
                \includegraphics{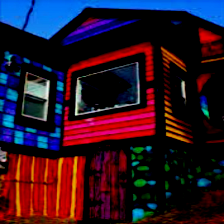}
                \includegraphics{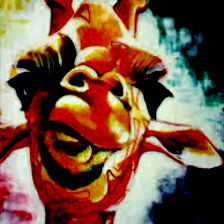}
                }
             \\
             \rotatebox[origin=c]{90}{\HUGE ALT} &
             \makecell{
                \includegraphics{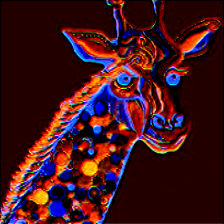}
                \includegraphics{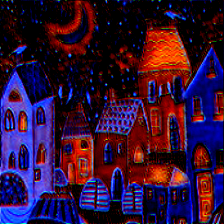}
                \includegraphics{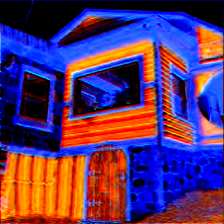}
                \includegraphics{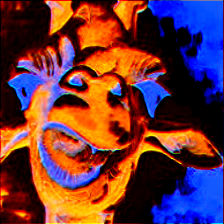}
                }
             \\
             \rotatebox[origin=c]{90}{\HUGE ABA} &
             \makecell{
                \includegraphics{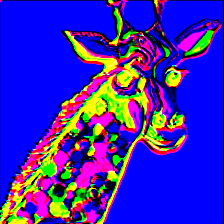}
                \includegraphics{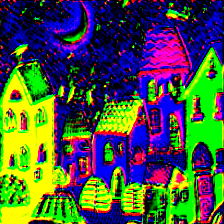}
                \includegraphics{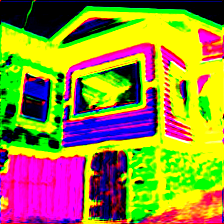}
                \includegraphics{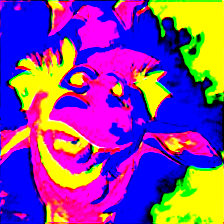}
                }
             \\
             
        \end{tabular}
    \end{adjustbox}
    \caption{Qualitative comparison of PACS images augmented by RandConv, ALT and our ABA.}
    \label{fig:sample}
\end{figure}

\section{Analysis} 
In this section we provide further insights on our results, analyses, and ablation studies.
\subsection{Key Insights from Results}\label{sec:ana}
\paragraph{Limitations of Data Augmentation.}
First, we observe that every data augmentation method has its limitation on datasets and types of SSDG. 
For example, AdvBNN, which incorporates the adversarial training of the feature extractor for robustness against the adversarial perturbations, exhibits inferior performance compared to ERM on all four datasets.
Augmix performs well on PACS, but not outstanding on the Digits dataset, both of which are style generalization dataset.
It even worsens the performance on the generalization of subpopulation and medical imaging.
RandConv boosts performance on style generalization, and medical imaging generalization, but hurts the subpopulation shift generalization.
Among these baseline models, ALT can keep outperforming most state-of-art methods, but 1) 1 layer ALT does not show significant improvement over RandConv on Digits dataset and 2) ALT sometimes needs additional RandConv or Augmix module to improve the performance.

\medskip\noindent\textbf{ABA without additional modules outperforms previous state-of-the-art.}
The next insight is that even without any explicit module, ABA outperforms all other data augmentation methods, even 1-layer ABA without Randconv module.
This indicates that adversarial learned Bayesian convolutional neural networks are powerful for generating augmented images to train a classifier for generalization -- we can see the benefit of using ABA compared with ALT and RandConv. 
Moreover, we did not clearly observe the benefits of adding additional RandConv or Augmix module to our ABA method. 
For example, in the PACS dataset and cartoon as the source domain, adding RandConv can improve performance for the 1-layer ABA, but not for the 5-layer ABA.
The addition of Augmix augmentation improves performance on the Digits and Camelyon17 dataset, but does not yield similar improvement on the PACS and Living17 dataset.
In order to facilitate a fair comparison with RandConv and ALT which employ 1-layer and 5-layer convolution-based augmentations respectively, we conduct most of our experiments using 1-layer and 5-layer ABA.
However, for style shift, our model attains state-of-the-art performance with 3-layer ABA.

\medskip\noindent\textbf{ABA results in diverse and distributed augmentations.}
In Figure~\ref{fig:tsne}, we analyze the feature distribution introduced by ABA on Digits dataset and compare it with the source distribution (MNIST-10K), target distribution (SYNTH), and feature distributions from ALT~\cite{gokhale2023improving} and RandConv~\cite{xu2021robust}. 
Both ALT and ABA use 1-layer convolutional network. 
We observe that RandConv can generate some augmented images outside the source domain but still close to it, while ALT, using adversarial learning, mostly generates images outside the source domain. 
Our ABA method is spread widely across the tSNE~\cite{hinton2002stochastic} space, thanks to the advantage of adversarial learning and Bayesian inference. 
We also show the qualitative results of augmented images by RandConv, ALT and ABA in Figure~\ref{fig:sample}. 
We use 1-layer convolutional network for both ALT and ABA.
We find that our method can stylize images while still retaining important texture and shape information.

\subsection{Ablation study on Hyperparameters}
As illustrated in Section~\ref{sec:ABA}, we control the strength of the adversarial samples by adjusting the adversarial learning rate $\eta$ and the number of adversarial steps $T_{adv}$.
For ABA, another hyperparameter that determines the model is the number of convolutional layers $L$.
We also explore the impact of the number of sampling per Bayesian convolutional layer on 1-layer ABA.
In our paper, to manage computational costs, we adopt a single sampling per layer in the multiple layers BNN.
When our method is used without additional augmentation methods, we sample twice per entire BNN network.
We investigate the effect of each of these parameters on SSDG in Figure~\ref{fig:hyper}.
The experiments are conducted on the Digits dataset.

The first plot shows that the number of adversarial steps has little impact on SSDG, once it surpasses 5 steps.
While the best results are achieved with 10 adversarial steps, all other results still outperform the previous baseline models.
In the second plot, we analyze how the adversarial learning rate affects the results.
We find that  $\eta=5e^{-6}$ achieves the best performance, while $\eta=5e^{-5}$ or $\eta=5e^{-7}$ may generate adversarial samples that are either too strong or too weak for domain generalization.
The third plot demonstrates the importance of the number of ABA layers, with 3-layer ABA achieving the best results. 
With 1 or 2 layers, the network may not be capable of generating strong enough augmented images, while increasing the number of layers may result in augmented images that are too strong and hurt performance.
The last plot shows the impact of the number of sampling of 1-layer ABA.
With increasing the number of samplings results in more diverse training samples and improved target accuracy, but the performance starts to decline when the number of sampling exceeds 10.
It is clear that these hyperparameters impact the performance of ABA on SSDG.

\section{Conclusion}
In this paper, we demonstrate how adversarial learning combined with Bayesian convolutional neural network can generate more diverse samples, leading to an improvement in the performance of image classifiers on the single-source domain generalization task.
Our method, ABA, outperforms all existing methods on multiple benchmark datasets spanning different types of domain shift.
The promising results from this work spark potential future research, such as exploring whether the Bayesian neural network as a feature extractor can improve SSDG.

\section*{Acknowledgements}
This work was supported by NSF RI grants \#1750082 and \#2132724.
The views and opinions of the authors expressed herein do not necessarily state or reflect those of the funding agencies and employers.

{\small
\bibliographystyle{ieee_fullname}
\bibliography{egbib}
}

\appendix

\section*{Appendix}
The appendix contains the details of Living17 dataset, additional results on PACS dataset and more visualizations.

\section{Details of Living17}
We show the partition of the source domain and the target domain in Table~\ref{tab:living17_class}. The dataset contains 17 superclasses.
Each superclass has four subclasses.
We select two of the four for the source domain and the remaining two as the target domain.
The partition follows ~\cite{santurkar2021breeds}.
The class id in the table follows the setting from the ImageNet dataset~\cite{deng2009imagenet}.

\begin{table}[!h]
    \centering
    \small
    
    \begin{tabular}{@{}p{0.39\linewidth}cc@{}}
    \toprule
    \multirow{2}{*}{\textbf{Super Class}}   & \multicolumn{2}{c}{\textbf{Class IDs}} \\
     & \textbf{Source Domain} & \textbf{Target Domain}\\
    \midrule
    salamander & 27, 29 & 26, 28 \\
    turtle & 37, 34 & 33, 35\\
    lizard & 41, 44 & 47, 38\\
    snake, serpent, ophidian & 60, 57 & 65, 61\\
    spider & 76, 72 & 74, 77\\
    grouse & 81, 83 & 82, 80\\
    parrot & 88, 90 & 87, 89\\
    crab & 118, 120 & 119, 121\\
    dog, domestic dog, Canis familiaris & 163, 154 & 257, 157\\
    wolf & 272, 271 & 270, 269\\
    fox & 280, 279 & 277, 278\\
    domestic cat, house cat, Felis domesticus, Felis catus & 282, 285 & 283, 284  \\
    bear& 297, 295 & 296, 294\\
    beetle & 305, 306 & 302, 303\\
    butterfly & 325, 321 & 324, 322\\
    ape & 368, 365 & 366, 367\\
    monkey & 377, 380 & 381, 379\\
    \bottomrule
    \end{tabular}

    \caption{The partition of Living17 for source domain and target domain. For each superclass, the source domain contains two subclasses and the target domain contains two subclasses.}
    \label{tab:living17_class}
\end{table}

\section{Details results of PACS}
We show the detailed results, including the standard deviation for our model on the PACS dataset.
We compare our model with ERM, Augmix~\cite{hendrycks*2020augmix}, and Convolutional-based augmentations, such as RandConv~\cite{xu2021robust} and ALT~\cite{gokhale2023improving}.
Results are shown in Table~\ref{tab:pacs_p}, \ref{tab:pacs_a}, \ref{tab:pacs_c}, \ref{tab:pacs_s}, corresponding to Photo, Art painting, Cartoon and Sketch as the source domain. 

\section{Visualizations}
\begin{figure*}
    \centering
    \begin{adjustbox}{width=\linewidth}
        \begin{tabular}{ccc}
             \includegraphics{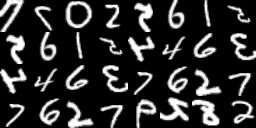} &
             \includegraphics{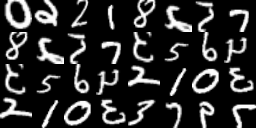} &
             \includegraphics{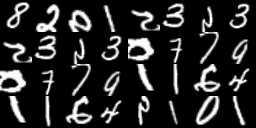} \\
             \includegraphics{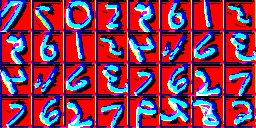} &
             \includegraphics{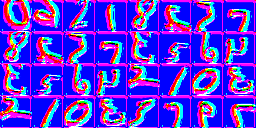} &
             \includegraphics{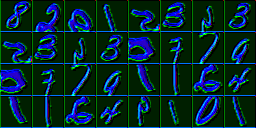} \\
        \end{tabular}
    \end{adjustbox}
    
    \caption{Digits: Images augmented by $\text{ABA}_{\text{3-layer}}$ with MNIST-10K as source dataset.}
    \label{fig:sample_digit}
\end{figure*}

\begin{figure*}
    \centering
    \begin{adjustbox}{width=\linewidth}
        \begin{tabular}{ccc}
             \includegraphics{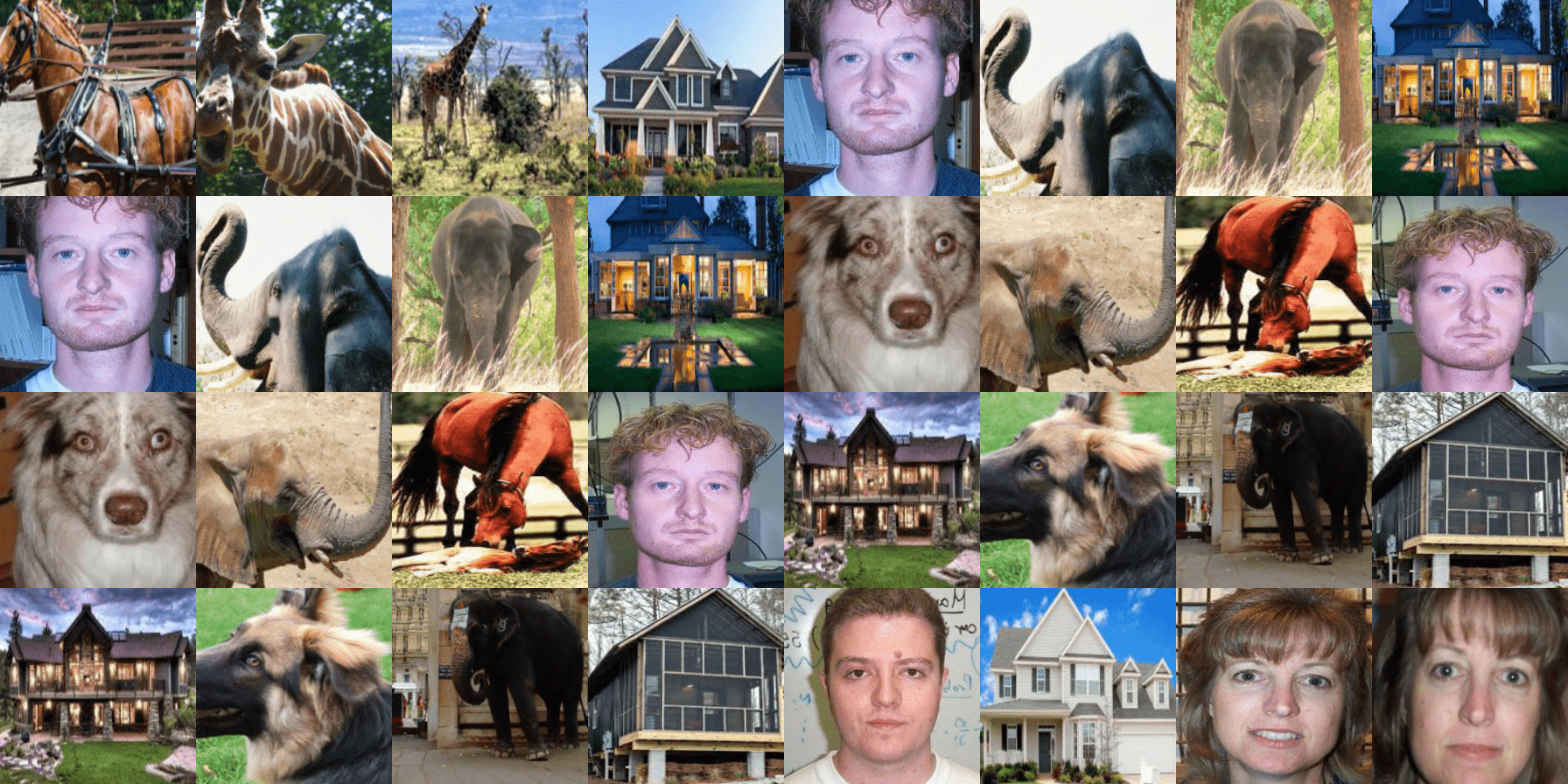} &
             \includegraphics{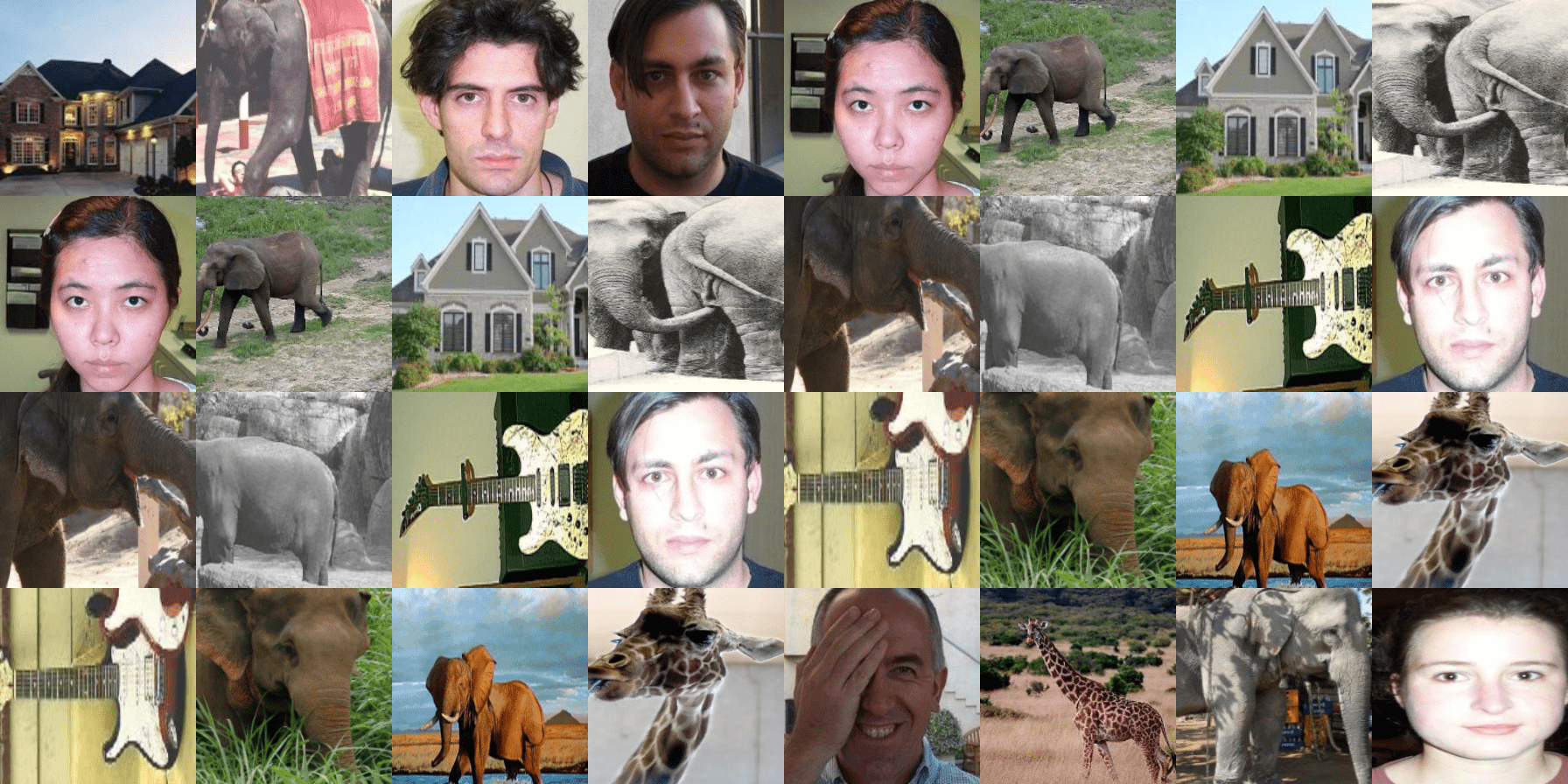} &
             \includegraphics{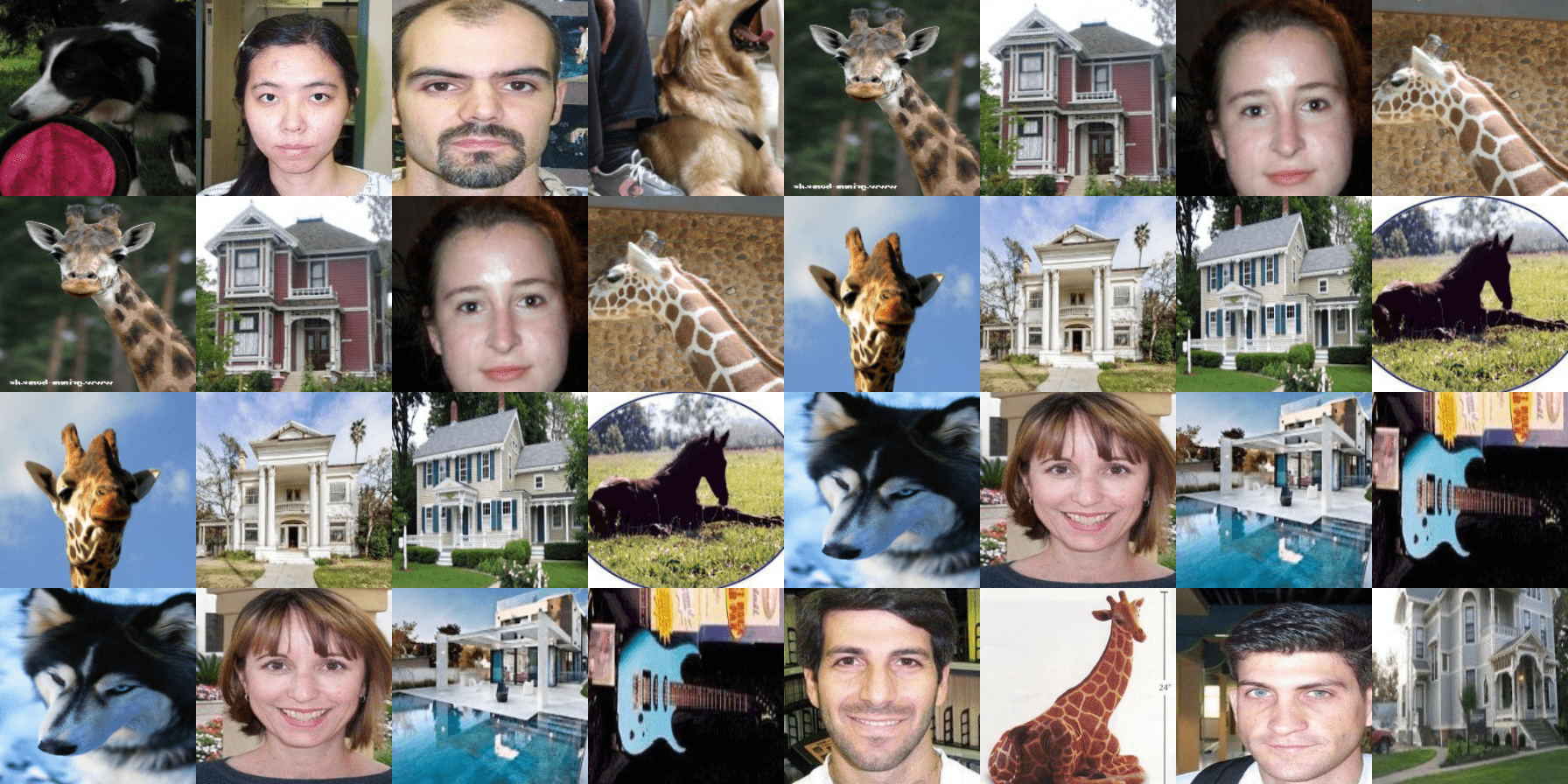} \\
             \includegraphics{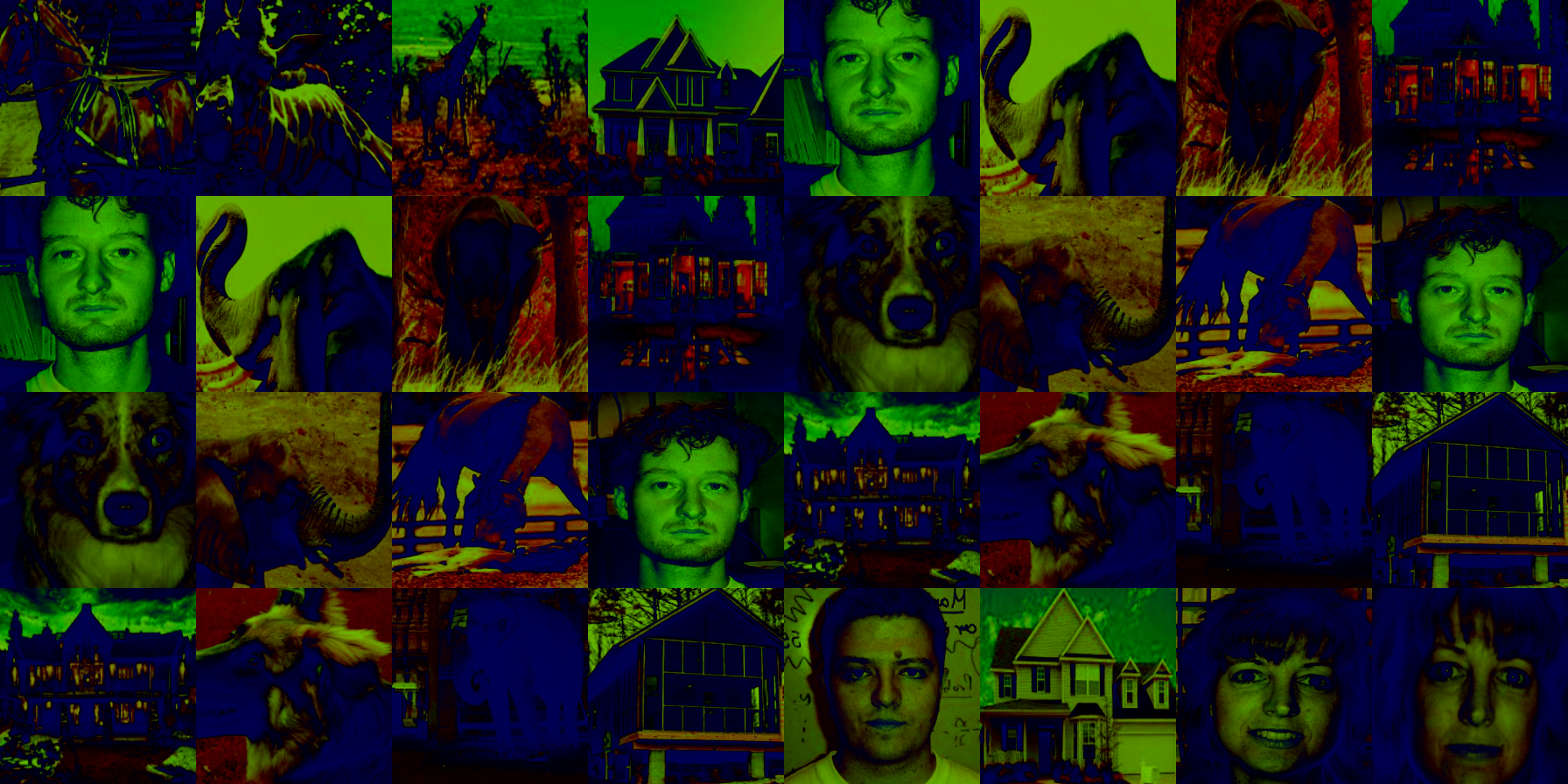} &
             \includegraphics{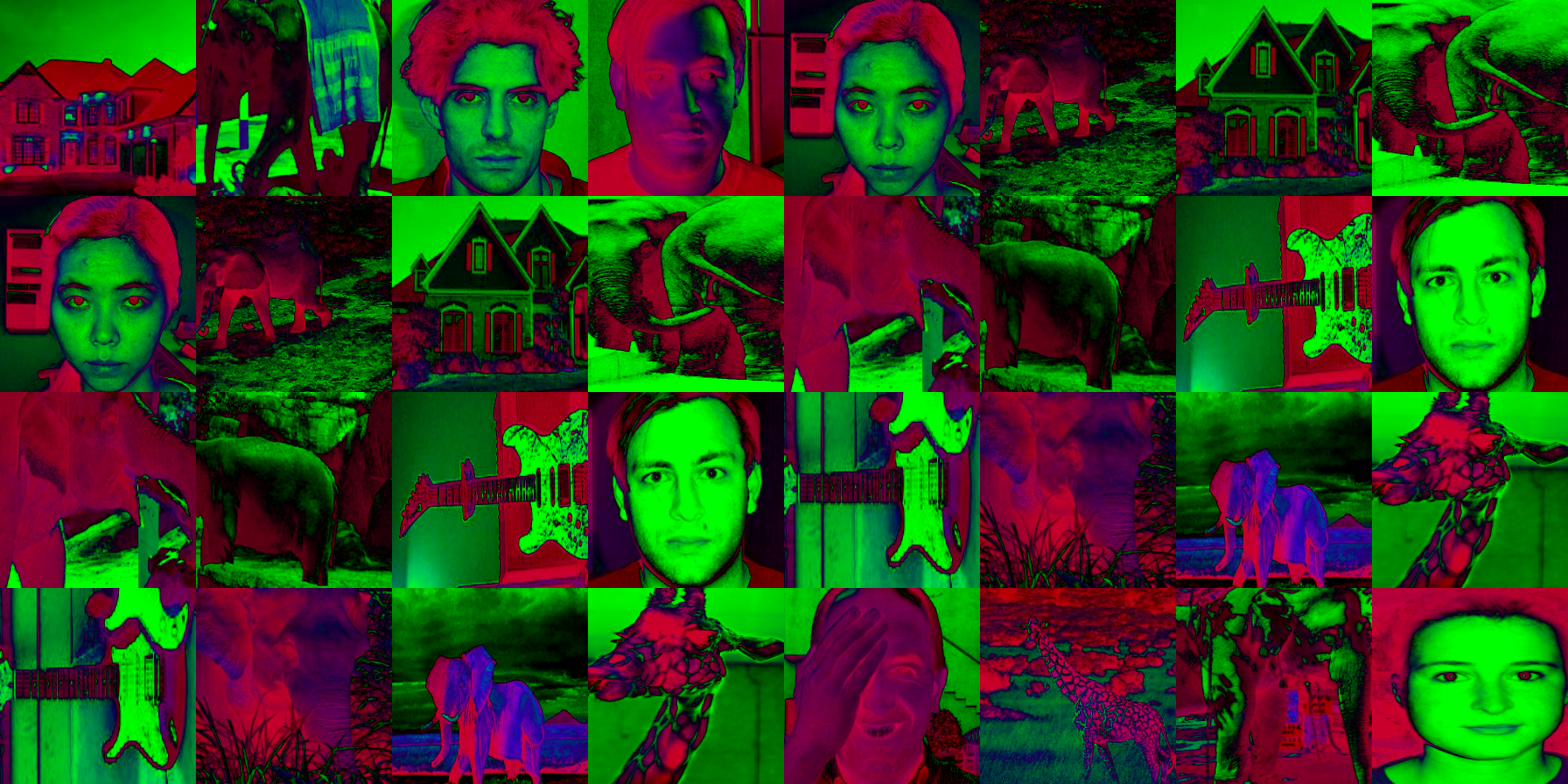} &
             \includegraphics{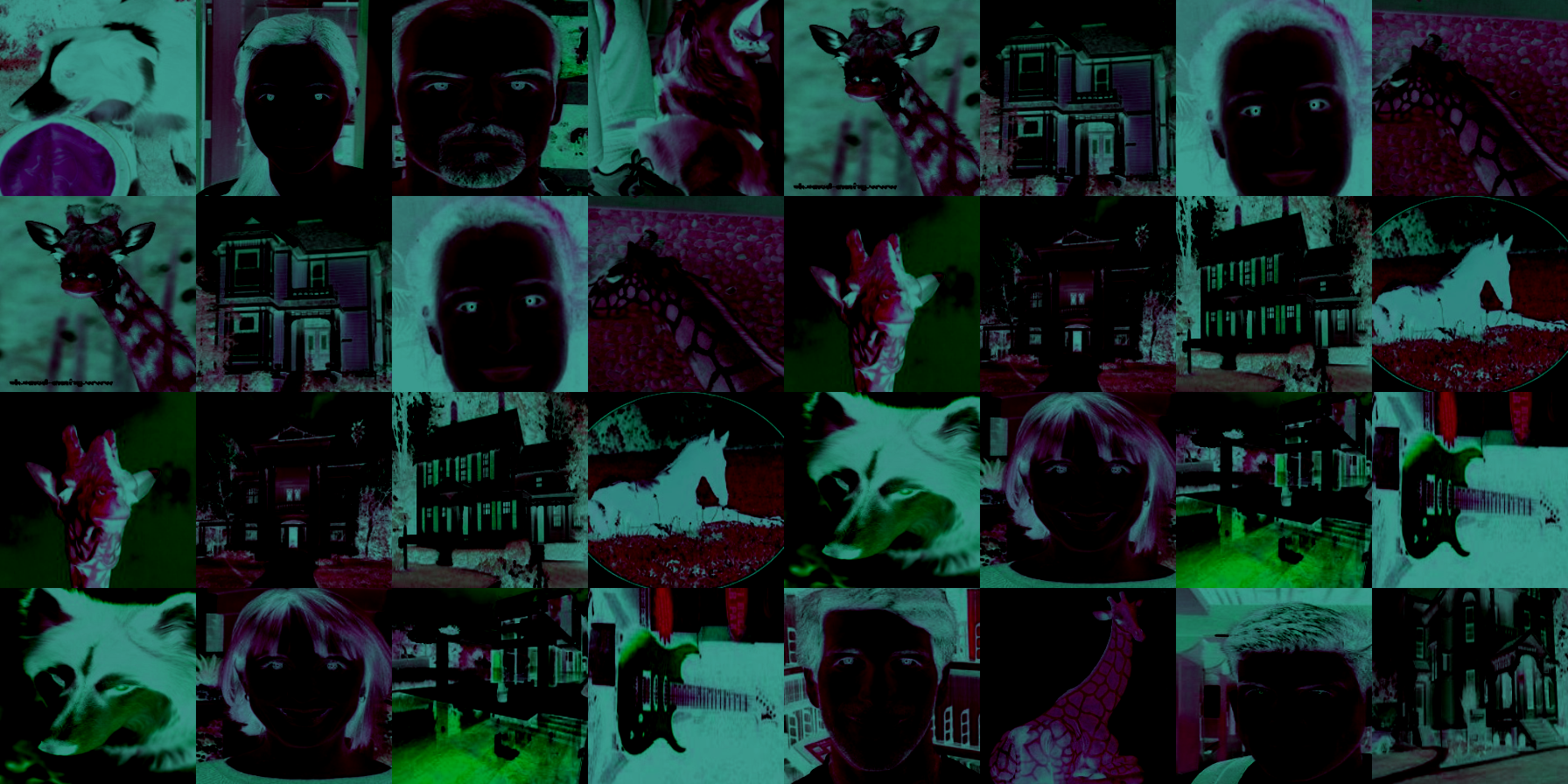} \\
        \end{tabular}
    \end{adjustbox}
    
    \caption{PACS: Images augmented by $\text{ABA}_{\text{3-layer}}$ with photo as source dataset.}
    \label{fig:sample_pacs_p}
\end{figure*}

\begin{figure*}
    \centering
    \begin{adjustbox}{width=\linewidth}
        \begin{tabular}{ccc}
             \includegraphics{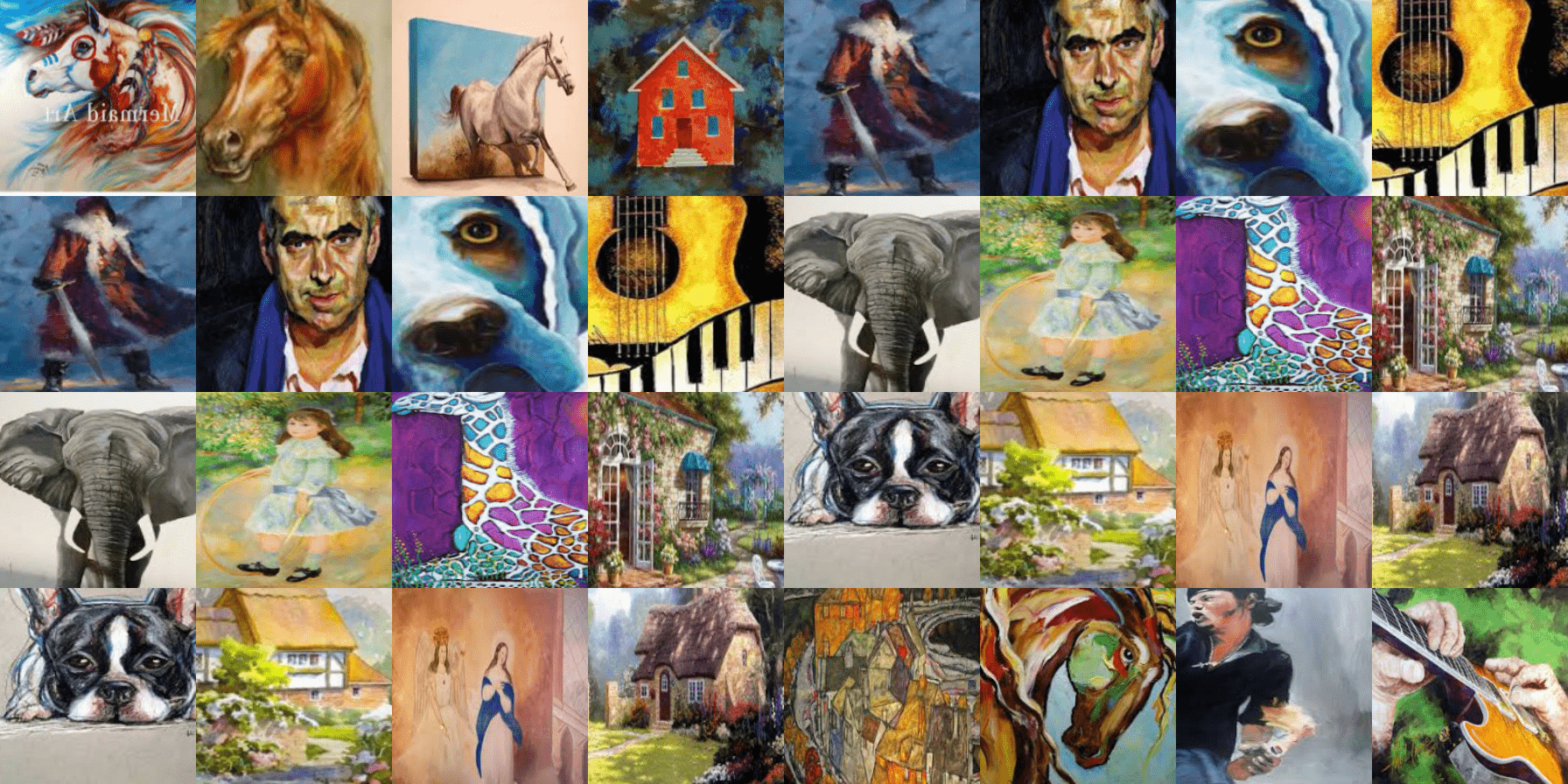} &
             \includegraphics{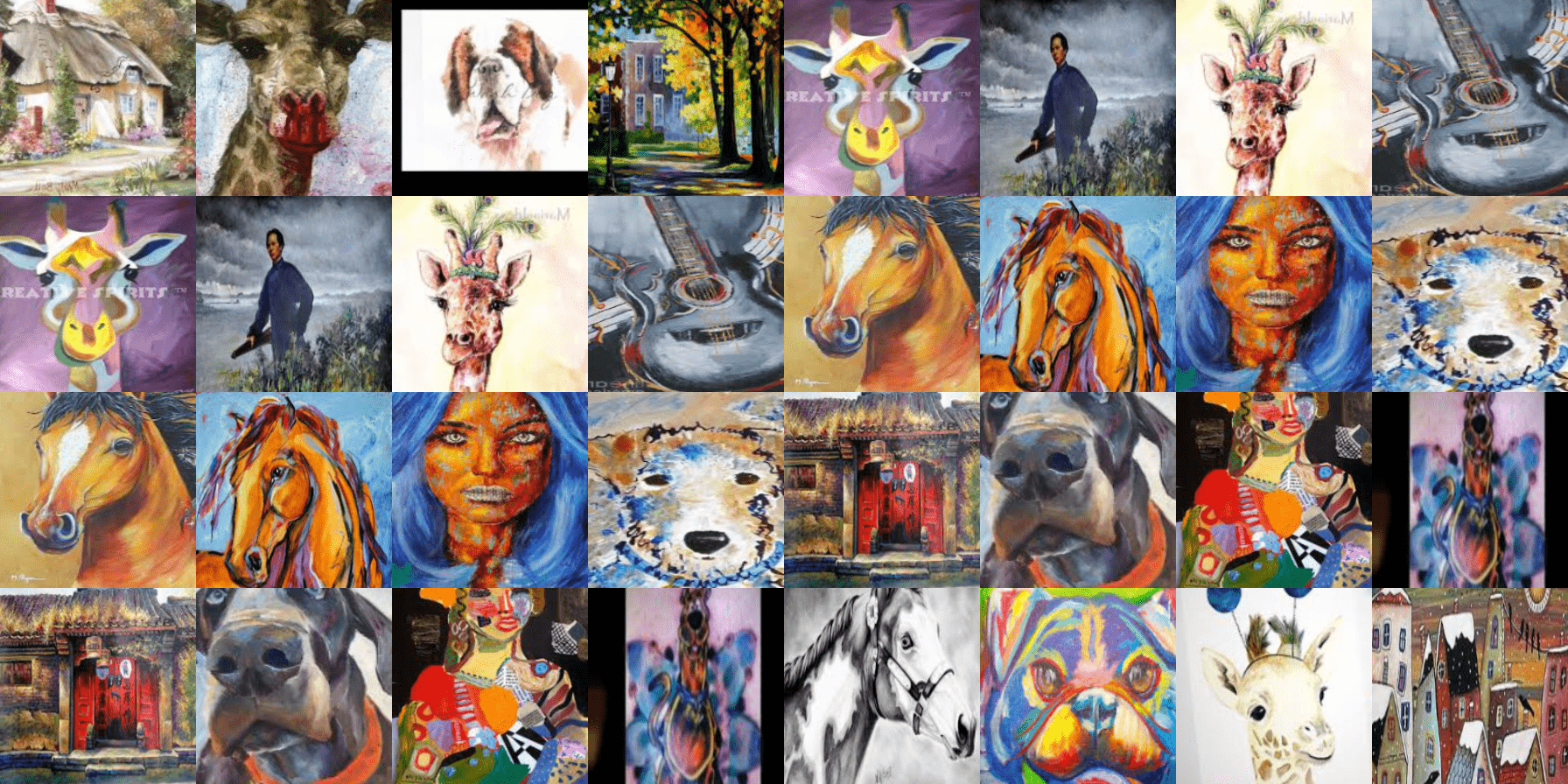} &
             \includegraphics{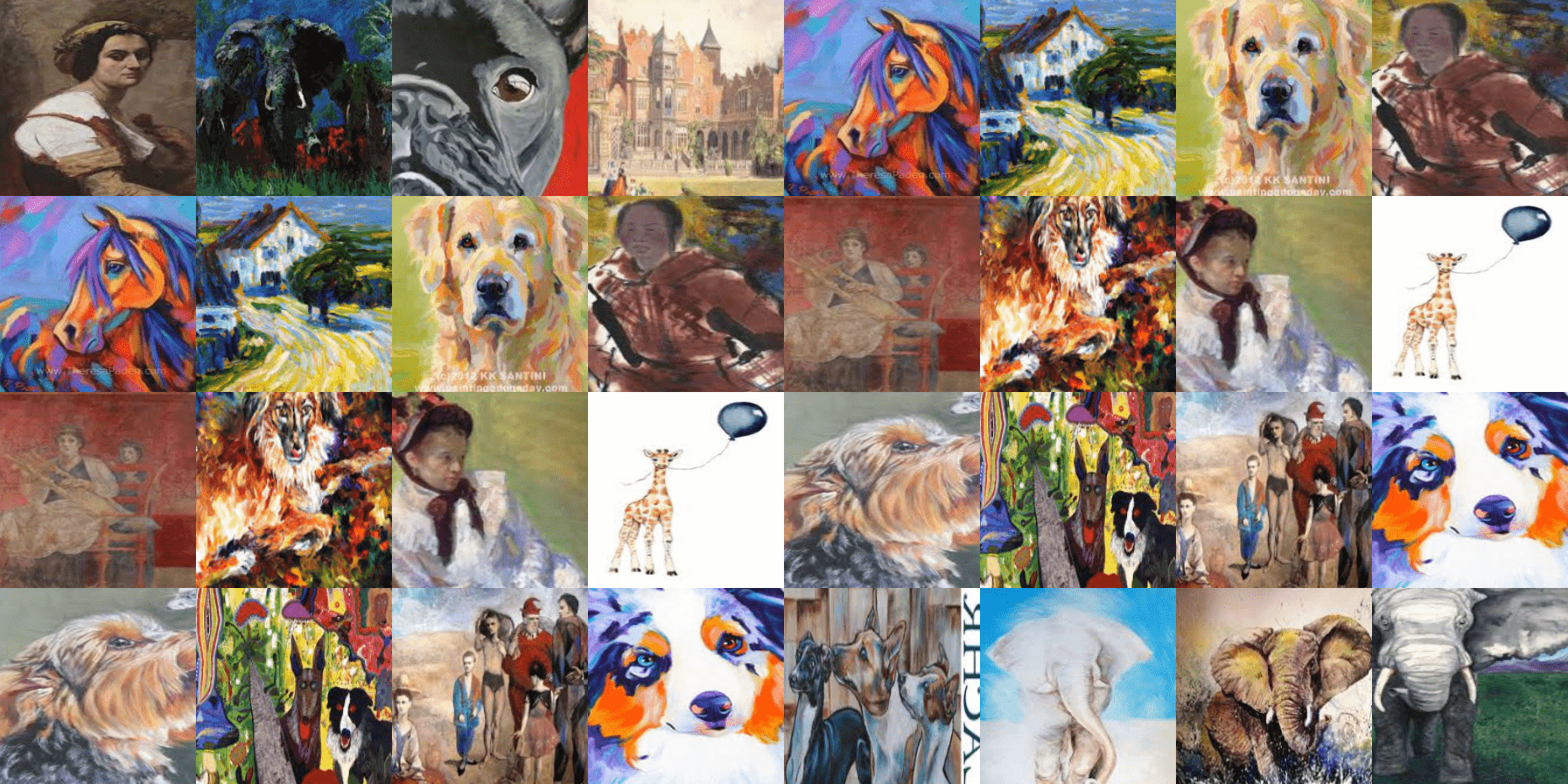} \\
             \includegraphics{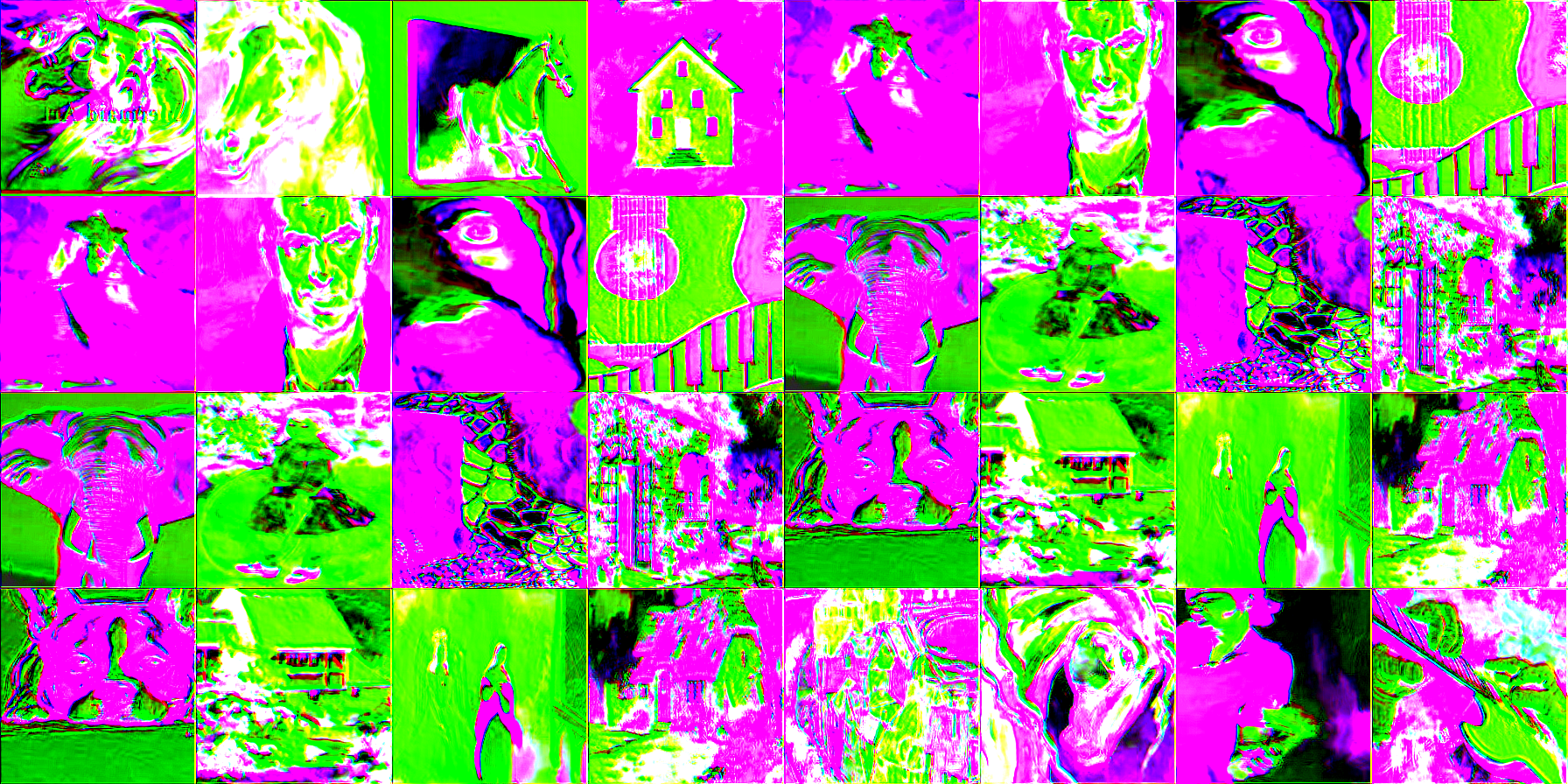} &
             \includegraphics{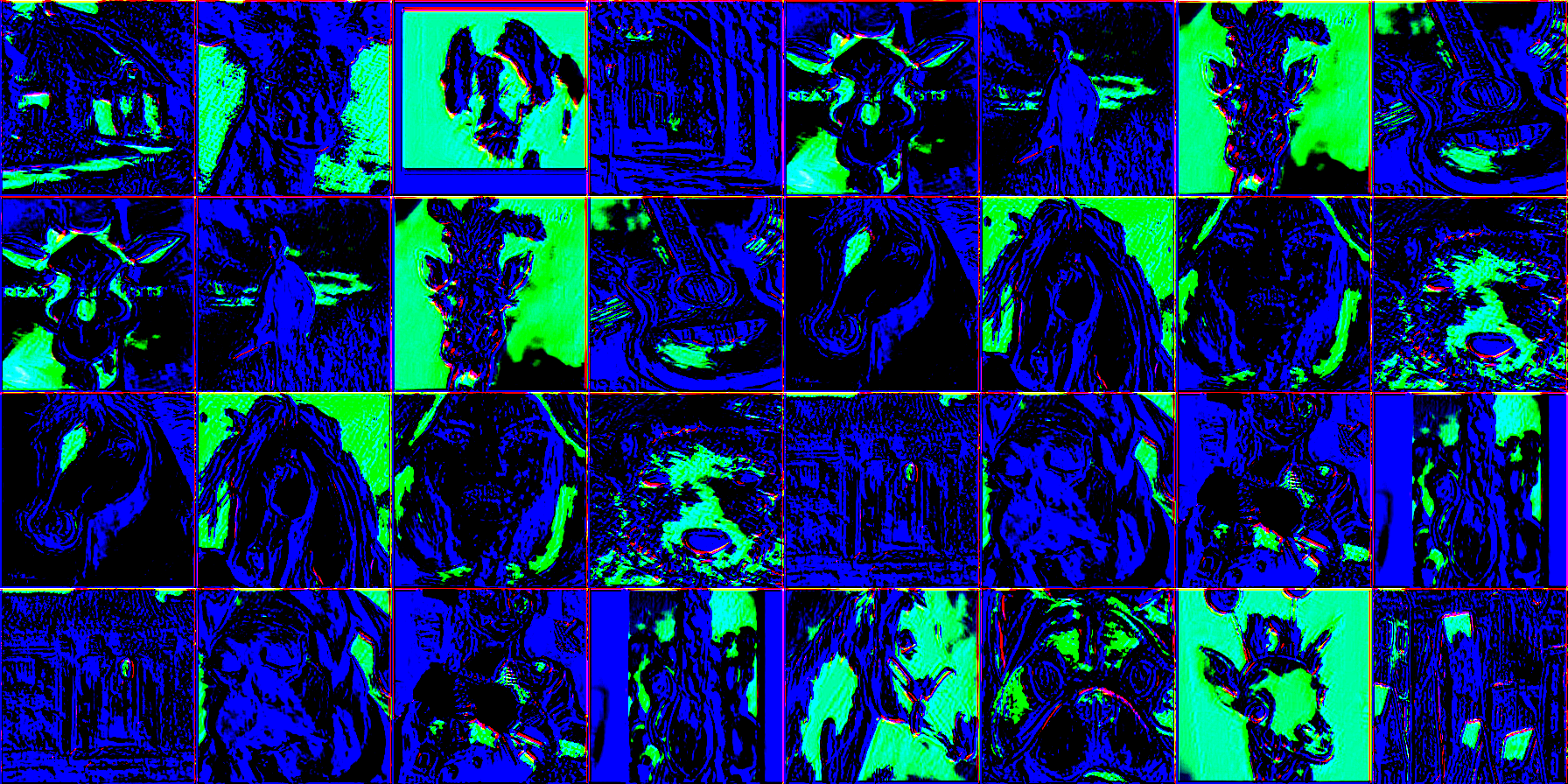} &
             \includegraphics{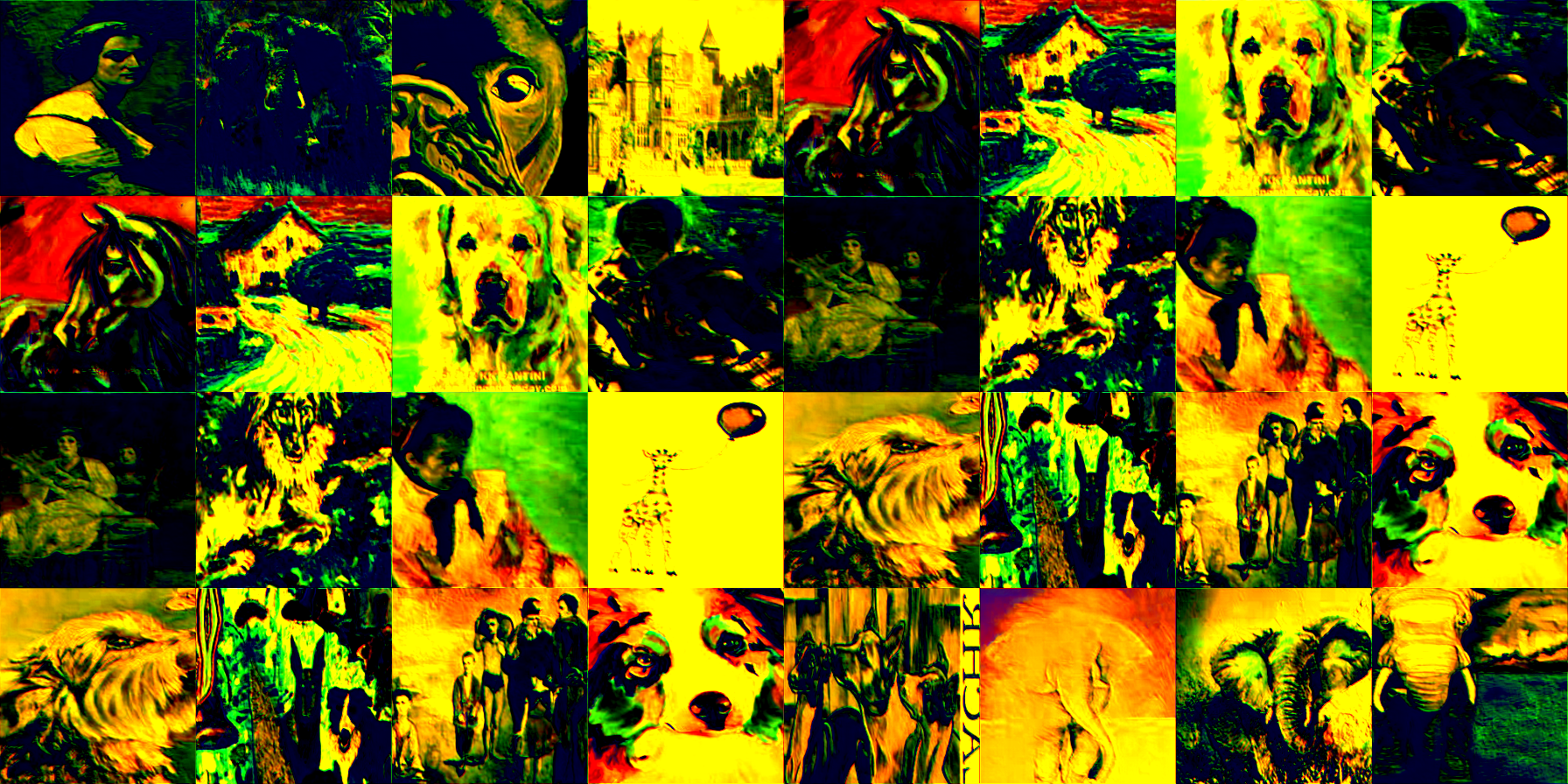} \\
        \end{tabular}
    \end{adjustbox}
    
    \caption{PACS: Images augmented by $\text{ABA}_{\text{3-layer}}$ with art painting as source dataset.}
    \label{fig:sample_pacs_a}
\end{figure*}

\begin{figure*}
    \centering
    \begin{adjustbox}{width=\linewidth}
        \begin{tabular}{ccc}
             \includegraphics{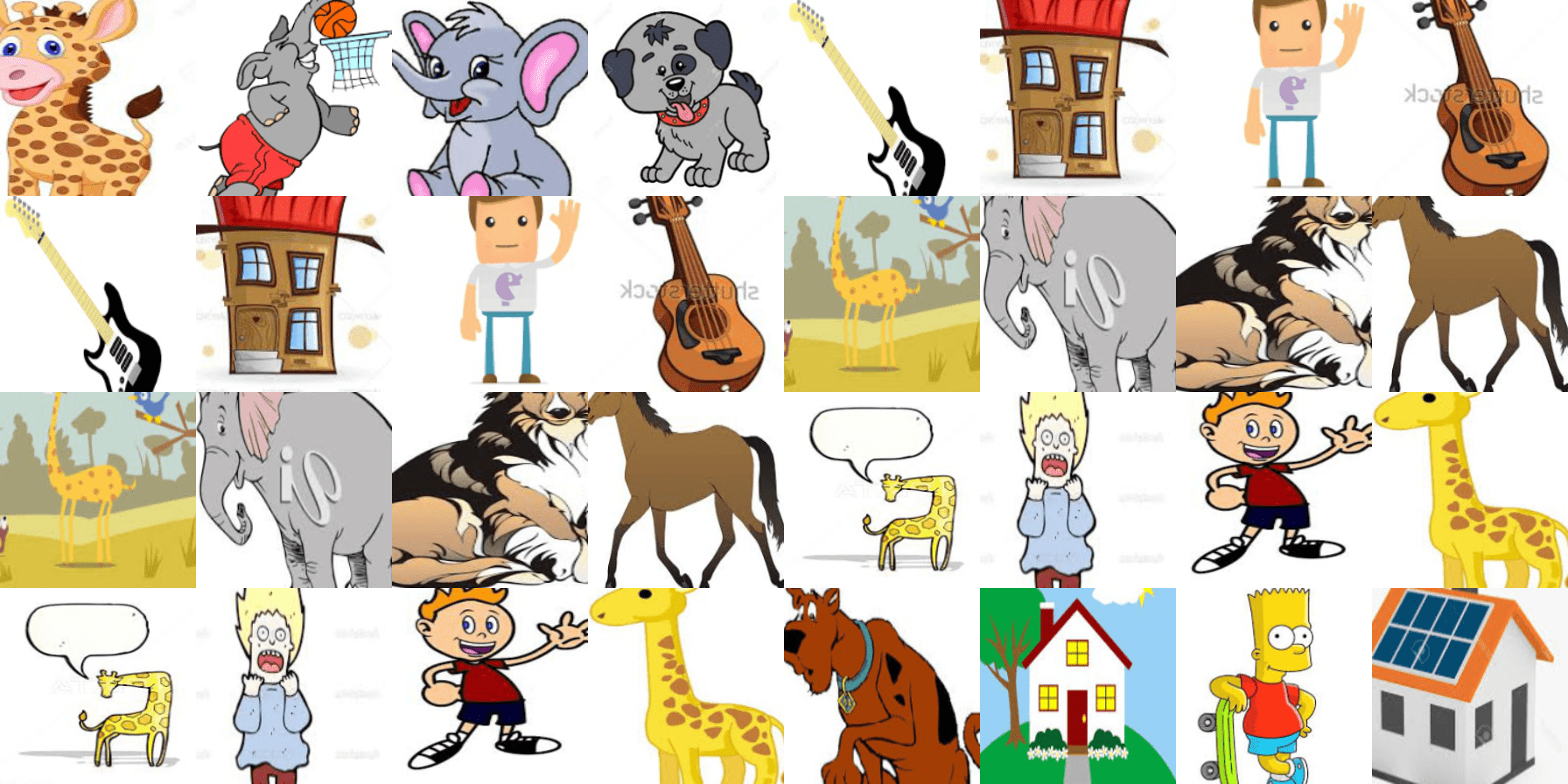} &
             \includegraphics{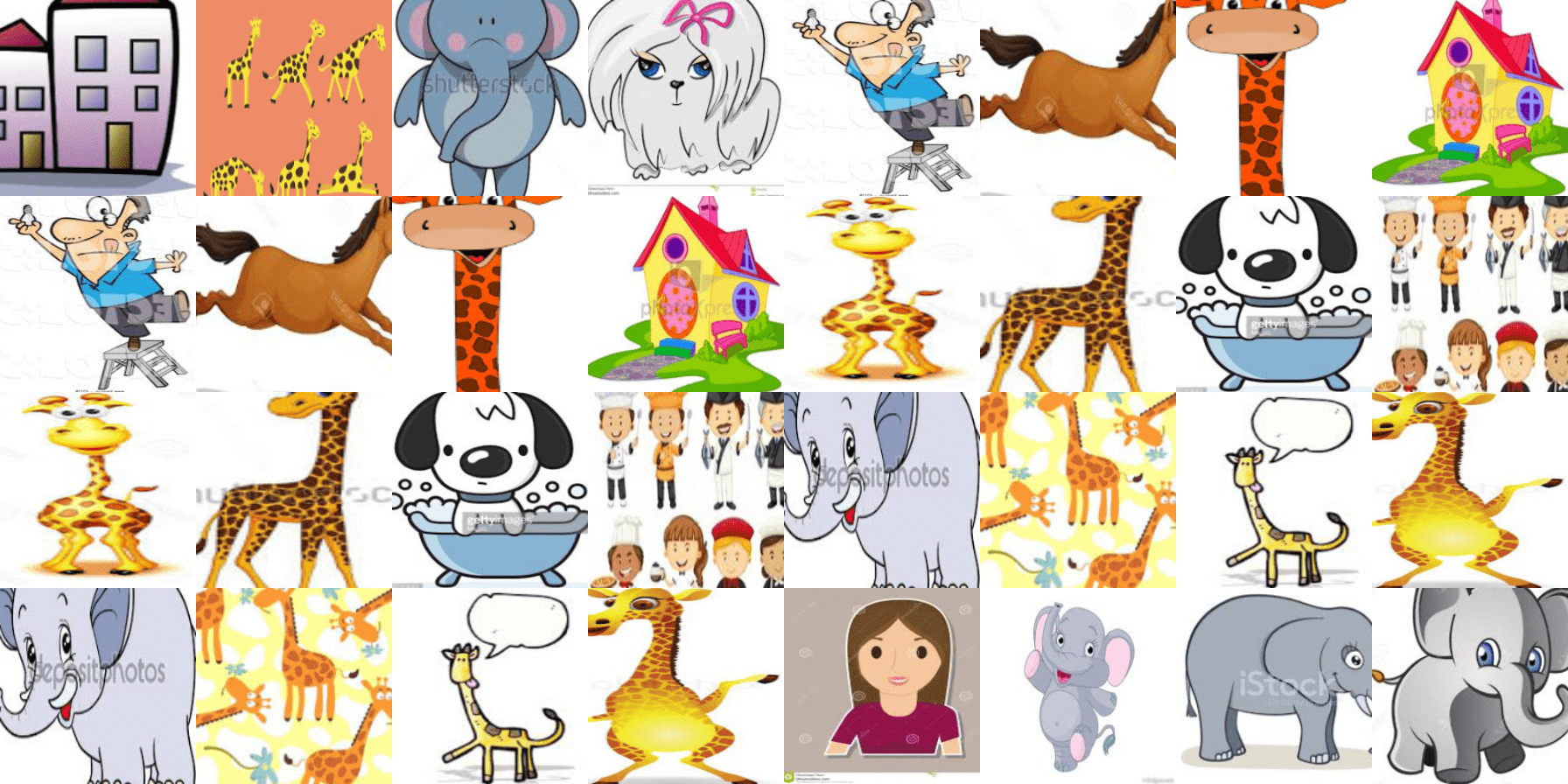} &
             \includegraphics{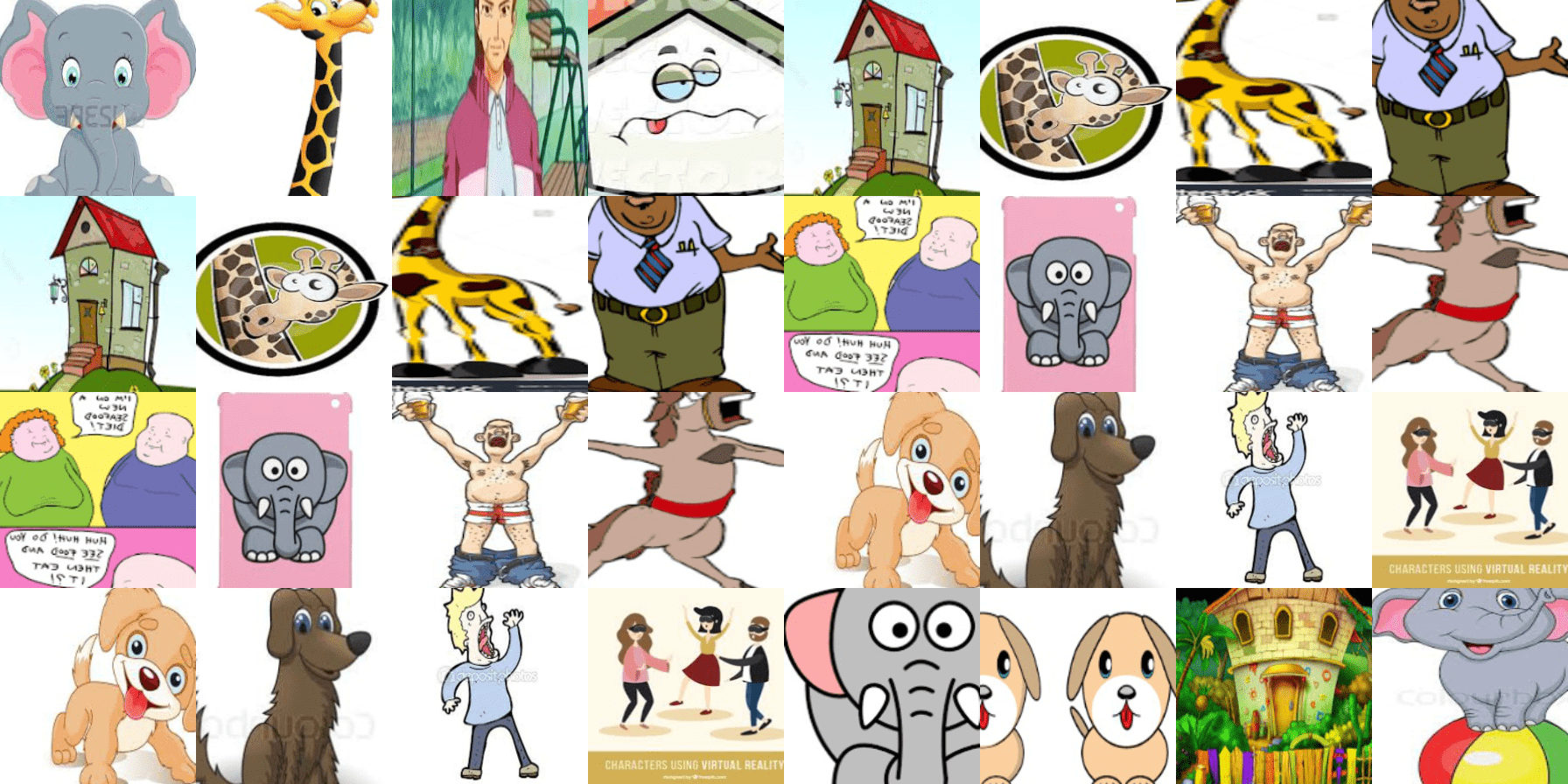} \\
             \includegraphics{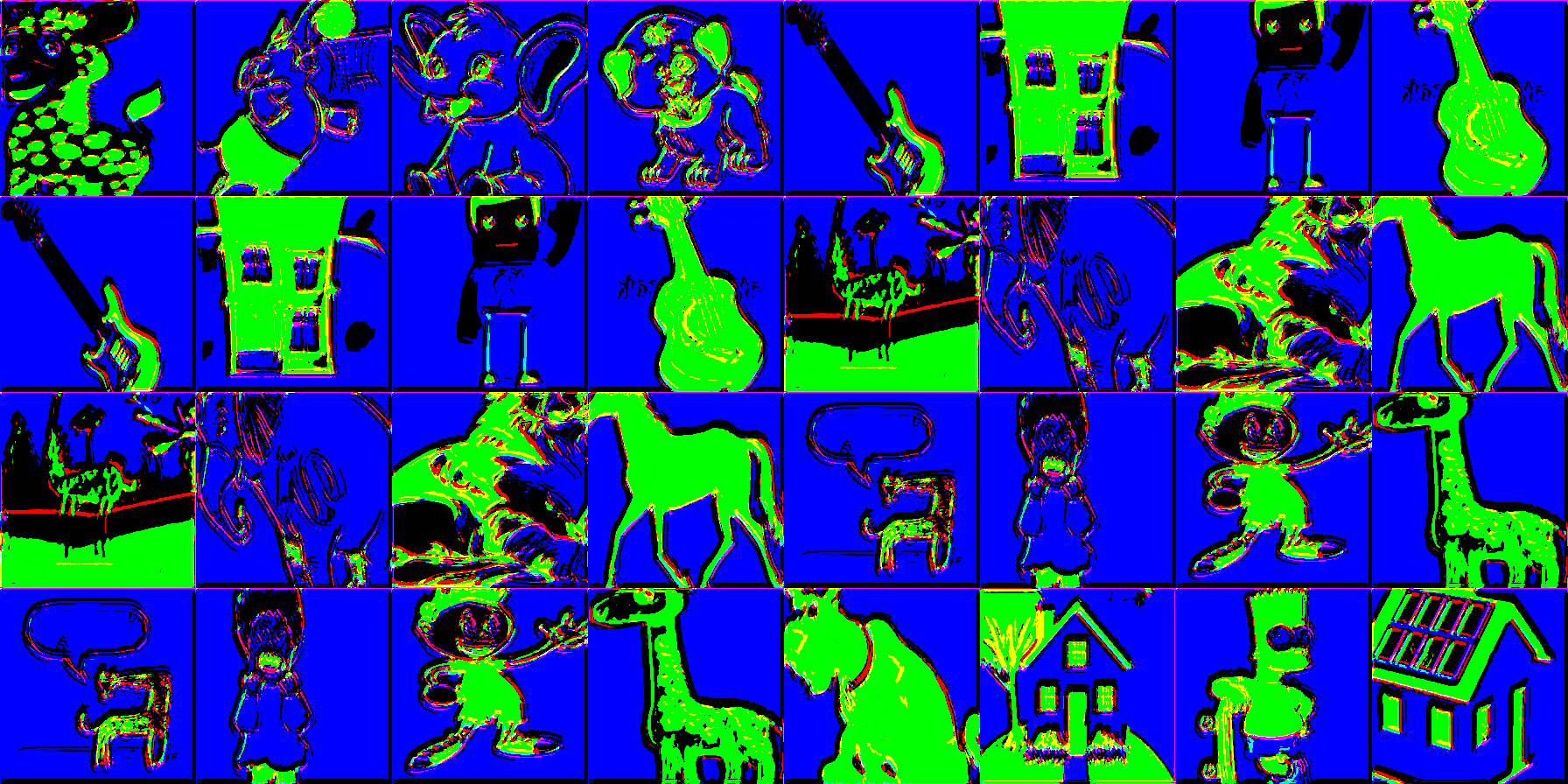} &
             \includegraphics{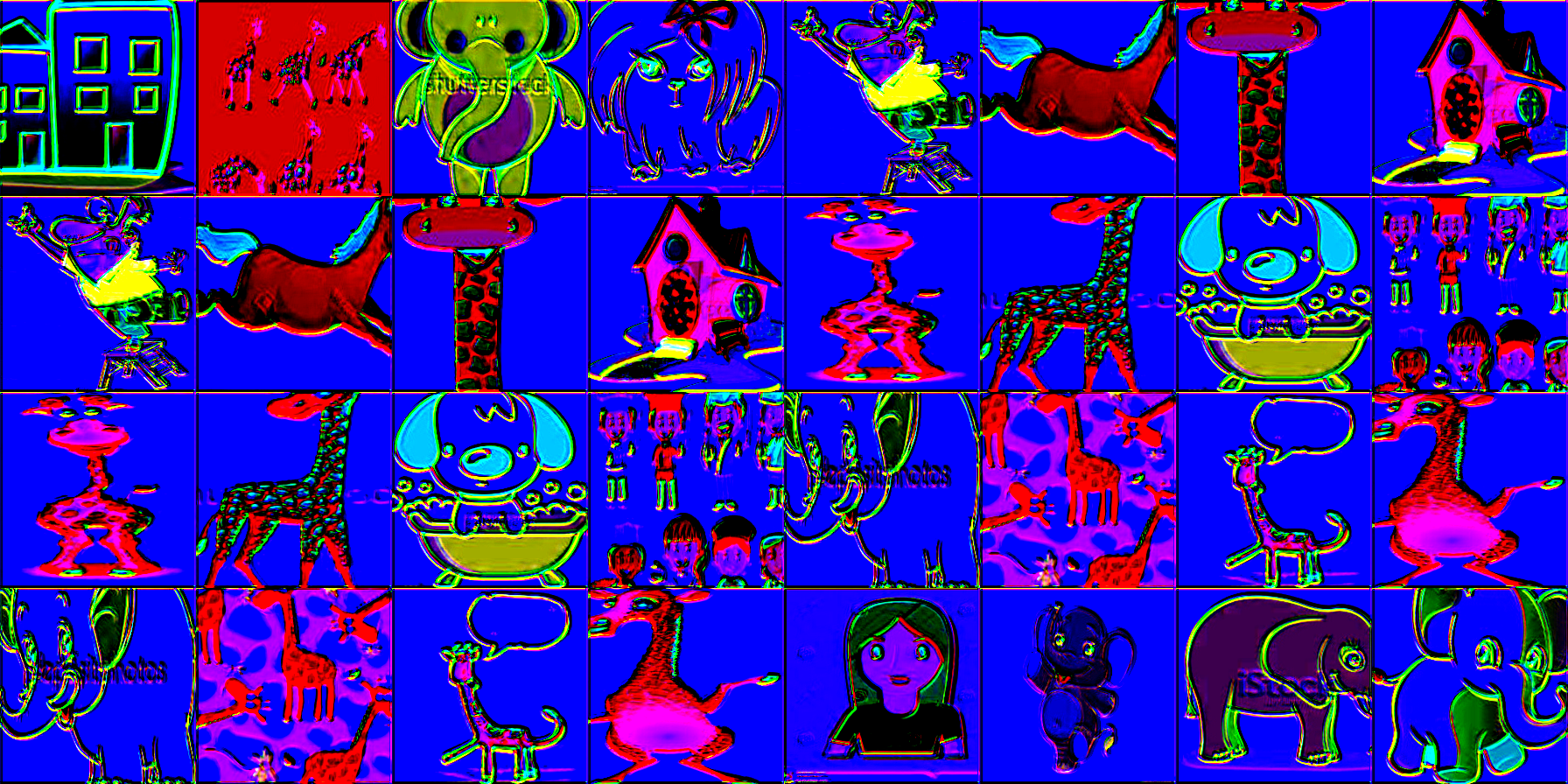} &
             \includegraphics{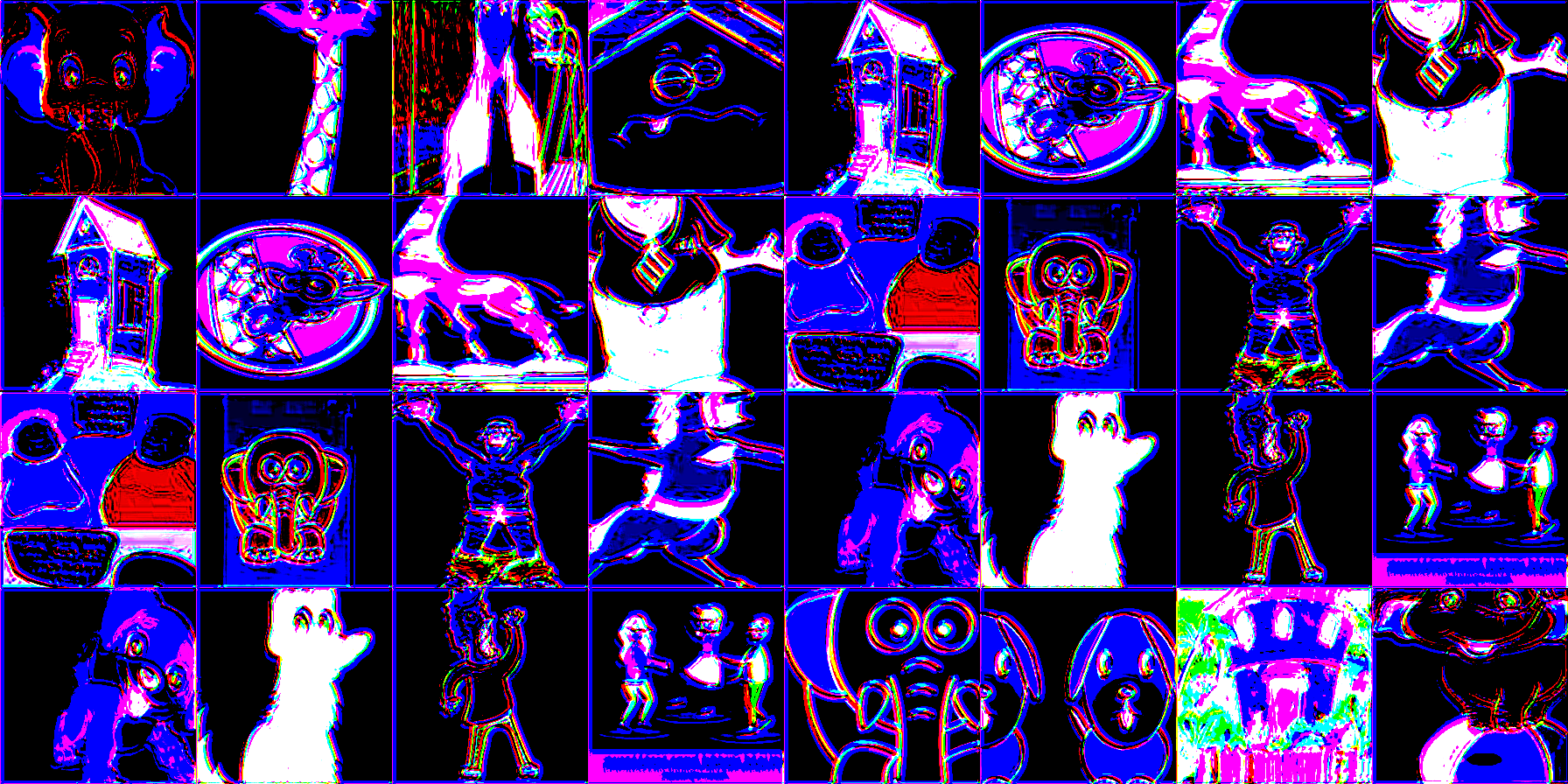} \\
        \end{tabular}
    \end{adjustbox}
    
    \caption{PACS: Images augmented by $\text{ABA}_{\text{3-layer}}$ with cartoon as source dataset.}
    \label{fig:sample_pacs_c}
\end{figure*}

\begin{figure*}
    \centering
    \begin{adjustbox}{width=\linewidth}
        \begin{tabular}{ccc}
             \includegraphics{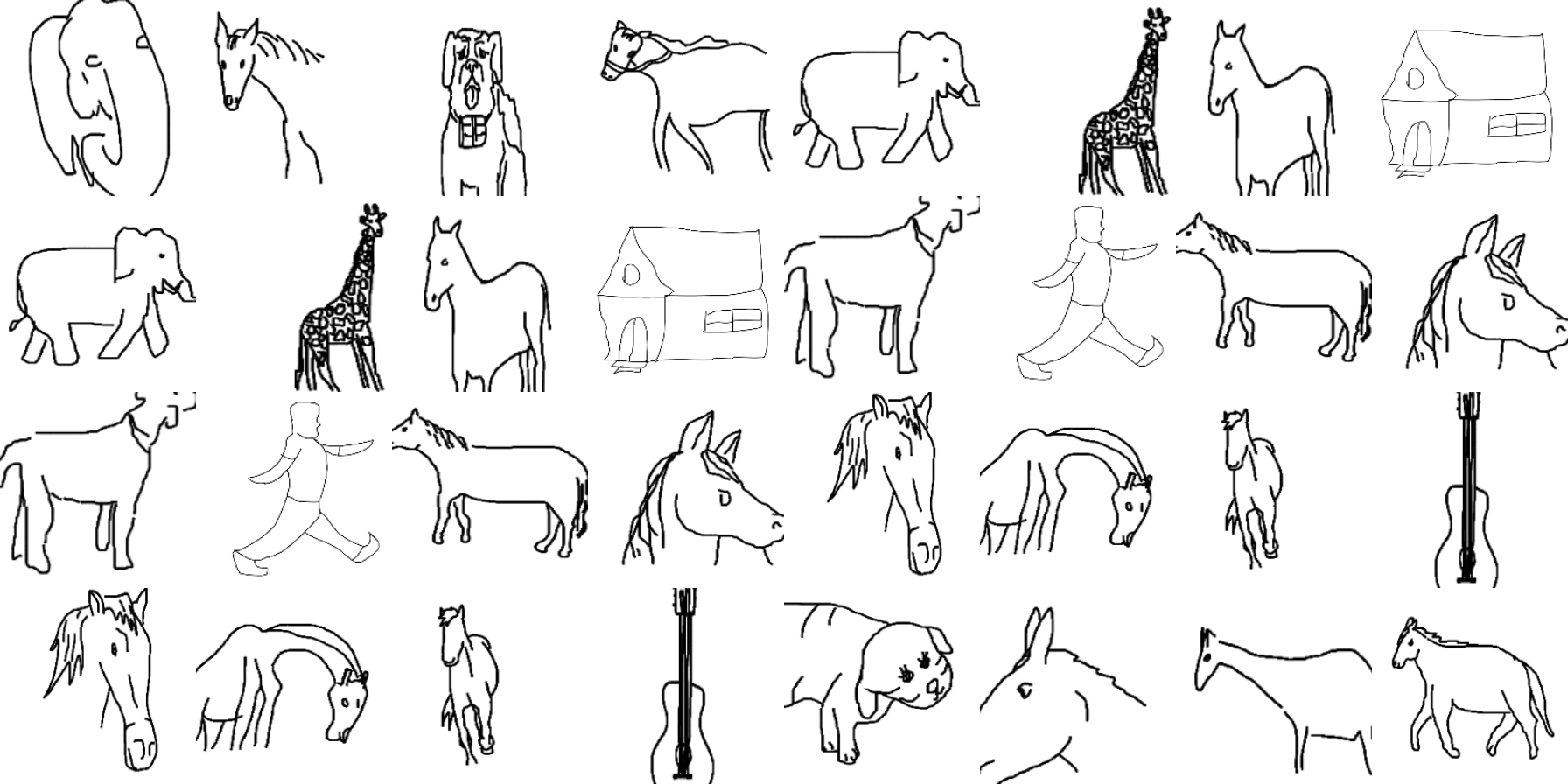} &
             \includegraphics{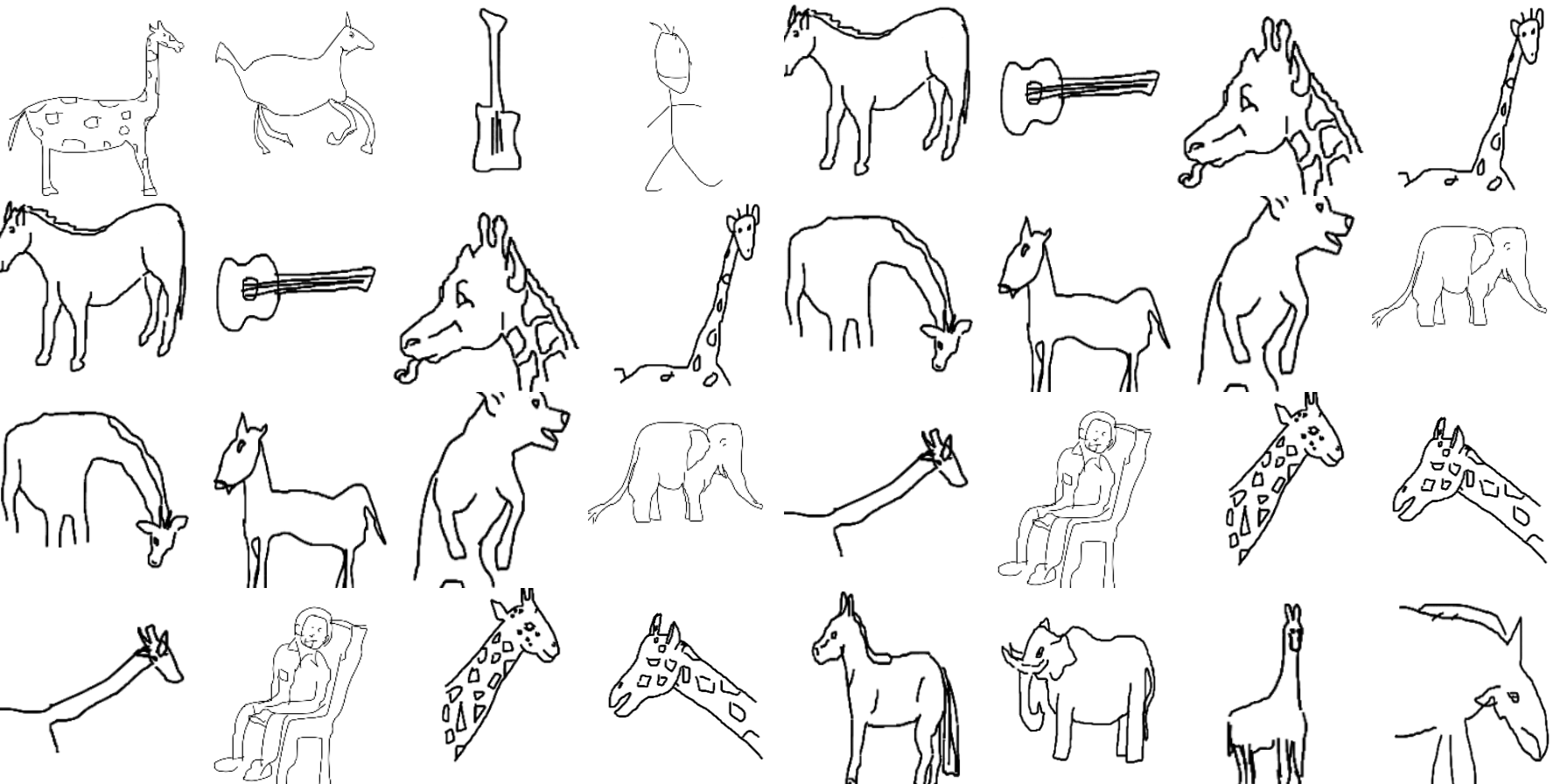} &
             \includegraphics{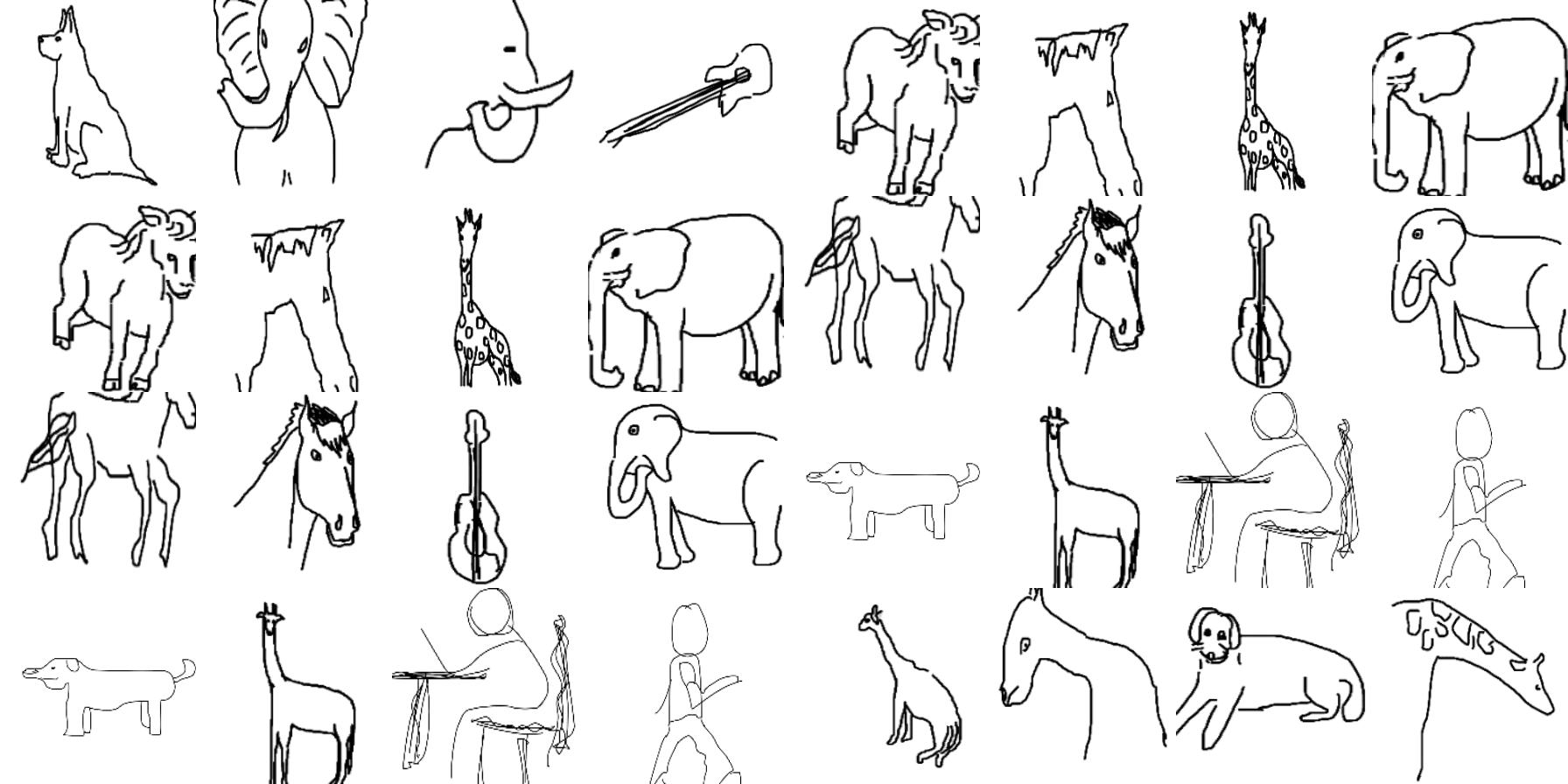} \\
             \includegraphics{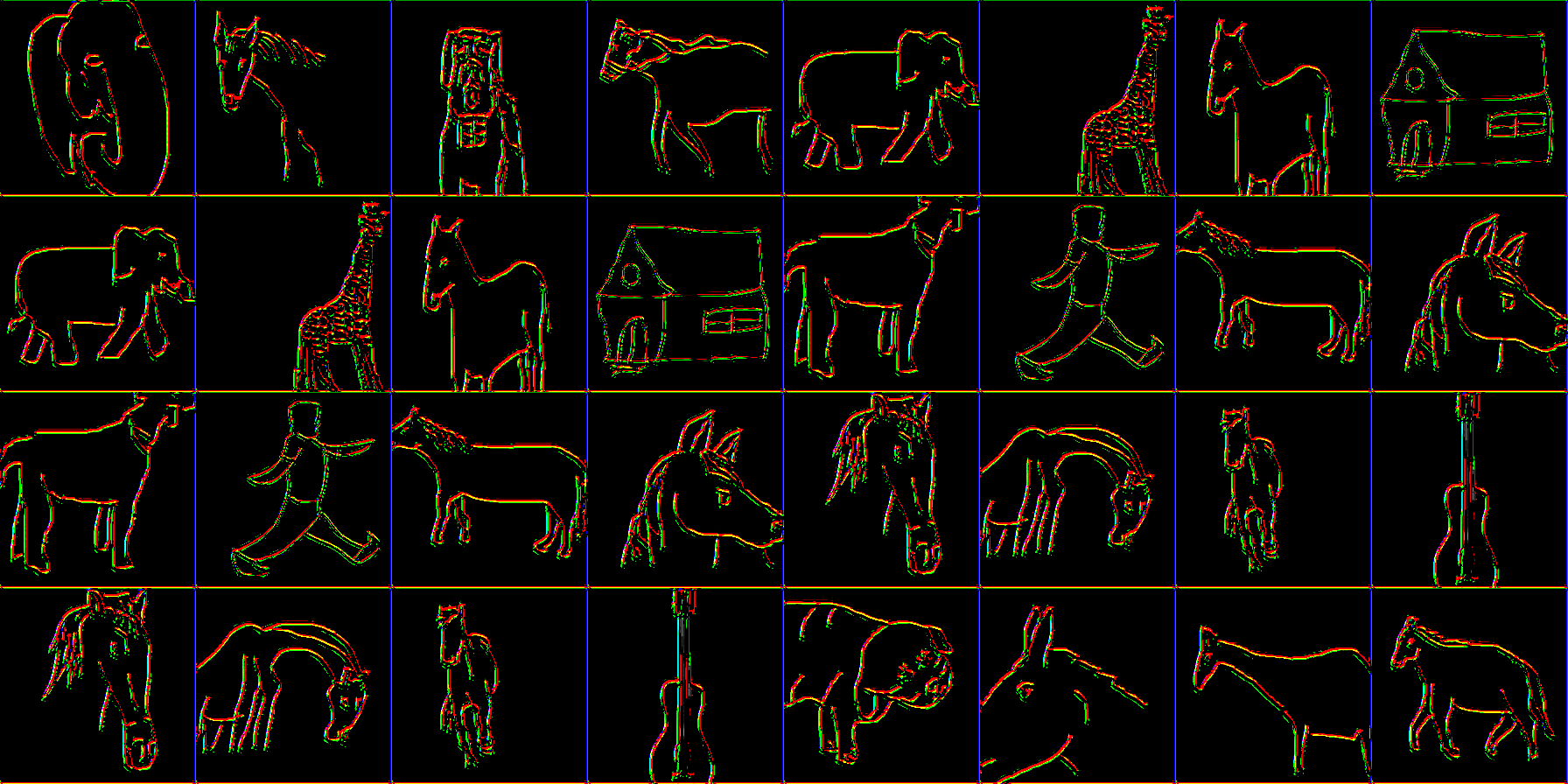} &
             \includegraphics{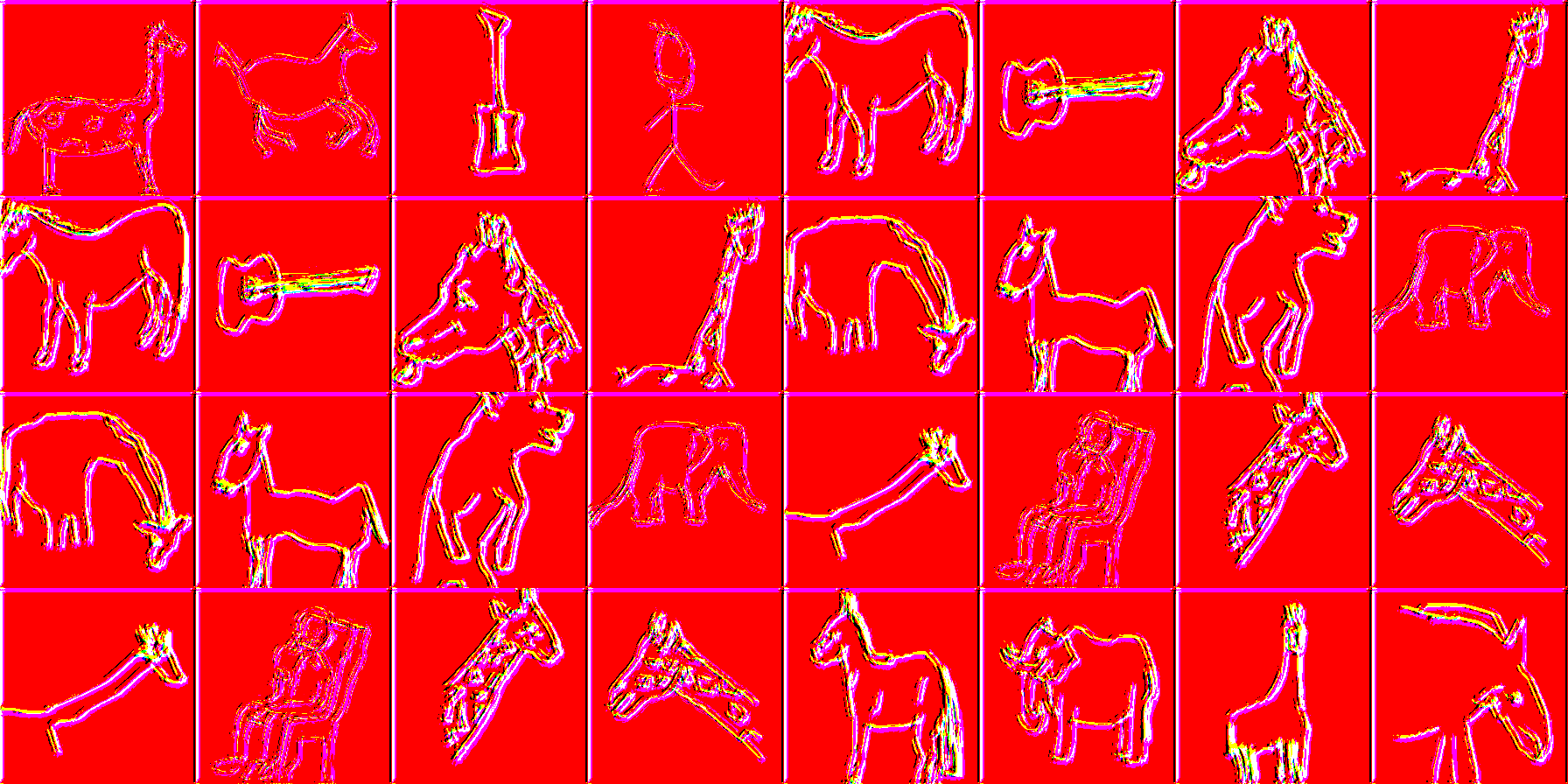} &
             \includegraphics{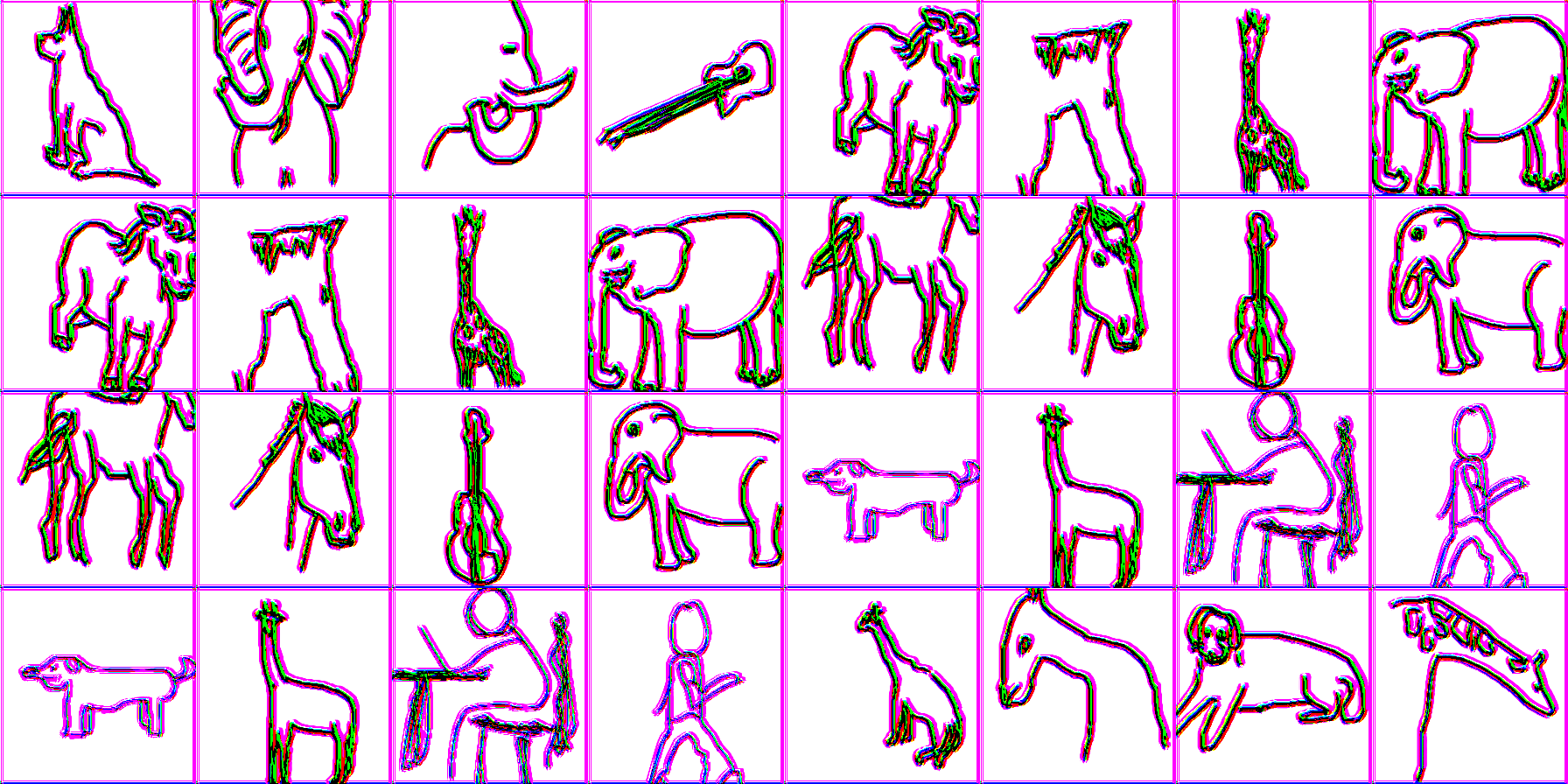} \\
        \end{tabular}
    \end{adjustbox}
    
    \caption{PACS: Images augmented by $\text{ABA}_{\text{3-layer}}$ with sketch as source dataset.}
    \label{fig:sample_pacs_s}
\end{figure*}

\begin{figure*}
    \centering
    \begin{adjustbox}{width=\linewidth}
        \begin{tabular}{ccc}
             \includegraphics{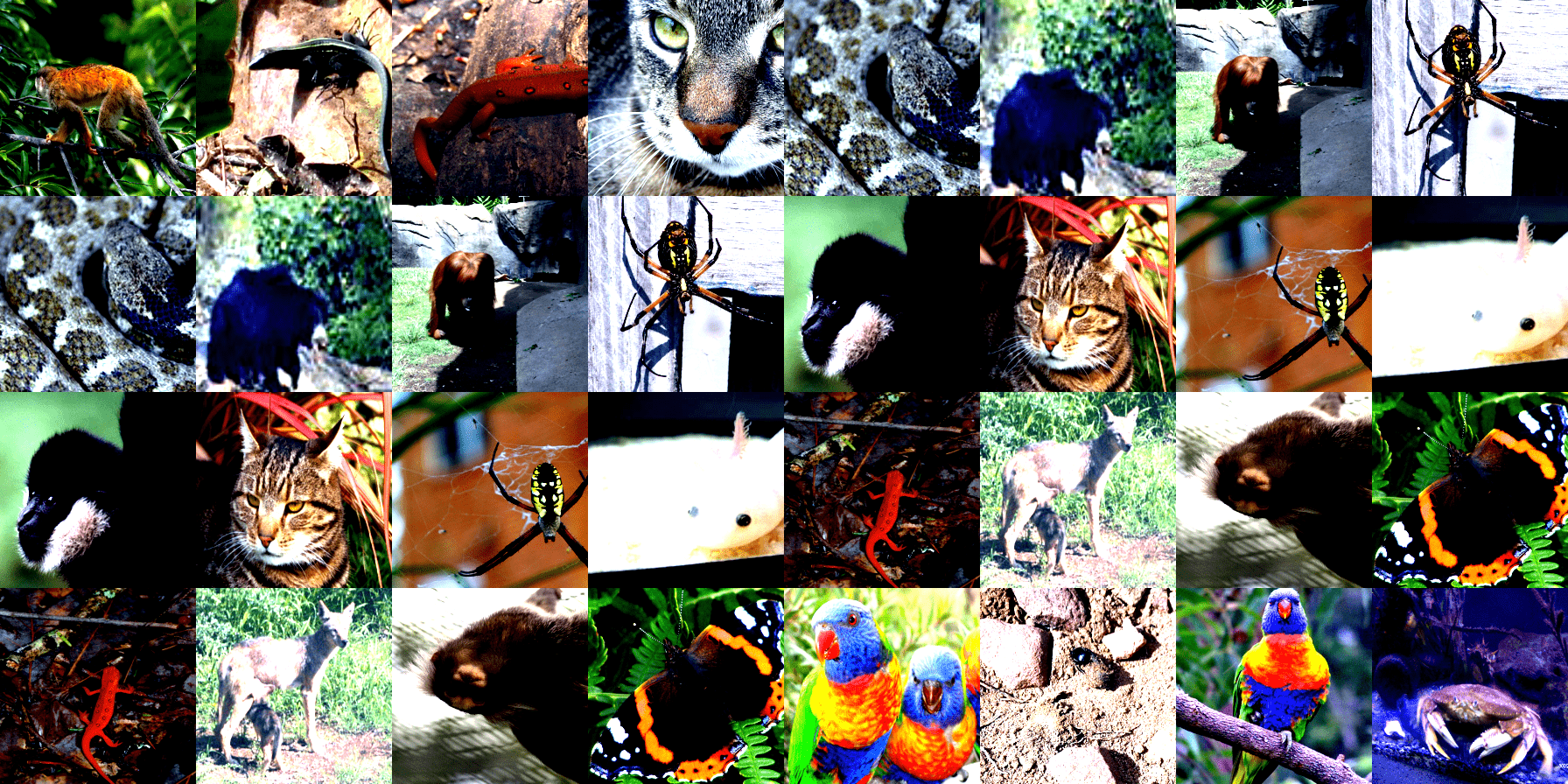} &
             \includegraphics{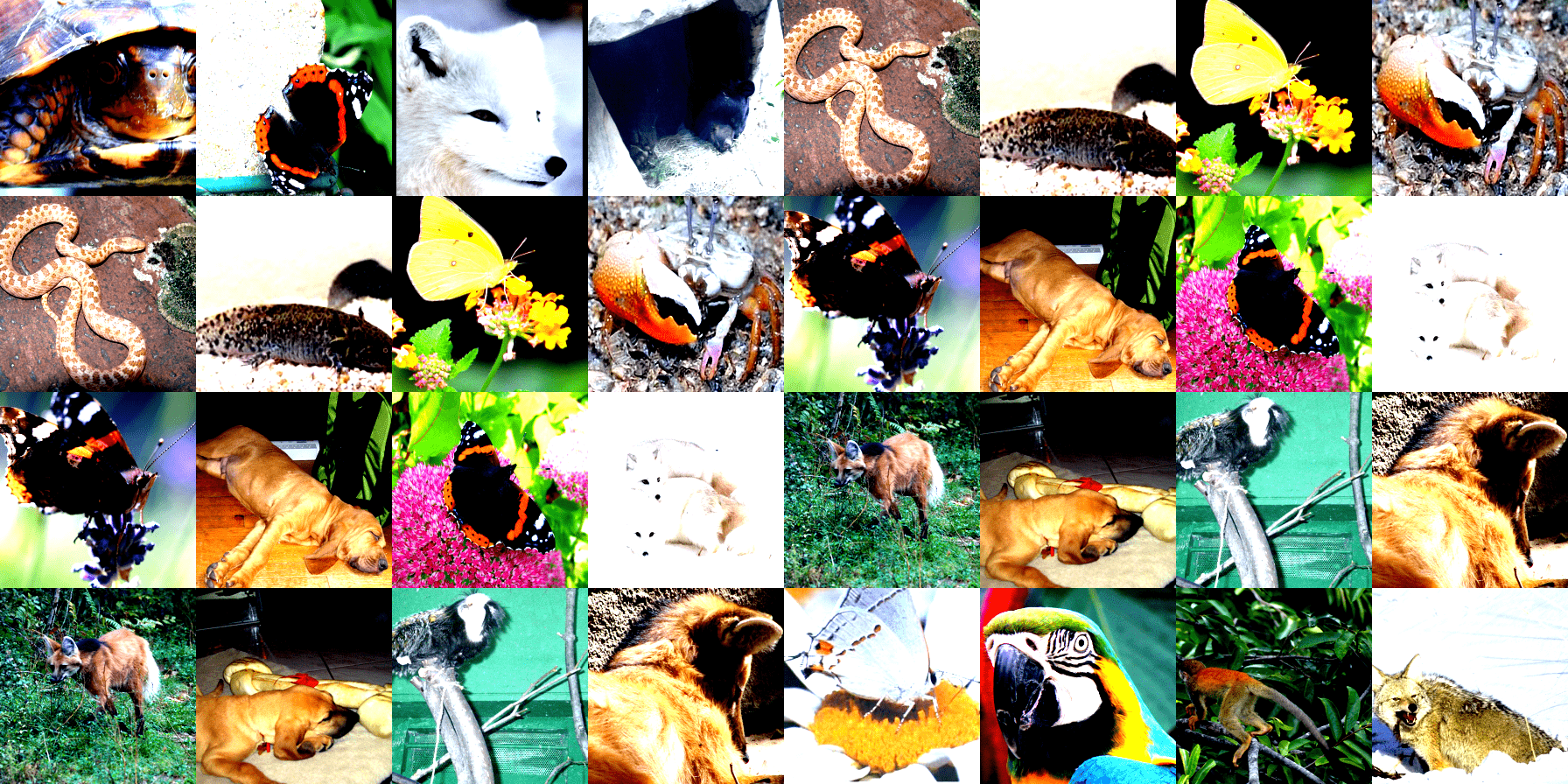} &
             \includegraphics{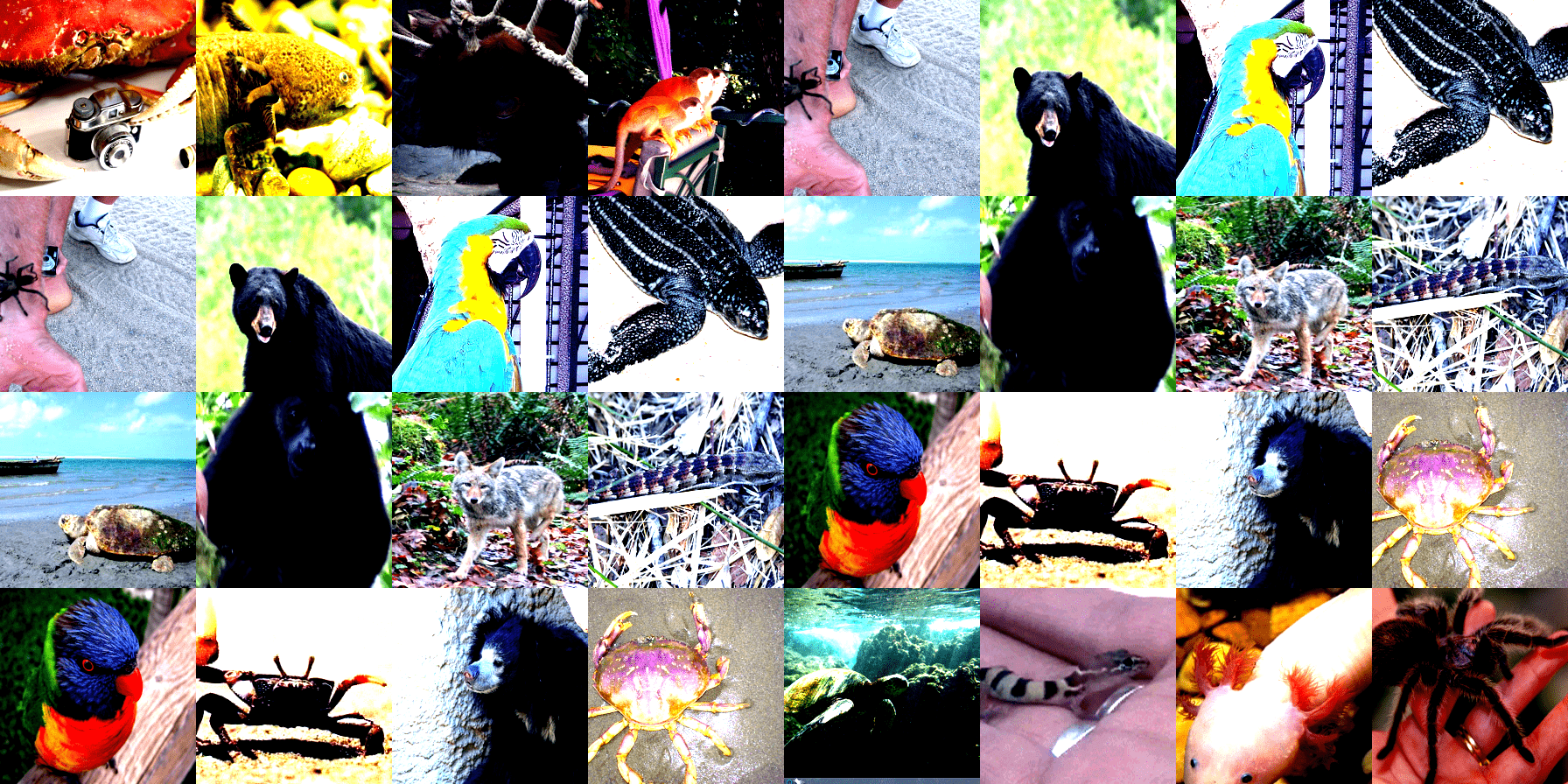} \\
             \includegraphics{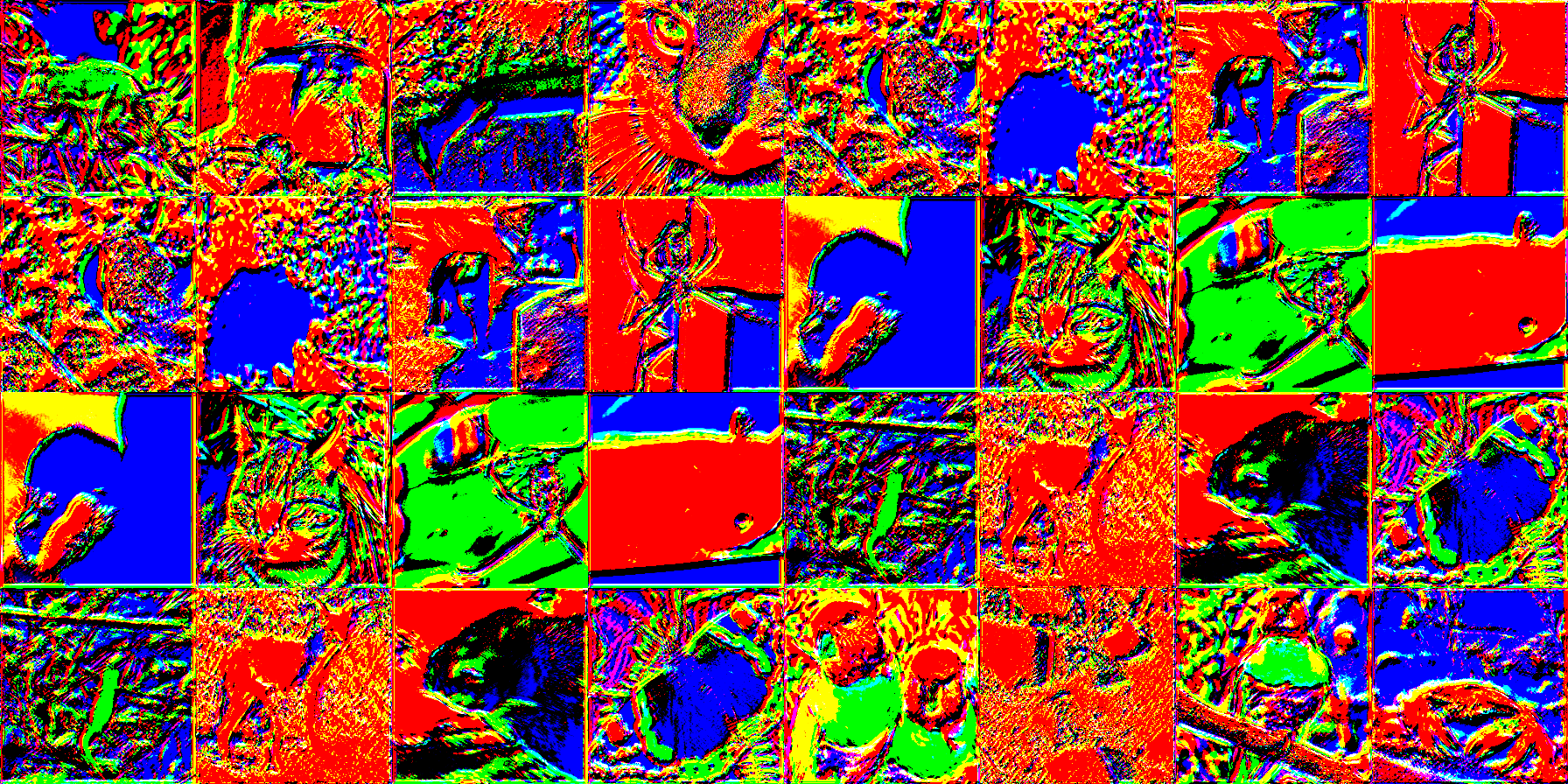} &
             \includegraphics{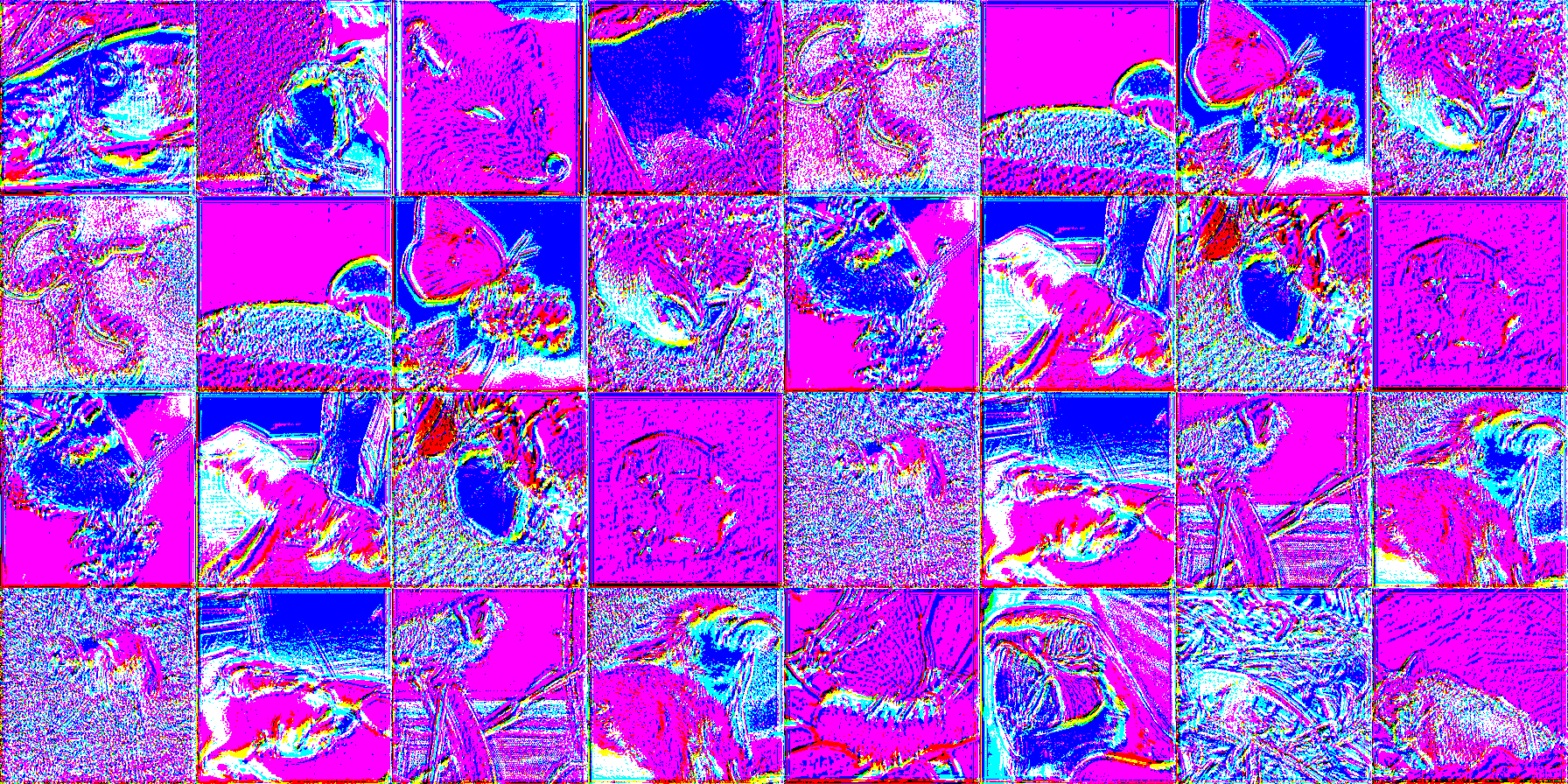} &
             \includegraphics{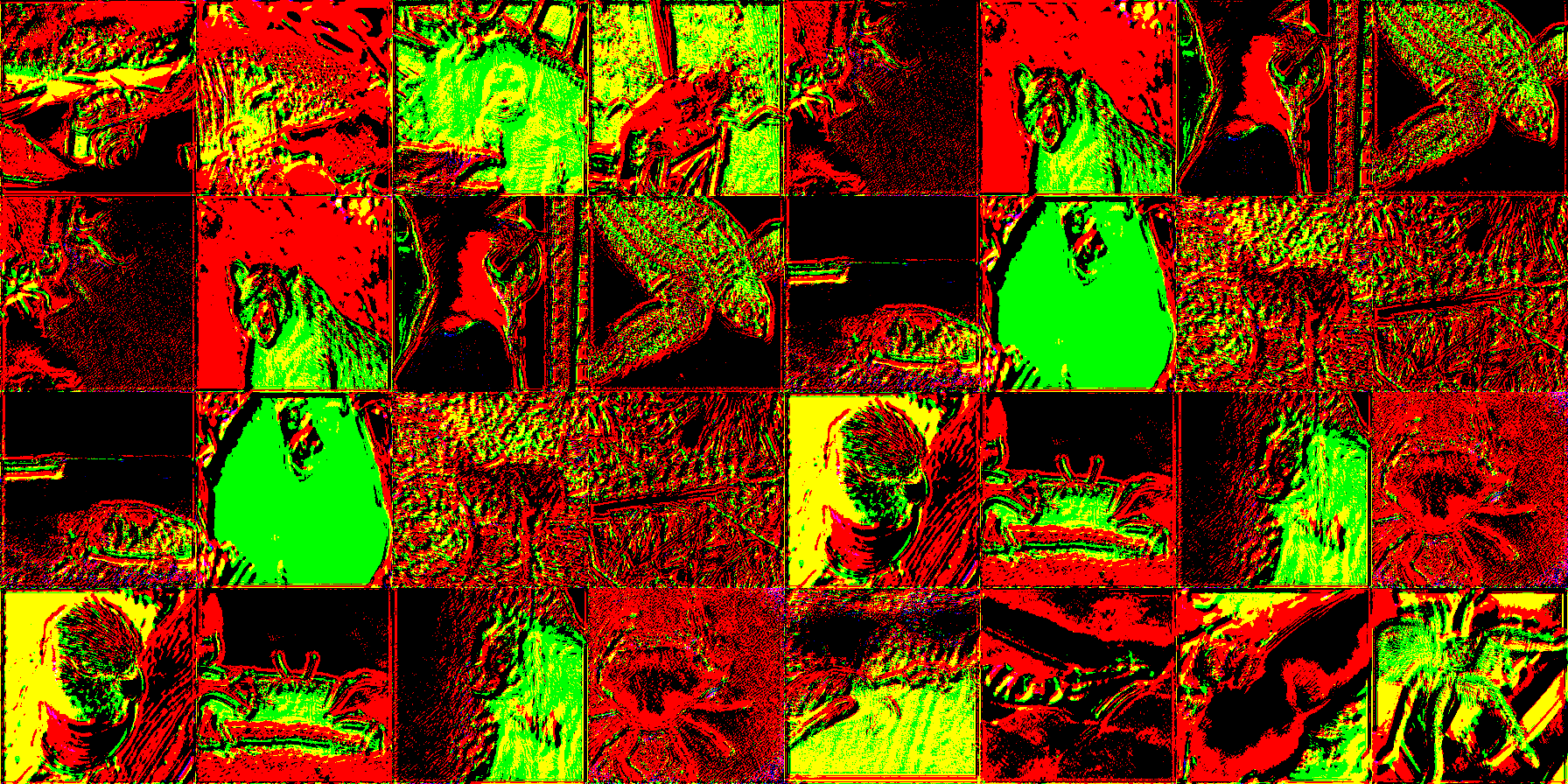} \\
        \end{tabular}
    \end{adjustbox}
    
    \caption{Living17: Images augmented by $\text{ABA}_{\text{5-layer}}$.}
    \label{fig:sample_living}
\end{figure*}

\begin{figure*}
    \centering
    \begin{adjustbox}{width=\linewidth}
        \begin{tabular}{ccc}
             \includegraphics{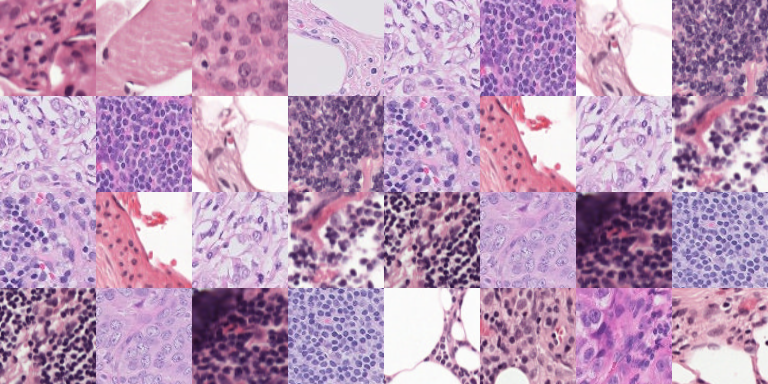} &
             \includegraphics{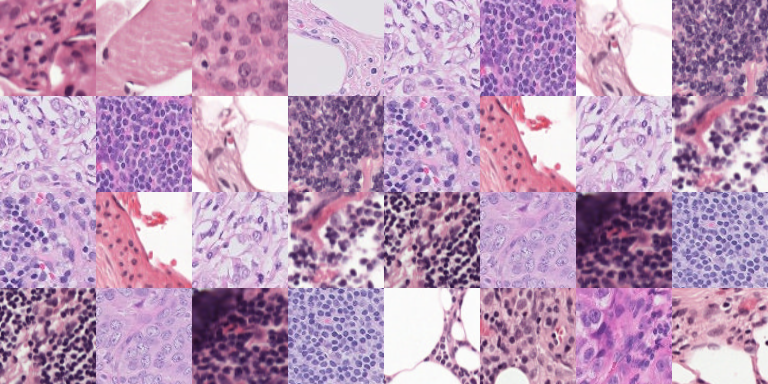} &
             \includegraphics{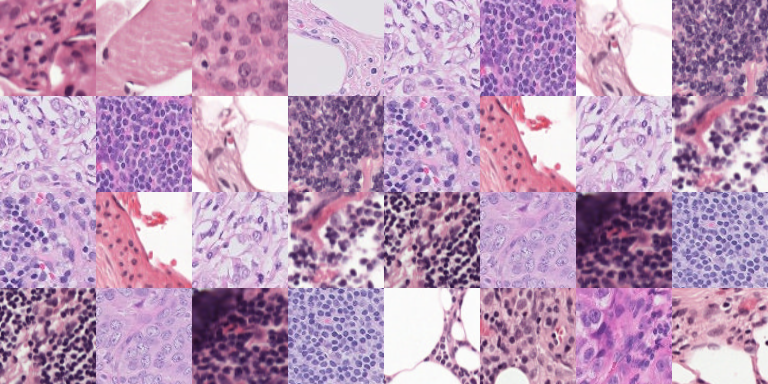} \\
             \includegraphics{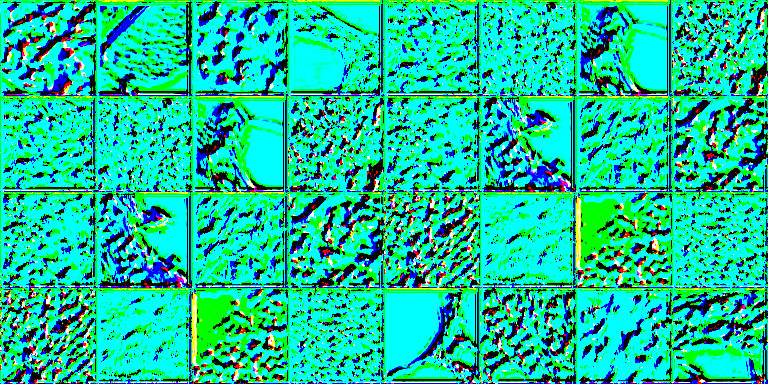} &
             \includegraphics{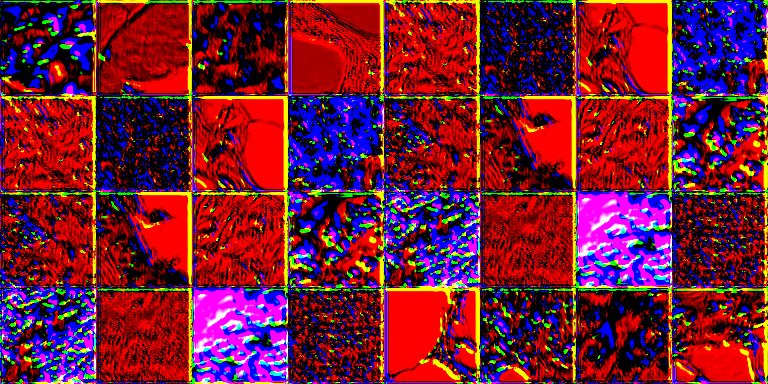} &
             \includegraphics{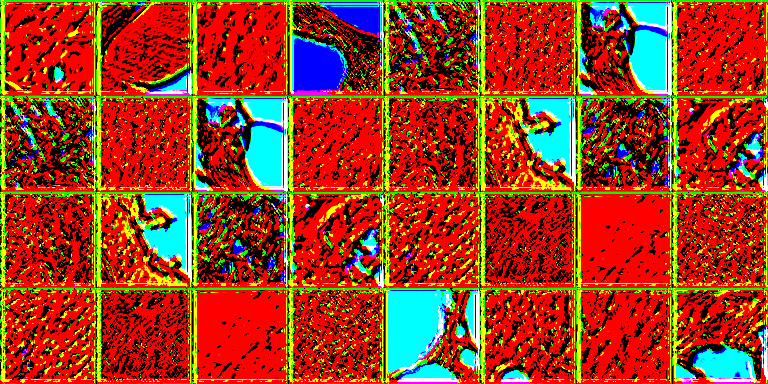} \\
        \end{tabular}
    \end{adjustbox}
    
    \caption{Camelyon17: Images augmented by $\text{ABA}_{\text{5-layer+RandConv}}$.}
    \label{fig:sample_wilds}
\end{figure*}

\begin{figure*}
    \centering
    \begin{adjustbox}{width=\linewidth}
    \begin{tabular}{ccc}
         \includegraphics{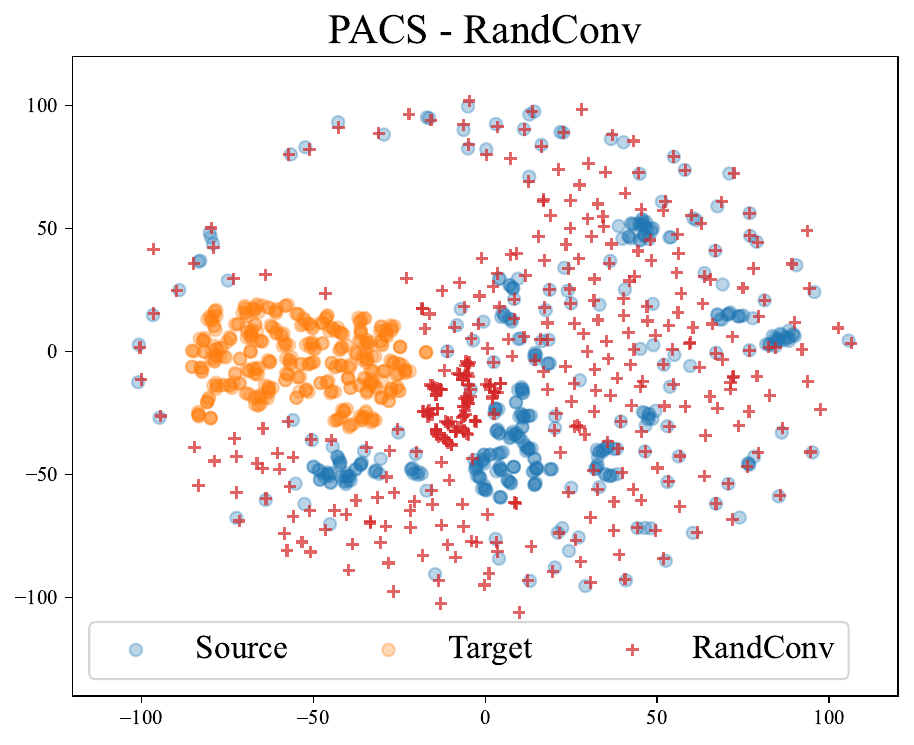} &
         \includegraphics{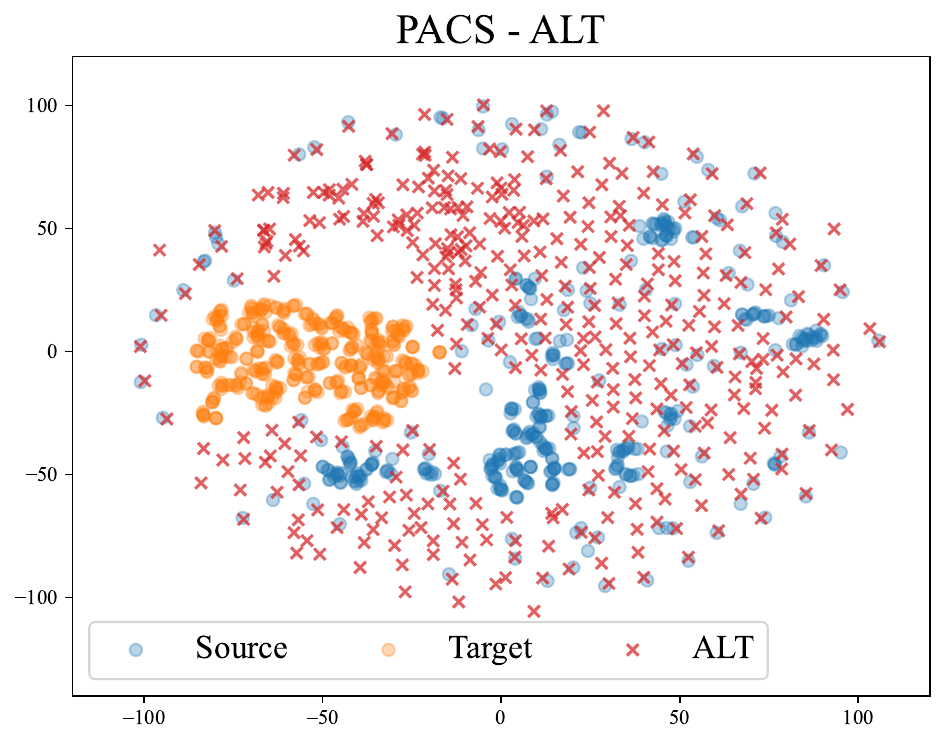} &
         \includegraphics{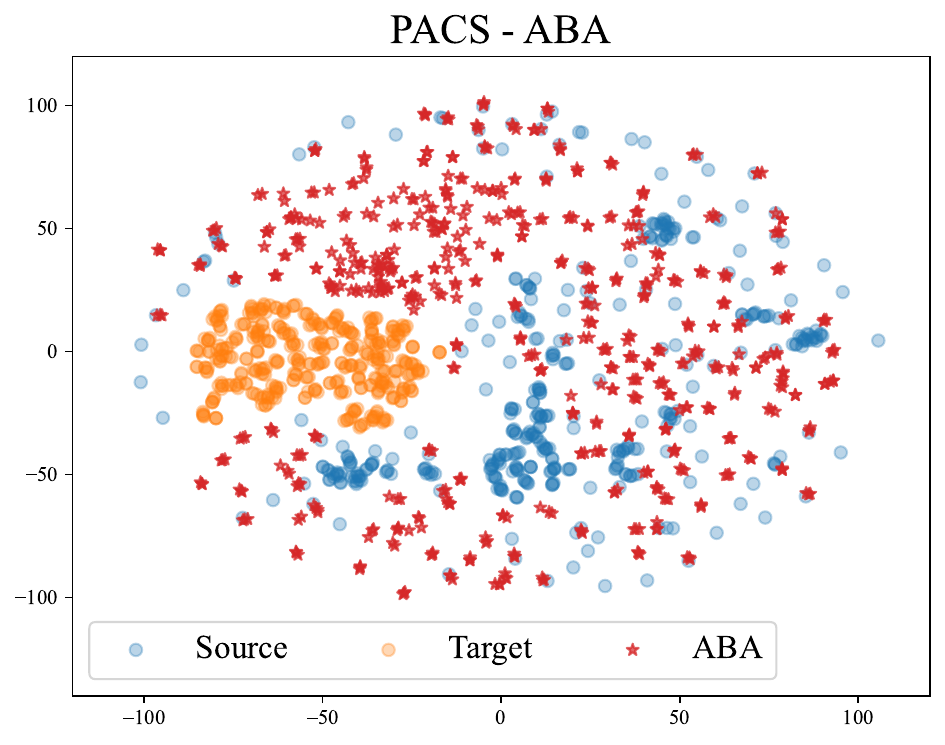} 
    \end{tabular}
    \end{adjustbox}
    \caption{TSNE plot for source domain, target domain and augmented image distribution by RandConv, ALT, ABA.}
    \label{fig:tsne_pacs}
\end{figure*}

\begin{figure*}
    \centering
    \begin{adjustbox}{width=\linewidth}
    \begin{tabular}{ccc}
         \includegraphics{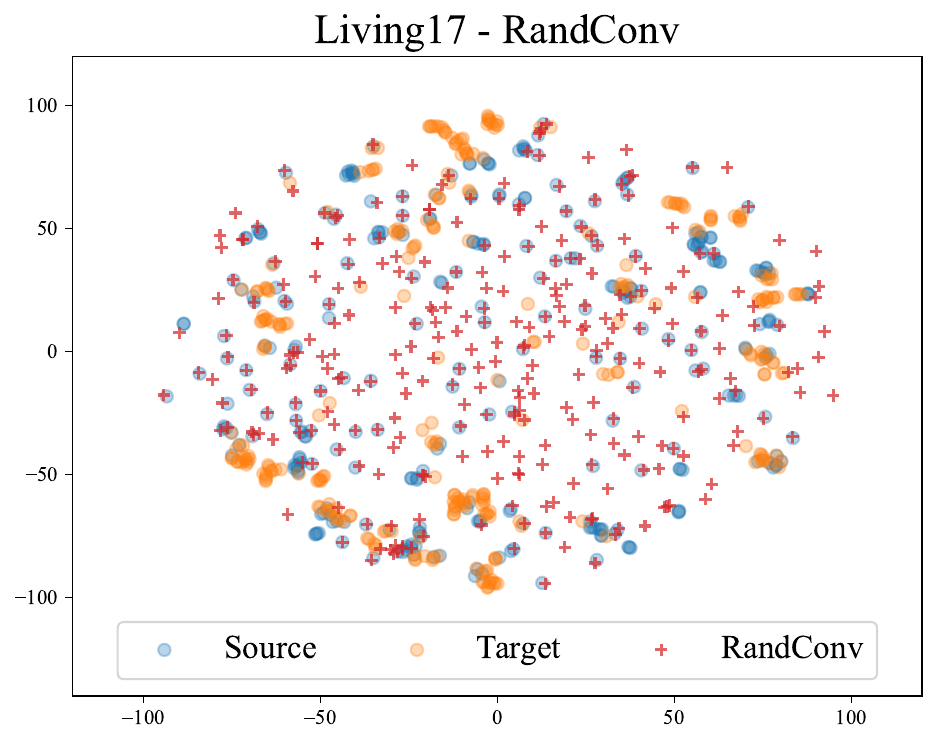} &
         \includegraphics{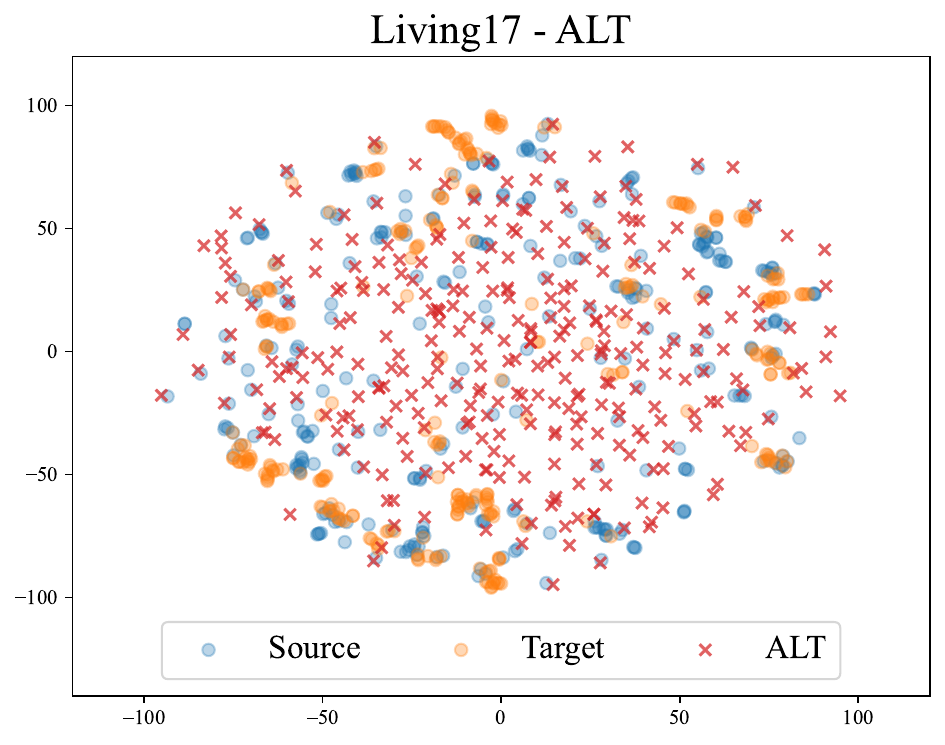} &
         \includegraphics{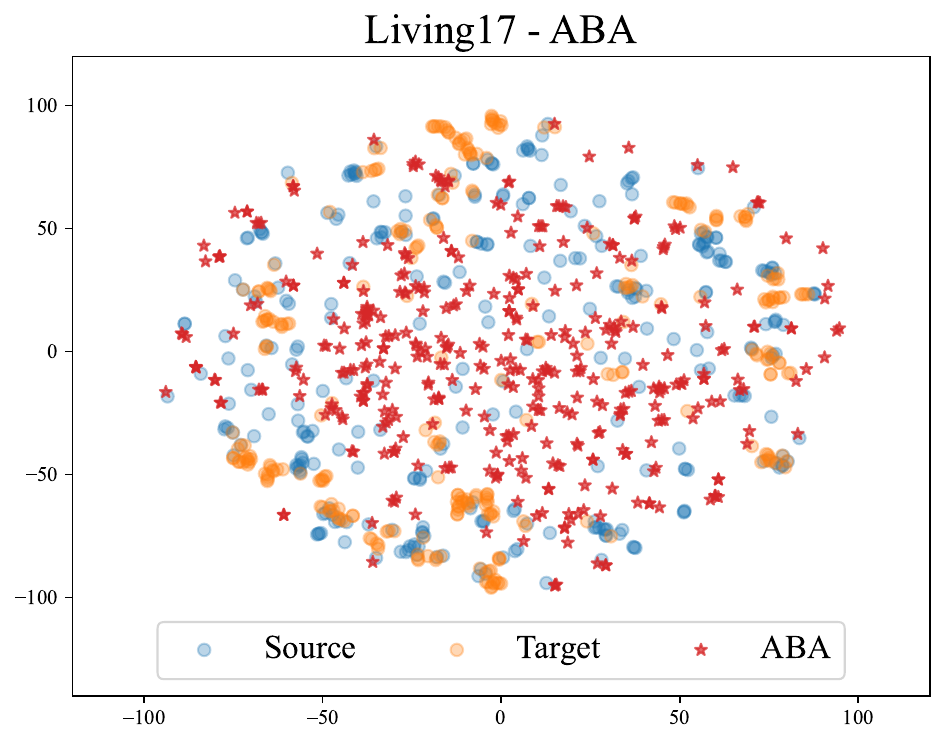} \\

    \end{tabular}
    \end{adjustbox}
    \caption{TSNE plot for source domain, target domain and augmented image distribution by RandConv, ALT, ABA.}
    \label{fig:tsne_living}
\end{figure*}

\begin{figure*}
    \centering
    \begin{adjustbox}{width=\linewidth}
    \begin{tabular}{ccc}
         \includegraphics{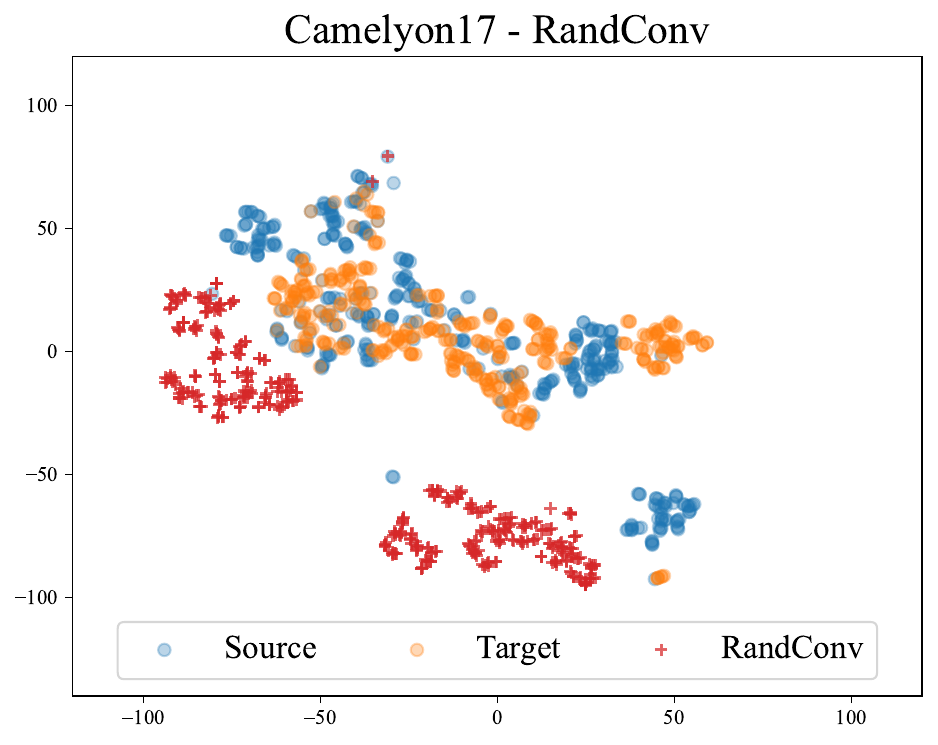} &
         \includegraphics{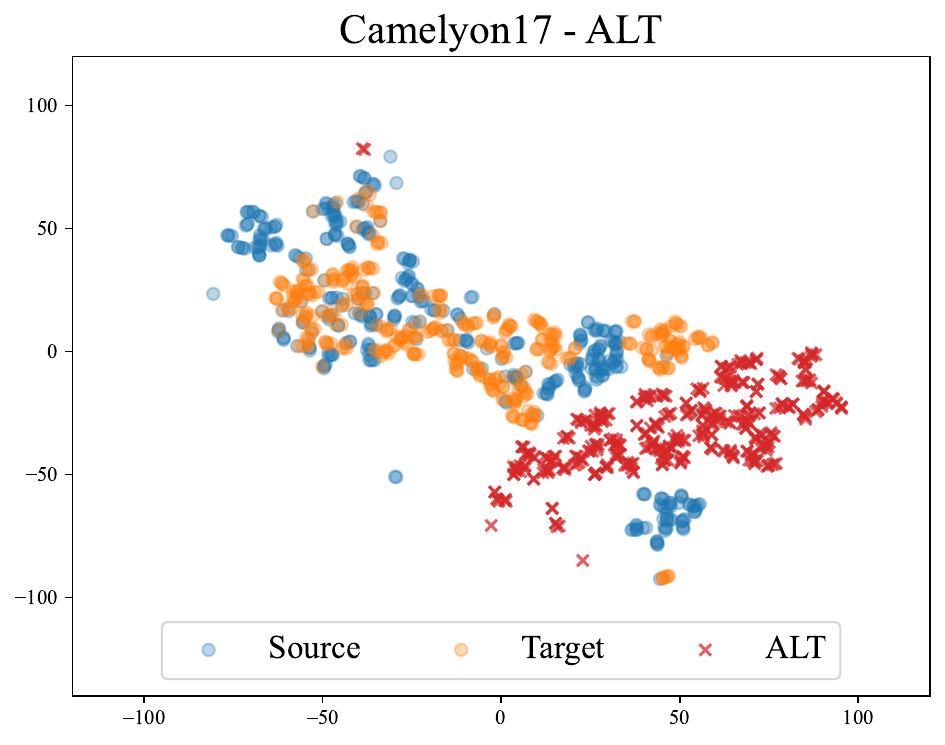} &
         \includegraphics{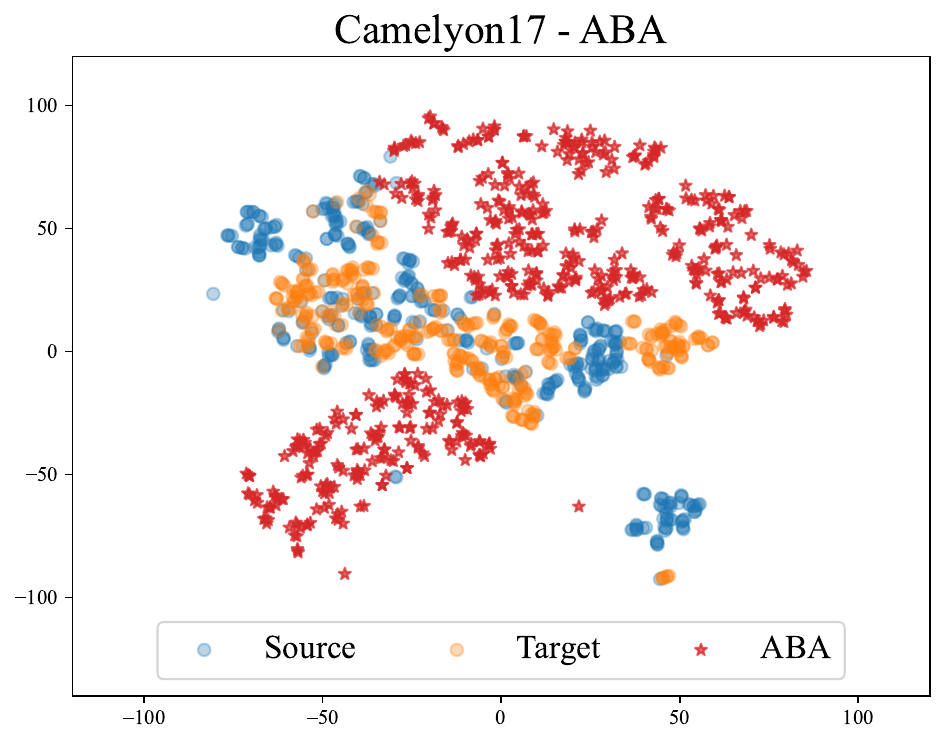} \\

    \end{tabular}
    \end{adjustbox}
    \caption{TSNE plot for source domain, target domain and augmented image distribution by RandConv, ALT, ABA.}
    \label{fig:tsne_wilds}
\end{figure*}

We show the additional qualitative results.
Figure~\ref{fig:sample_digit}, \ref{fig:sample_living}, \ref{fig:sample_wilds} show the augmented images by our method on the source domain of Digits dataset, Living17 dataset and Camelyon17 respectively. 
Figure~\ref{fig:sample_pacs_p}, \ref{fig:sample_pacs_a}, \ref{fig:sample_pacs_c}, \ref{fig:sample_pacs_s} show the augmented images as photo, art painting, cartoon and sketch as source domain.
The first row shows the input images, and the second row shows the augmented image by our methods.
Similarly, we illustrate the diversity introduced by our methods in comparison to the source distribution, the target distribution, and the distribution of RandConv, ALT augmentations, as shown in Figure~\ref{fig:tsne_pacs}, \ref{fig:tsne_living}, \ref{fig:tsne_wilds}.
For PACS dataset, we adopt 1-layer ALT and ABA, while for Living17 and Camelyon, we adopt 5-layer ALT and ABA.
For PACS dataset, the Photo domain is the source domain, and the Sketch domain is the target domain.
For Camelyon17 dataset, the combination of Hospital 1,2,3 is the source domain, and Hospital 4 is the target domain.
We observe that in both datasets, the distribution of our augmented method is spread widely across the tSNE space, consistent with the result from the Digits dataset in Figure~\ref{fig:tsne} of the main paper.

\begin{table*}[th]
    \centering
    \resizebox{\textwidth}{!}{
    \begin{tabular}{lllllll}
    \toprule
    \textbf{Method} & \textbf{Photo*} & \textbf{Art} & \textbf{Cartoon} & \textbf{Sketch} & \textbf{Target Avg.} & \textbf{All Avg.}\\
    \midrule
    ERM & - & 64.1 &  23.6 & 29.1 & 38.9 & - \\
    Augmix & 99.532 \scriptsize(0.438)  & 68.633 \scriptsize(0.950) & 33.788 \scriptsize(1.205) & 36.304 \scriptsize(2.801) & 46.242 \scriptsize(1.122) & 59.564 \scriptsize(0.930)\\ 
    \hline
    \color{purple} \scriptsize 1-layer convolutional-based augmentations\\
    RandConv & 96.407 \scriptsize(0.757)  & 61.309 \scriptsize(2.316) & 37.577 \scriptsize(2.257) & 50.463 \scriptsize(9.018) & 49.783 \scriptsize(4.255) & 61.439 \scriptsize(3.217)\\
    $\text{ALT}_{\text{1-layer}}$ & 99.298 \scriptsize(0.438) & 70.000 \scriptsize(0.399) & 39.761 \scriptsize(2.443) & 42.728 \scriptsize(5.928) & 50.830 \scriptsize(2.127) & 62.947 \scriptsize(1.689)\\
    $\text{ALT}_{\text{1-layer+RandConv}}$ & 99.298 \scriptsize(0.438) & 69.941 \scriptsize(0.213) & 38.874 \scriptsize(1.686) & 47.905 \scriptsize(0.924) & 52.240 \scriptsize(0.820) & 64.005 \scriptsize(0.660) \\
    $\text{ABA}_{\text{1-layer}}$ &  99.123 \scriptsize(0.292) & \bfseries 70.667 \scriptsize(0.517) & \bfseries 42.129 \scriptsize(1.813) & \bfseries 50.681 \scriptsize(3.094) & \bfseries 54.492 \scriptsize(1.352) & \bfseries 65.650 \scriptsize(1.047)\\
    $\text{ABA}_{\text{1-layer+RandConv}}$ & 99.064 \scriptsize(0.286) & 70.303 \scriptsize(0.843) & 40.546 \scriptsize(1.612) & 46.114 \scriptsize(3.697) & 52.321 \scriptsize(1.823) & 64.007 \scriptsize(1.375) \\
    \hline
    \color{purple} \scriptsize 3-layer convolutional-based augmentations\\
    $\text{ABA}_{\text{3-layer}}$ & 98.713 \scriptsize(0.234) & 67.041 \scriptsize(0.687) & \bfseries 46.698 \scriptsize(1.267) & \bfseries 62.825 \scriptsize(1.625) & \bfseries 58.855 \scriptsize(0.828) & \bfseries 68.819 \scriptsize(0.661)\\
    $\text{ABA}_{\text{3-layer+RandConv}}$ &99.061 \scriptsize(0.610) & \bfseries 68.353 \scriptsize(0.842) &44.830 \scriptsize(2.606) &57.682 \scriptsize(2.520) &56.958 \scriptsize(0.801) &67.482 \scriptsize(0.491) \\
    \hline
    \color{purple} \scriptsize 5-layer convolutional-based augmentations\\
    $\text{ALT}_{\text{5-layer}}$ & 99.064 \scriptsize(0.286) & \bfseries 68.770 \scriptsize(0.932) & 43.387 \scriptsize(1.142)  & 50.832 \scriptsize(2.937) & 54.330 \scriptsize(1.078) & 65.513 \scriptsize(0.757) \\
    $\text{ALT}_{\text{5-layer+RandConv}}$ & 98.947 \scriptsize(0.234) & 68.740 \scriptsize(0.702) & 40.828 \scriptsize(2.517) & 56.024 \scriptsize(2.009) & 55.197 \scriptsize(0.498) & 66.135 \scriptsize(0.330)\\
    $\text{ALT}_{\text{5-layer+Augmix}}$ & 99.298 \scriptsize(0.438) & 68.506 \scriptsize(0.836) & 43.507 \scriptsize(2.615) & 53.271 \scriptsize(4.149) & 55.094 \scriptsize(1.876) & 66.145 \scriptsize(1.387)\\
    $\text{ABA}_{\text{5-layer}}$ &98.480 \scriptsize(0.286) &67.197 \scriptsize(0.485) & \bfseries 47.381 \scriptsize(2.023) & \bfseries 62.566 \scriptsize(2.638) & \bfseries 59.048 \scriptsize(1.429) & \bfseries 68.906 \scriptsize(1.084) \\
    $\text{ABA}_{\text{5-layer+RandConv}}$&98.830 \scriptsize(0.002) &67.881 \scriptsize(0.935) &45.990 \scriptsize(3.440) &58.916 \scriptsize(2.640) &57.595 \scriptsize(1.263) &67.904 \scriptsize(0.947) \\
    $\text{ABA}_{\text{5-layer+Augmix}}$&98.830 \scriptsize(0.477) &66.243 \scriptsize(0.403) &46.573 \scriptsize(1.660) &60.813 \scriptsize(1.209) &57.876 \scriptsize(0.221) &68.115 \scriptsize(0.115) \\
    \bottomrule
    \end{tabular}
    }
    \caption{SSDG performance on PACS for the P $\rightarrow$ ACS. $^*$Source Domain. \textbf{bold}: best result.}
    \label{tab:pacs_p}
\end{table*}

\begin{table*}
    \centering
    \resizebox{\textwidth}{!}{
    \begin{tabular}{lllllll}
    \toprule
    \textbf{Method} & \textbf{Photo} & \textbf{Art*} & \textbf{Cartoon} & \textbf{Sketch} & \textbf{Target Avg.} & \textbf{All Avg.} \\
    \midrule
    ERM & 95.2 & -  & 62.3 & 49.0 & 68.8 & - \\
    Augmix & 95.317 \scriptsize(0.422) & 93.077 \scriptsize(1.276) & 64.061 \scriptsize(0.361) & 55.027 \scriptsize(2.195) & 71.469 \scriptsize(0.637) & 76.871 \scriptsize(0.581) \\
    \hline
    \color{purple} \scriptsize 1-layer convolutional-based augmentations\\
    RandConv & 87.281 \scriptsize(0.796) & 85.437 \scriptsize(0.532) & 61.143 \scriptsize(2.752) & 60.519 \scriptsize(4.050) & 69.648 \scriptsize(2.152) & 73.595 \scriptsize(1.582) \\
    $\text{ALT}_{\text{1-layer}}$ & \bfseries 95.365 \scriptsize(0.285) & 92.500 \scriptsize(1.239) &64.462 \scriptsize(1.018) &61.792 \scriptsize(3.433) &73.873 \scriptsize(1.314) &78.530 \scriptsize(1.043)\\
    $\text{ALT}_{\text{1-layer+RandConv}}$ &94.766 \scriptsize(0.409) &91.442 \scriptsize(1.532) &64.480 \scriptsize(1.083) &61.140 \scriptsize(3.971) &73.462 \scriptsize(1.288) &77.957 \scriptsize(0.836)  \\
    $\text{ABA}_{\text{1-layer}}$ &94.311 \scriptsize(0.273) &91.442 \scriptsize(1.439) &  \bfseries 65.239 \scriptsize(1.436) &67.228 \scriptsize(3.555) &75.593 \scriptsize(1.563) &79.555 \scriptsize(1.289) \\
    $\text{ABA}_{\text{1-layer+RandConv}}$ &94.515 \scriptsize(0.449) &91.346 \scriptsize(1.426) &65.162 \scriptsize(1.258) & \bfseries 67.661 \scriptsize(3.889) & \bfseries 75.779 \scriptsize(1.636) & \bfseries 79.671 \scriptsize(1.123) \\
    \hline
    \color{purple} \scriptsize 3-layer convolutional-based augmentations\\
    $\text{ABA}_{\text{3-layer}}$ &92.635 \scriptsize(0.386) &91.346 \scriptsize(1.325) & \bfseries 64.625 \scriptsize(0.576) &\bfseries 68.745 \scriptsize(1.789) &75.335 \scriptsize(0.889) &79.338 \scriptsize(0.976) \\
    $\text{ABA}_{\text{3-layer+RandConv}}$ &\bfseries 93.592 \scriptsize(0.36) & 91.443 \scriptsize(2.050) &64.601 \scriptsize(0.710) &67.849 \scriptsize(1.485) &\bfseries 75.341 \scriptsize(0.520) &\bfseries 79.372 \scriptsize(0.713) \\
    \hline
    \color{purple} \scriptsize 5-layer convolutional-based augmentations\\
    $\text{ALT}_{\text{5-layer}}$ & \bfseries 94.934 \scriptsize(0.269) & 91.058 \scriptsize(0.720) & 63.524 \scriptsize(1.821) & 63.813 \scriptsize(2.249) & 74.090 \scriptsize(1.086) & 78.332 \scriptsize(0.845) \\
    $\text{ALT}_{\text{5-layer+RandConv}}$ & 93.593 \scriptsize(0.328) & 92.596 \scriptsize(1.036) & 64.044 \scriptsize(0.635) & 65.991 \scriptsize(1.130) & 74.543 \scriptsize(0.537) & 79.056 \scriptsize(0.609) \\
    $\text{ALT}_{\text{5-layer+Augmix}}$ & 93.174 \scriptsize(0.437) & 91.442 \scriptsize(0.638) &\bfseries 65.683 \scriptsize(1.656) &\bfseries 68.226 \scriptsize(2.453) &\bfseries 75.694 \scriptsize(1.214) & 79.631 \scriptsize(0.856)\\
    $\text{ABA}_{\text{5-layer}}$ &92.108 \scriptsize(0.442) &91.827 \scriptsize(1.426) &64.872 \scriptsize(0.994) &67.172 \scriptsize(1.135) &74.717 \scriptsize(0.757) &78.995 \scriptsize(0.518)  \\
    $\text{ABA}_{\text{5-layer+RandConv}}$ &93.210 \scriptsize(0.402) &92.212 \scriptsize(0.981) &  65.580 \scriptsize(1.326) & 68.068 \scriptsize(2.521) & 75.619 \scriptsize(1.015) &\bfseries 79.767 \scriptsize(0.695) \\
    $\text{ABA}_{\text{5-layer+Augmix}}$ &92.255 \scriptsize(0.539) &90.545 \scriptsize(1.379) & 64.022 \scriptsize(0.625) & 67.829 \scriptsize(2.017) & 74.702 \scriptsize(0.968) & 78.663 \scriptsize(0.834) \\
    \bottomrule
    \end{tabular}
    }
    \caption{SSDG performance on PACS for the A $\rightarrow$ PCS. $^*$Source Domain. \textbf{bold}: best result.}
    \label{tab:pacs_a}
\end{table*}

\begin{table*}
    \centering
    \resizebox{\textwidth}{!}{
    \begin{tabular}{lllllll}
    \toprule
    \textbf{Method} & \textbf{Photo} & \textbf{Art} & \textbf{Cartoon*} & \textbf{Sketch} & \textbf{Target Avg.} & \textbf{All Avg.}\\
    \midrule
    ERM & 83.6 & 65.7  &- & 60.7 & 70.0 & -\\
    Augmix & 84.599 \scriptsize(0.997) & 68.281 \scriptsize(2.085) & 96.287 \scriptsize(0.940) & 71.097 \scriptsize(0.609) & 74.659 \scriptsize(1.088) & 80.066 \scriptsize(0.897) \\
    \hline
    \color{purple} \scriptsize 1-layer convolutional-based augmentations\\
    RandConv & 73.677 \scriptsize(1.814) & 57.051 \scriptsize(1.764) & 91.660 \scriptsize(0.876) & \bfseries 72.855 \scriptsize(2.314) & 67.861 \scriptsize(1.550) & 73.810 \scriptsize(1.317) \\
    $\text{ALT}_{\text{1-layer}}$ &85.018 \scriptsize(0.874) &67.764 \scriptsize(1.195) &95.865 \scriptsize(0.977) &72.232 \scriptsize(1.039) &75.005 \scriptsize(0.620) &80.220 \scriptsize(0.530) \\
    $\text{ABA}_{\text{1-layer+RandConv}}$ &84.683 \scriptsize(0.997) &68.232 \scriptsize(1.247) &94.937 \scriptsize(0.755) &72.553 \scriptsize(1.159) &75.156 \scriptsize(0.672) &80.101 \scriptsize(0.622) \\
    $\text{ABA}_{\text{1-layer}}$ &84.575 \scriptsize(0.881) &69.463 \scriptsize(1.429) &95.274 \scriptsize(0.560) &72.792 \scriptsize(1.505) &75.610 \scriptsize(0.894) &80.526 \scriptsize(0.636)   \\
    $\text{ABA}_{\text{1-layer+RandConv}}$ &\bfseries85.377 \scriptsize(0.743) &\bfseries 70.371 \scriptsize(1.078) &95.190 \scriptsize(0.430) &72.298 \scriptsize(1.828) &\bfseries 76.016 \scriptsize(0.559) &\bfseries 80.809 \scriptsize(0.468) \\
    \hline
    \color{purple} \scriptsize 3-layer convolutional-based augmentations\\
    $\text{ABA}_{\text{3-layer}}$ &\bfseries 85.269 \scriptsize(1.077) &71.660 \scriptsize(0.977) & 95.274 \scriptsize(0.413) &\bfseries 75.531 \scriptsize(0.791) &\bfseries 77.487 \scriptsize(0.573) &\bfseries 81.934 \scriptsize(0.427) \\
    $\text{ABA}_{\text{3-layer+RandConv}}$ & 85.191 \scriptsize(0.922) & \bfseries 72.034 \scriptsize(1.215) &95.112 \scriptsize(0.578) &74.413 \scriptsize(1.162) &77.219 \scriptsize(0.853) &81.681 \scriptsize(0.630) \\
    \hline
    \color{purple} \scriptsize 5-layer convolutional-based augmentations\\
    $\text{ALT}_{\text{5-layer}}$ &84.575 \scriptsize(1.047) & 68.867 \scriptsize(2.126) & 94.768 \scriptsize(0.430) & 74.421 \scriptsize(0.441) & 75.954 \scriptsize(1.119) & 80.658 \scriptsize(0.929)\\
    $\text{ALT}_{\text{5-layer+RandConv}}$ & 83.916 \scriptsize(0.510) & 68.086 \scriptsize(1.901) & 95.190 \scriptsize(0.686) &\bfseries 74.487 \scriptsize(0.505) & 75.496 \scriptsize(0.799) & 80.420 \scriptsize(0.644) \\
    $\text{ALT}_{\text{5-layer+Augmix}}$ &\bfseries 85.964 \scriptsize(1.098) & 71.943 \scriptsize(1.234) & 94.599 \scriptsize(0.560) & 74.172 \scriptsize(0.752) &\bfseries 77.360 \scriptsize(0.734) & 81.670 \scriptsize(0.667)\\
    $\text{ABA}_{\text{5-layer}}$ & 85.413 \scriptsize(0.885) & 71.660 \scriptsize(0.635) & 95.359 \scriptsize(0.462) &74.426 \scriptsize(1.393) & 77.166 \scriptsize(0.347) &81.715 \scriptsize(0.312) \\
    $\text{ABA}_{\text{5-layer+RandConv}}$ &84.647 \scriptsize(0.498) &71.152 \scriptsize(0.814) &94.937 \scriptsize(0.462) &74.192 \scriptsize(0.950) &76.664 \scriptsize(0.237) &81.232 \scriptsize(0.170) \\
    $\text{ABA}_{\text{5-layer+Augmix}}$ & 85.888 \scriptsize(0.707) &\bfseries 73.470 \scriptsize(1.122) & 95.781 \scriptsize(0.911) & 72.512 \scriptsize(0.818) & 77.290 \scriptsize(0.786) &\bfseries 81.913 \scriptsize(0.569)\\
    \bottomrule
    \end{tabular}
    }
    \caption{SSDG performance on PACS for the C $\rightarrow$ PAS. $^*$Source Domain. \textbf{bold}: best result.}
    \label{tab:pacs_c}
\end{table*}

\begin{table*}
    \centering
    \resizebox{\textwidth}{!}{
    \begin{tabular}{lllllll}
    \toprule
    \textbf{Method} & \textbf{Photo} & \textbf{Art} & \textbf{Cartoon} & \textbf{Sketch*} & \textbf{Target Avg.} & \textbf{All Avg.}\\
    \midrule
    ERM & 35.6 & 28.0 & 54.5 & - & 39.4 & - \\
    Augmix & 46.731 \scriptsize(2.916) & 37.852 \scriptsize(1.878) & 58.575 \scriptsize(1.747) & 94.221 \scriptsize(0.711) & 47.719 \scriptsize(1.723) & 59.345 \scriptsize(1.268) \\
    \hline
    \color{purple} \scriptsize 1-layer convolutional-based augmentations\\
    RandConv & 46.132 \scriptsize(4.879) & \bfseries 52.168 \scriptsize(1.623) & \bfseries 63.942 \scriptsize(2.219) & 94.264 \scriptsize(0.673) &  \bfseries54.081 \scriptsize(1.959) & \bfseries64.126 \scriptsize(1.465) \\
    $\text{ALT}_{\text{1-layer}}$ & 46.024 \scriptsize(3.148) &38.154 \scriptsize(2.171) &59.334 \scriptsize(1.35) &94.422 \scriptsize(0.432) &47.838 \scriptsize(1.947) &59.484 \scriptsize(1.423) \\
    $\text{ALT}_{\text{1-layer+RandConv}}$ & 49.030 \scriptsize(3.242) &38.770 \scriptsize(2.128) &59.829 \scriptsize(1.358) &94.673 \scriptsize(0.293) &49.210 \scriptsize(2.142) &60.576 \scriptsize(1.672) \\
    $\text{ABA}_{\text{1-layer}}$&\bfseries 52.587 \scriptsize(2.390) & 43.643 \scriptsize(5.125) & 62.287 \scriptsize(2.124) &94.523 \scriptsize(0.293) &52.839 \scriptsize(2.804) &63.260 \scriptsize(2.150) \\
    $\text{ABA}_{\text{1-layer+RandConv}}$ &49.713 \scriptsize(3.434) &39.199 \scriptsize(1.736) &61.715 \scriptsize(1.382) &94.673 \scriptsize(0.700) &50.209 \scriptsize(1.934) &61.325 \scriptsize(1.471) \\
    \hline
    \color{purple} \scriptsize 3-layer convolutional-based augmentations\\
    $\text{ABA}_{\text{3-layer}}$&52.587 \scriptsize(2.967) &\bfseries45.635 \scriptsize(3.936) &\bfseries63.063 \scriptsize(1.511) &94.623 \scriptsize(0.201) &\bfseries53.762 \scriptsize(2.457) &\bfseries63.977 \scriptsize(1.804) \\
    $\text{ABA}_{\text{3-layer+RandConv}}$&\bfseries52.780 \scriptsize(1.472) &44.764 \scriptsize(1.767) &63.022 \scriptsize(0.815) &94.822 \scriptsize(0.694) &53.523 \scriptsize(0.900) &63.842 \scriptsize(0.736) \\
    \hline
    \color{purple} \scriptsize 5-layer convolutional-based augmentations\\
    $\text{ALT}_{\text{5-layer}}$ & 49.305 \scriptsize(2.775) & 39.658 \scriptsize(3.423) & 61.109 \scriptsize(1.853) &  94.573 \scriptsize(0.466) & 50.024 \scriptsize(2.408) & 61.161 \scriptsize(1.726) \\
    $\text{ALT}_{\text{5-layer+RandConv}}$ & 51.305 \scriptsize(0.866) & 41.787 \scriptsize(1.174) & 62.773 \scriptsize(1.089) & 94.724 \scriptsize(0.527) & 51.955 \scriptsize(0.791) & 62.647 \scriptsize(0.571) \\
    $\text{ALT}_{\text{5-layer+Augmix}}$ & 49.078 \scriptsize(2.072) & 40.186 \scriptsize(2.494) & 62.901 \scriptsize(0.358) & 94.271 \scriptsize(0.624) & 50.721 \scriptsize(1.414) & 61.609 \scriptsize(1.103)\\
    $\text{ABA}_{\text{5-layer}}$ &49.461 \scriptsize(2.584) &45.576 \scriptsize(4.069) &\bfseries 64.522 \scriptsize(0.425) &94.171 \scriptsize(0.801) &53.186 \scriptsize(2.072) &63.433 \scriptsize(1.662) \\
    $\text{ABA}_{\text{5-layer+RandConv}}$&\bfseries52.982 \scriptsize(1.960) &\bfseries 46.299 \scriptsize(2.702) &63.106 \scriptsize(0.651) &93.719 \scriptsize(0.449) &\bfseries 54.129 \scriptsize(1.334) &\bfseries64.026 \scriptsize(0.970) \\
    $\text{ABA}_{\text{5-layer+Augmix}}$ & 51.377 \scriptsize(0.240) & 43.335 \scriptsize(0.610) & 62.351 \scriptsize(0.747) & 94.472 \scriptsize(0.001) & 52.354 \scriptsize(0.034) & 62.884 \scriptsize(0.026) \\
    \bottomrule
    \end{tabular}
    }
    \caption{SSDG performance on PACS for the S $\rightarrow$ PAC. $^*$Source Domain. \textbf{bold}: best result.}
    \label{tab:pacs_s}
\end{table*}

\end{document}